\documentclass[
11pt,              
letterpaper,       
fleqn,
headnosepline,	   
oneside,           
onecolumn,         
openright,         
noonelinecaption,
cleardoubleempty,
smallheadings]
{scrartcl}         
\addtokomafont{caption}{\sffamily\footnotesize}
\setkomafont{pagenumber}{\sffamily\footnotesize}
\addtokomafont{pagenumber}{\sffamily\footnotesize}
\setkomafont{captionlabel}{\sffamily\bfseries\footnotesize}
\setcapindent{0em} 

\usepackage{makecell}
\usepackage[usestackEOL]{stackengine}
\usepackage{graphicx}
\usepackage{psfrag}
\usepackage{amsmath}
\usepackage{amssymb}
\usepackage{color}
\usepackage{xcolor}
\usepackage{rotate}
\usepackage{makeidx}
\usepackage{cite}
\usepackage{eufrak}
\usepackage[varumlaut]{yfonts}
\usepackage{helvet}   
\usepackage{mathpazo}
\normalfont
\usepackage[T1]{fontenc}
\usepackage{textcomp}
\usepackage{longtable} 
\usepackage[only,llbracket,rrbracket]{stmaryrd}
\usepackage{bm}
\begin{document}
\newcommand{\beq} {\begin{equation}}
\newcommand{\eeq} {\end{equation}}
\newcommand{\D}   {\displaystyle}
\newcommand{\divg}{\mbox{\rm{div}}\,}
\newcommand{\clearemptydoublepage}{\newpage{\pagestyle{empty}\cleardoublepage}}
\newcommand{\Divg}{\mbox{\rm{Div}}\,}
\newtheorem{remark}      {\bf{\sffamily{Remark}}}
\newtheorem{definition}  {\bf{\sffamily{Definition}}}
\renewcommand{\sc}{}
\renewcommand{\Psi}{\psi}
\renewcommand{\varrho}{\vartheta}
\renewcommand{\arraystretch}{1.3}
\sloppy
\def\sca   #1{\mbox{\rm #1}{}}
\def\mat   #1{\mbox{\bf #1}{}}
\def\vec   #1{\mbox{\boldmath $#1$}{}}
\def\ten   #1{\mbox{\boldmath $#1$}{}}
\def\scas  #1{\mbox{{\scriptsize{${\rm{#1}}$}}}{}}
\def\vecs  #1{\mbox{{\boldmath{\scriptsize{$#1$}}}}{}}
\def\tens  #1{\mbox{{\boldmath{\scriptsize{$#1$}}}}{}}
\def\up    #1{^{\mbox{\rm{\footnotesize{#1}}}}}
\def\down  #1{_{\mbox{\rm{\footnotesize{#1}}}}}
\def\ltr   #1{\mbox{\sf{#1}}}
\def\bltr  #1{\mbox{\sffamily{\bfseries{{#1}}}}}
\vspace*{0.8cm}
\begin{center}
{\sffamily\bfseries\Large{On sparse regression, Lp-regularization}}\\[4pt]
{\sffamily\bfseries\Large{and automated model discovery}}\\
\vspace*{0.8cm}
Jeremy A. McCulloch$^*$,
Skyler R. St. Pierre$^*$,
Kevin Linka$^{**}$,
Ellen Kuhl$^*$\\ [6.pt]
$^*$ Department of Mechanical Engineering \\
Stanford University, Stanford, California, United States \\[4.pt]
$^{**}$ Institute for Continuum and Material Mechanics\\
Hamburg University of Technology, Hamburg, Germany\\
\vspace*{0.9cm}
{\small{\it{
This manuscript is dedicated to Robert L. Taylor on the occasion of his 90th birthday and to his many contributions to the finite element method and his open source software FEAP. Congratulations, Bob: tang,,1!}}}
\end{center}
\vspace*{0.6cm}
{\sffamily{\bfseries{Abstract.}}}
Sparse regression and feature extraction are the cornerstones of knowledge discovery from massive data.
Their goal is to discover interpretable and predictive models that provide simple relationships among scientific variables.
While the statistical tools for model discovery are well established in the context of linear regression, their generalization to nonlinear regression in material modeling is highly problem-specific and insufficiently understood.  
Here we explore the potential of neural networks for automatic model discovery and induce sparsity by a hybrid approach that combines two strategies: regularization and physical constraints. 
We integrate the concept of Lp regularization for subset selection with constitutive neural networks that leverage our domain knowledge in kinematics and thermodynamics. 
We train our networks with both, synthetic and real data, and perform several thousand discovery runs to infer common guidelines and trends:
L2 regularization or ridge regression is unsuitable for model discovery; 
L1 regularization or lasso promotes sparsity, but induces strong bias that may aggressively change the results; 
only L0 regularization allows us to transparently fine-tune the trade-off between interpretability and predictability, simplicity and accuracy, and bias and variance.
With these insights, we demonstrate that Lp regularized constitutive neural networks can simultaneously discover both, interpretable models and physically meaningful parameters.
We anticipate that our findings will generalize to alternative discovery techniques such as sparse and symbolic regression, and to other domains such as biology, chemistry, or medicine. 
Our ability to automatically discover material models from data could have tremendous applications in generative material design and open new opportunities to manipulate matter, alter properties of existing materials, and discover new materials with user-defined properties.

\vspace*{0.5cm}
{\sffamily{\bfseries{Keywords.}}}
constitutive modeling;
automated model discovery;
sparse regression;
Lp regularization;
hyperelasticity

\clearpage
\section{Motivation}
The ability to discover meaningful constitutive models from data would forever change how we understand, model, and design new materials and structures. 
Massive advancements in data science are now bringing us closer than ever towards this goal \cite{alber19,brunton19}.
Throughout the past three years, numerous research groups have begun to harness the potential of neural networks and fit constitutive models to experimental data \cite{asad22,fuhg22a,ghaderi20,holthusen23,holzapfel21,huang20,klein22,masi21,peirlinck23,tac22,tac23,wang23}, an approach that is now widely known as {\it{constitutive neural networks}} \cite{linka21}. 
While initial studies have used neural networks exclusively as black box regression operators \cite{ghaboussi91}, recent approaches 
are increasingly recognizing their potential to discover not only the model parameters, but also the model itself \cite{linka23}.
The paradigm of {\it{automated model discovery}} was first formalized in the context of nonlinear dynamical systems more than a decade ago
to discover Lagrangians and Hamiltonians for 
oscillators \cite{bongard07}, 
pendula, biological processes \cite{schmidt09},
or turbulent fluid flows \cite{brunton16}.
It is now rapidly gaining popularity in the context of constitutive modeling \cite{peng21}, and several promising techniques have emerged to decipher constitutive relations between stresses and strains \cite{peirlinck23}, and even integrate them automatically into a finite element analysis \cite{peirlinck23a,simo85,zienkiewicz87}.
These not only include constitutive neural networks
\cite{linka21}, but also sparse regression \cite{flaschel21}, genetic programming in the form of symbolic regression \cite{abdusalamov23}, and variational system identification \cite{wang21}.\\[6.pt]
The holy grail in automated model discovery is to identify 
{\it{generalizable}} and truly {\it{interpretable}} models 
with {\it{physically meaningful}} parameters \cite{brunton19,flaschel23,linka23a}.  
Ideally, we want to discover a concise, yet simple and interpretable model with only a few relevant terms that best explains experimental data, while remaining robust to outliers and noise.  
In terms of statistical learning, this translates model discovery into a {\it{subset selection}} or feature extraction task. 
Subset selection and shrinkage methods are by no means new; in fact, they have been extensively studied for many decades \cite{frank93,hoerl70,santosa86,tibshirani96}. 
In the context of linear regression, these methods have become standard textbook knowledge \cite{hastie09}.  
In the context of nonlinear regression, when analytical solutions are rare, subset selection is much more nuanced, general recommendations are difficult, and feature extraction becomes highly problem-specific \cite{james13}. 
To be clear, this limitation is not exclusively inherent to automated model discovery with constitutive neural networks--it applies to 
distilling scientific knowledge from data in general \cite{schmidt09}.
This includes alternative model discovery approaches like 
sparse regression \cite{brunton16,flaschel21}, or
symbolic regression \cite{abdusalamov23,koza92,schmidt09}.
The key question to the success of discovering new knowledge from data is:
How do we robustly discover the best interpretable model with a small subset of relevant terms?
And, probably equally importantly: 
What is the trade-off between {\it{interpretability}} and prediction {\it{accuracy}}?
To frame these questions more broadly, let us first revisit the notions of regression and neural networks in the context of constitutive modeling: \\[6.pt]
\noindent
{\bf{\sffamily{Regression.}}}
Regression is a statistical method to examine the relationship between a {\it{dependent variable}}, in constitutive modeling in solid mechanics the stress $\ten{\sigma}$, and one or more {\it{independent variables}}, in this case the strain $\ten{\varepsilon}$, using a model that depends on a set of {\it{model parameters}} $\vec{\theta}$. Here regression has two main objectives: characterizing the form and strength of the relationship between stress and strain to enable predictions, and providing insights into how stress and strain are correlated \cite{hastie09}. 
Popular types of regression are 
{\it{logistic regression}}, assuming for example a binary relationship;
{\it{linear regression}} \cite{seber77}, assuming a relationship that is linear in the model parameters $\vec{\theta}$, or
{\it{nonlinear regression}}, assuming a relationship that is nonlinear in the model parameters $\vec{\theta}$, 
as we do throughout this manuscript. 
Regression is the cornerstone of statistical learning \cite{james13}. It provides tools to decipher relationships within data, but its application to constitutive modeling requires attention to physical constraints including objectivity, symmetry, incompressibility, polyconvexity, or thermodynamic consistency \cite{antman05,fuhg22,hartmann01,linden23,tac23a,truesdell65}.
As a natural consequence, we cannot just use {\it{any}} set of functions to build our constitutive model: While polynomial functions between stresses and strains associated with a linear regression would be ideal from an optimization point of view, these models may violate thermodynamic constraints, which favor exponential or power functions associated with a nonlinear regression.\\[6.pt] 
\noindent
{\bf{\sffamily{Linear regression.}}}
Linear regression \cite{seber77} seeks to model the relationship between a {\it{dependent variable}}, in our case the stress $\sigma$, and a set of one or more {\it{independent variables}}, in our case the set of strains $\varepsilon_i$ at different load levels $i$, using a function that depends {\it{linearly}} on the {\it{model parameter}} $\theta = \{\, {\rm{E}} \,\}$, in this case the elastic modulus or Young's modulus. 
The regression estimates this parameter by minimizing the 
difference between the predicted stress, $\sigma_i = {\rm{E}} \, \varepsilon_i$, for given strains $\varepsilon_i$ and stiffness ${\rm{E}}$, and the experimentally measured stresses $\hat{\sigma}_i$, divided by the number of data points $n_{\rm{data}}$.
A common measure for this difference is the mean squared error
based on the $L_2$-norm \cite{hastie09},
$ || \, (\,\circ\,)  \, ||
= || \, (\,\circ\,)  \, ||_2
= |\,(\,\circ\,)^2 \,|^{1/2}$, for which the minimization problem becomes 
\beq
  L({\rm{E}}; {\varepsilon}) 
= \frac{1}{n_{\rm{data}}}
  \sum_{i=1}^{n_{\rm{data}}}  
 || \, {\rm{E}} \, {\varepsilon}_i - \hat{\sigma}_i  \, ||^2  
 \rightarrow \min_{\rm{E}}
 \qquad \mbox{... \, estimate model parameter} \; \mbox{E} \,.
 \label{regression_linear}
\eeq
When phrased as least square's problems, 
provided 
linear regression problems have a {\it{convex}} objective function with a {\it{unique global minimum}}.
For our example (\ref{regression_linear}), we can find it by evaluating the vanishing derivative,  
$\partial L / \partial {\rm{E}} 
=\sum_{i=1}^{n_{\rm{data}}}  
 2 \varepsilon_i \, ( \, {\rm{E}} \, \varepsilon_i - \hat{\sigma}_i  )   
\doteq 0$.
Here, the minimization problem is not only linear in the model parameter ${\rm{E}}$, but also in the dependent variable $\varepsilon$, and we obtain an explicit solution for the elastic modulus,
${\rm{E}}
=\sum_{i}^{n_{\rm{data}}} \varepsilon_i \, \hat{\sigma}_i 
\,/\,
 \sum_{i}^{n_{\rm{data}}} \varepsilon_i \, \varepsilon_i$.
For linear regression with multiple model parameters $\vec{\theta}$, for which the minimization problem is a linear function in the dependent variables $\varepsilon$, we obtain a {\it{coupled system of equations}}, 
$\partial L / \partial \vec{\theta} \doteq \vec{0}$,
with an explicit solution for the parameter vector $\vec{\theta}$.
For linear regression with one or multiple model parameters $\vec{\theta}$, for which the minimization problem is a nonlinear function in the dependent variables $\varepsilon$, we obtain a similar set of one or more equations 
$\partial L / \partial \vec{\theta} \doteq \vec{0}$, which may require an iterative solution for the parameter vector $\vec{\theta}$.
Importantly, {\it{any}} regression that is linear in the model parameters $\vec{\theta}$--independent of whether it is linear, polynomial, or generally nonlinear in the independent variables $\varepsilon_i$--is considered a linear regression problem 
that, when phrased as a least square's problem with appropriate data, results in a convex quadratic function in the model parameters $\vec{\theta}$ with a unique global minimum. \\[6.pt]
\noindent
{\bf{\sffamily{Nonlinear regression.}}}
Nonlinear regression \cite{seber89} seeks to model the relationship between a {\it{dependent variable}}, in our case the Piola stress $\ten{P}$, and a set of one or more {\it{independent variables}}, in our case the set of deformation gradients $\ten{F}_i$ at different load levels $i$, 
using a function that depends {\it{nonlinearly}} on a set of 
{\it{model parameters}} $\vec{\theta}$ \cite{bates88}. 
The regression estimates these parameters by minimizing the 
difference between the predicted stress, $\ten{P} (\ten{F}_i, \vec{\theta})$, for given deformation gradients $\ten{F}_i$ and model parameters $\vec{\theta}$, and the experimentally measured stresses $\hat{\ten{P}}_i$, divided by the number of data points $n_{\rm{data}}$.
Similar to linear regression, we can measure for this difference as the mean squared error
based on the $L_2$-norm \cite{hastie09},
$ || \, (\,\circ\,)  \, ||
= || \, (\,\circ\,)  \, ||_2
= |\,(\,\circ\,)^2 \,|^{1/2}$, for which the minimization problem becomes 
\beq
  L(\vec{\theta}; \ten{F}) 
= \frac{1}{n_{\rm{data}}}
  \sum_{i=1}^{n_{\rm{data}}}  
 || \, \ten{P} (\ten{F}_i, \vec{\theta}) - \hat{\ten{P}}_i  \, ||^2  
 \rightarrow \min_{\vecs{\theta}}
 \qquad \mbox{... \, estimate model parameters} \; \vec{\theta} \,.
 \label{regression_nonlinear}
\eeq
In general, nonlinear regression problems have a {\it{non-convex}} objective function with  {\it{multiple local minima}}. Solving non-convex optimization problems requires iterative algorithms that are at risk of converging to a local minimum instead of the global minimum, and their solution is often highly sensitive to the initial conditions that we select for the parameter vector. 
Depending on the nature of the problem, 
the solution we find may involve a large and dense parameter vector 
$\vec{\theta}$, and overfitting may occur when 
the number of parameters is larger than the number of data points,
$n_{\rm{para}} > n_{\rm{data}}$.
Notably, even for many data points, we may face overfitting
when the data are noisy or 
not rich enough to sufficiently activate all the parameters.
For example, with tension and compression tests alone, we cannot estimate model parameters for shear. \\[6.pt]
\noindent
{\bf{\sffamily{Sparse regression.}}}
Sparse regression is a special type of regression that seeks to prevent overfitting by inducing sparsity in the parameter vector $\vec{\theta}$ by setting a large number of parameters to zero \cite{tibshirani96}. 
Sparse regression is particularly useful in high-dimensional settings, since it generates models with a small subset of non-zero parameters \cite{hastie09}, which tends to make the model more interpretable \cite{james13}. 
Historically, the need for sparse regression emerged prominently with the advent of high-dimensional datasets for which the number of parameters can easily exceed the number of independent observations \cite{champion19}. 
A prominent example is SINDy, an algorithm for sparse identification in nonlinear dynamics that promotes sparsity through sequential thresholded least-squares by 
iterating between a partial least-squares fit and a thresholding step to
sequentially drop the least relevant terms of a model \cite{brunton16}. 
Importantly, while these sparsification algorithms converge well in linear regression associated with convex objective functions \cite{zhang19}, their convergence is no longer guaranteed in nonlinear regression with non-convex objective functions.
The advantages of sparse regression are  {\it{improved interpretability}} by reducing the parameter set to only a few non-zero terms; {\it{feature selection}} by identifying the most relevant terms; and {\it{reduced risk of overfitting}} by promoting simpler models. These advantages come at a price: The disadvantages of sparse regression are {\it{selection bias}} by enforcing sparsity of the parameter estimates; {\it{additional hyperparameters}} that need to be tuned and require additional attention; and {\it{risk of misspecification}} by excluding relevant parameters if sparsity is enforced too aggressively. In conclusion, sparse regression offers a powerful toolset for high-dimensional modeling, but introduces a trade-off between interpretability and prediction accuracy.\\[6.pt]
\noindent
{\bf{\sffamily{Neural networks.}}}
Neural networks are a class of models and algorithms that can approximate a wide range of functions \cite{hornik89}. Their versatility not only makes them a powerful tool for classification, reinforcement learning, and generative tasks, but also for {\it{regression}} problems \cite{goodfellow16}, in our context, for regression in constitutive modeling \cite{ghaboussi91}. 
Neural networks consist of input, hidden, and output layers with several nodes in each layer. Their parameters are the network weights
$\vec{\theta}=\{ w_{i,j} \}\,$, where 
$i=1,...,n_{\rm{lay}}$ is the number of hidden layers and 
$j=1,...,n_{\rm{nod}}$ is the number of nodes per layer. 
During training, neural networks effectively perform a regression 
as they learn their parameters
by minimizing a loss function $L$ that penalizes the error between model and data. 
Similar to the classical nonlinear regression in equation (\ref{regression_nonlinear}), 
we can characterize this error as the mean squared error, 
the $L_2$-norm of the difference between the stress predicted by the model 
$\ten{P}(\ten{F}_i)$ and the experimentally measured stress $\hat{\ten{P}}_i$, 
divided by the number of data points $n_{\rm{data}}$ to train the model,
\beq
  L (\vec{\theta} ; \ten{F})
= \frac{1}{n_{\rm{data}}} 
  \sum_{i=1}^{n_{\rm{data}}}
|| \, \ten{P}(\ten{F}_i, \vec{\theta}) - \hat{\ten{P}}_i \, ||^2 
  \rightarrow \mbox{min}
  \qquad \mbox{... \, learn network weights} \;
  \vec{\theta}=\{ w_{i,j} \} \,.
\label{regression_NN}
\eeq
However, in contrast to traditional regression tools that have a fixed functional form, neural networks can easily adapt their shape 
which allows them to model complex functions \cite{ghaderi20},
either linear \cite{stpierre23} or nonlinear \cite{linka23a}
in the model parameters $\vec{\theta}$.
The advantages of neural networks are their 
{\it{universal approximation}} that allows them to approximate {\it{any}} continuous function for a sufficiently large number of weights \cite{hornik89},
and their inherent
{\it{flexibility}} that allows them to model high-dimensional nonlinear relationships like the constitutive behavior we seek to model here.
Their disadvantages are their
{\it{computational complexity}}, especially for densely connected architectures with multiple hidden layers;
{\it{risk of overfitting}} sparse or noisy data; and 
{\it{lack of interpretability}} that generally worsens with the number of layers and makes plain neural networks unsuitable for model discovery tasks.\\[6.pt]
\noindent
{\bf{\sffamily{Sparse neural networks.}}}
Sparse neural networks use a special type of network architecture for which a large number of weights are zero. This reduces the number of active connections between the nodes of consecutive layers. Sparsity can be induced {\it{during}} training by using special algorithms, or {\it{after}} training by pruning \cite{han15}.
The concepts of sparse neural networks and weight or node pruning--inspired by brain development and synaptic pruning--have gained increasing attention with the rise of deep learning and the need for computational efficiency \cite{lecun89}.
The advantages of sparse neural networks are their
{\it{computational efficiency}} and faster inference times;
{\it{reduced risk of overfitting}} by promoting smaller model sizes; and
their potential for {\it{regularization}} and improved {\it{generalization}}.
The disadvantages are
{\it{increased training complexity}} to induce sparsity; and
{\it{risk of decreased performance}} through overly aggressive sparsification or pruning.\\[6.pt]
The objective of this review is to explore neural networks 
for which sparsity is induced by a hybrid approach of two combined strategies:
{\it{regularization}} 
and {\it{physical constraints}}. 
Towards this goal,
first, we review popular subset selection and shrinkage methods to induce sparsity within the general framework of $L_p$ {\it{regularization}} in Section \ref{Lp_regularization}. Then, we discuss how to leverage physical constraints to induce sparsity within the framework of {\it{constitutive neural networks}} in Section \ref{neural_networks}. Finally, we propose a hybrid approach of $L_p$ regularized constitutive neural networks for automated model discovery and discuss the interpretability and prediction accuracy of their discovered models by means of a library of illustrative examples in Section \ref{Lp_networks}. We conclude by comparing the different approaches and providing guidelines and recommendations in Section \ref{conclusion}.   
\section{L$\!_{\mbox{\normalsize{p}}}$ regularization}
\label{Lp_regularization}
\begin{figure}[h]
\centering
\includegraphics[width=0.72\linewidth]{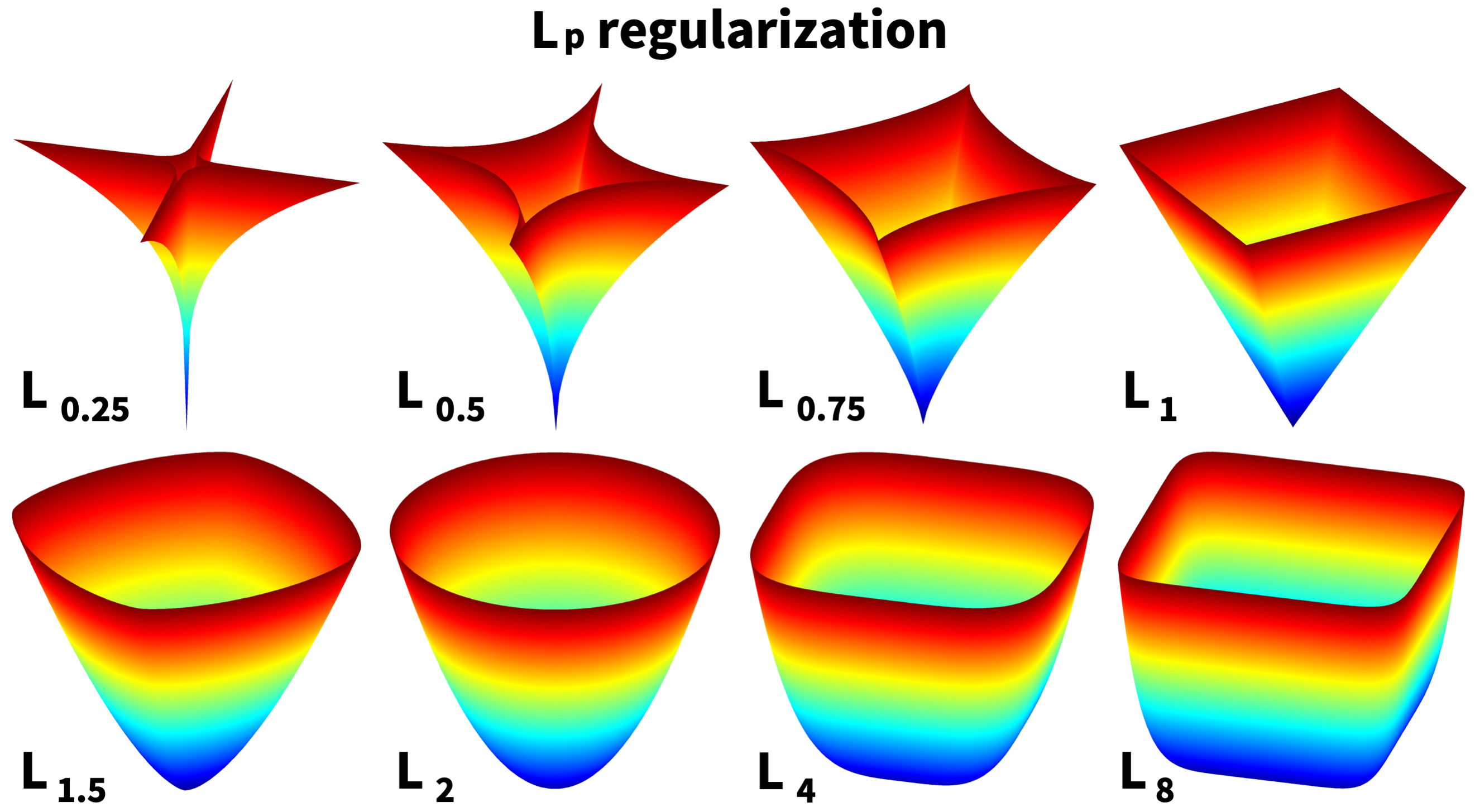}
\caption{{\bf{\sffamily{Lp regularization.}}} 
Contours of regularization term,
  $L_{\rm{p}} 
= \alpha \, \sum_{i=1}^{n_{\rm{para}}}  || \, \vec{\theta} \, ||_p^p$ 
with
$|| \, \vec{\theta} \, ||_p^p
= |\,w_{i}\,|^p$,
for varying powers, 
$p = [0.25, 0.5, 0.75, 1, 1.5, 2, 4, 8]$,
evaluated for two parameters, $w_1$ and $w_2$.
For $p \le 1$, top row, with the special case of 
$L_1$ regularization or lasso represented through the pyramid,
$L_{\rm{p}}$ regularization promotes sparsity by setting some weights exactly to zero, 
but is no longer strictly convex and can have multiple local minima.
For $p  >  1$, bottom row, with the special case of 
$L_2$ regularization or ridge regression represented through the ellipsoid, 
$L_{\rm{p}}$ regularization promotes stability by reducing outliers, 
while the regularization term remains convex.}
\label{fig01}
\end{figure}

The concept of $L_p$ regularization or bridge regression 
was first introduced three decades ago in the context of chemometrics with the goal to shrink the parameter space in chemical data analysis \cite{frank93}.
The method has re-gained attention as a powerful tool to promote sparsity in system identification \cite{brunton16}, and, most recently, in discovering constitutive models from data \cite{flaschel21}.
$L_p$ regularization is a generalized regularization technique that uses the $L_p$ norm of the parameter vector $\vec{\theta}$, the sum of the $p$-th power of the norm of its $i=1,...,n_{\rm{para}}$ coefficients $w_i$, raised to the inverse power,
$ || \, \vec{\theta} \, ||_p 
= [\, \mbox{$ \sum_{i=1}^{n_{\rm{para}}}$} |\, w_i \,|^p\,]^{1/p}$.
Bridge regression constrains the regression (\ref{regression_nonlinear}) by constraining this $L_p$ norm to be smaller than a non-negative constant $\epsilon\ge0$, 
\beq
  L(\vec{\theta}; \ten{F}) 
= \frac{1}{n_{\rm{data}}}
  \sum_{i=1}^{n_{\rm{data}}} 
 || \, \ten{P} (\ten{F}_i, \vec{\theta}) - \hat{\ten{P}}_i  \, ||^2 
  \rightarrow \mbox{min}
  \qquad
  \mbox{subject to}
  \qquad
  || \, \vec{\theta} \, ||_p^p 
= \sum_{i=1}^{n_{\rm{para}}} 
 |\, w_i \,|^p \le \epsilon \,.
\label{regression_constrained}
\eeq
It proves convenient to reformulate this {\it{constrained regression}} problem as a {\it{penalized regression}} problem, by penalizing the regression (\ref{regression_nonlinear}) with a penalty term that consists of the $L_p$ norm $|| \, \vec{\theta} \, ||_p^p$, multiplied by a non-negative penalty parameter $\alpha \ge 0$,
\beq
  L(\vec{\theta}; \ten{F}) 
= \frac{1}{n_{\rm{data}}}
  \sum_{i=1}^{n_{\rm{data}}} 
 || \, \ten{P} (\ten{F}_i, \vec{\theta}) - \hat{\ten{P}}_i  \, ||^2 
+ \alpha \, || \, \vec{\theta} \, ||_p^p
  \rightarrow \mbox{min}
  \qquad
  \mbox{with}
  \qquad
  || \, \vec{\theta} \, ||_p^p 
= \sum_{i=1}^{n_{\rm{para}}} 
  |\, w_i \,|^p \,.
\label{regression_penalized}
\eeq 
The constrained and penalized regression problems (\ref{regression_constrained}) and (\ref{regression_penalized}) are equivalent, which implies that for a given $\epsilon \ge 0$, there exists an $\alpha \ge 0$ such that the two problems share the same solution $\vec{\theta}$ \cite{fu98}.
The flexible power $p$ not only allows us to recover classical regularization techniques like $L_2$ or $L_1$ regularization as special cases, but also to interpolate smoothly between these different methods \cite{frank93}. 
The advantages of $L_p$ regularization are 
its inherent {\it{flexibility}} by allowing for a continuum of popular regularization techniques for varying powers $p$; and 
its potential to effectively {\it{promote sparsity}}.
Its disadvantages are 
the added complexity associated with the {\it{selection of hyperparameters}}, specifically the penalty parameter $\alpha$ and the power $p$; and the potential {\it{computational challenges}} associated with specific choices for $p$.
Figure \ref{fig01} illustrates the contours of the regularization term, 
for varying powers $p$ as $p = [0.25, 0.5, 0.75, 1, 1.5, 2, 4, 8]$, 
evaluated for two parameters, $w_1$ and $w_2$.
The top row illustrates the effect of powers smaller than or equal to one, $p \le 1$; the bottom row illustrates the effect powers larger than one, $p > 1$.
In the following, we highlight the most popular special cases of $L_p$ regularization, their history, and their advantages and disadvantages.
\\[6.pt]
\noindent
{\bf{\sffamily{L$_{\bf{2}}$ regularization or ridge regression.}}} 
$L_2$ regularization, commonly known to as ridge regression, 
was introduced more than half a century ago to address multicollinearity in regression analysis \cite{hoerl70}, 
and has gained attention for its ability to stabilize parameter estimates, especially when the parameters are closely correlated \cite{hastie09}.
It uses the $L_2$ norm of a vector, the Euclidian norm, the sum of the vector components squared,
$ || \, \vec{\theta} \, ||_2 
= [\, \sum_{i=1}^{n_{\rm{para}}} | \, w_i \, |^2\,]^{1/2}$.
Notably, the $L_2$ norm does {\it{not}} weigh all entries of the vector equally. 
Instead, it squares the vector entries which makes it highly sensitive to outliers as it penalizes the squared magnitude of the individual parameters $w_i$.
Ridge regression supplements the regression (\ref{regression_nonlinear}) with a penalty term that consists of the $L_2$ norm multiplied by a penalty parameter $\alpha$,
\beq
  L(\vec{\theta}; \ten{F}) 
= \frac{1}{n_{\rm{data}}}
  \sum_{i=1}^{n_{\rm{data}}} 
 || \, \ten{P} (\ten{F}_i, \vec{\theta}) - \hat{\ten{P}}_i  \, ||^2 
+ \alpha \, || \, \vec{\theta} \, ||_2^2
 \rightarrow \mbox{min}
 \qquad
 \mbox{with}
 \qquad
 || \, \vec{\theta} \, ||_2^2 
= \sum_{i=1}^{n_{\rm{para}}}  
 | \, w_i \, |^2 \,.
\eeq 
Its advantages are 
{\it{stability in multicollinearity}} by offering stable parameter estimates even in the presence of highly correlated predictors;
{\it{managing outliers}} and {\it{preventing overfitting}} 
by quadratically penalizing extreme coefficients; and 
{\it{computational efficiency}} even for large datasets. 
Its disadvantages are {\it{introducing bias}},
which may result in underestimating certain coefficients and effects;
and its {\it{inability to induce sparsity}}, which makes it unsuitable for our current focus of subset selection.
Figure \ref{fig01} shows that, for the special case of 
$L_2$ regularization, the regularization term adopts a convex ellipsoidal shape that
promotes stability by reducing outliers.\\[6.pt]
\noindent
{\bf{\sffamily{L$_{\bf{1}}$ regularization or lasso.}}} 
$L_1$ regularization was initially introduced as a method to analyze seismograms in geophysics almost four decades ago \cite{santosa86}, and has become widely known under the name of lasso, short for Least Absolute Shrinkage and Selection Operator \cite{tibshirani96}. Lasso has become popular for producing interpretable models \cite{fuhg22a}, while exhibiting the same stability properties as ridge regression. It uses the $L_1$ norm of a vector, the sum of the absolute values of its components, 
$|| \, \vec{\theta} \, ||_1 = \sum_{i=1}^{n_{\rm{para}}} |\, w_i \,|$.
Because of its similarities with a distance between city bocks, the $L_1$ norm is often referred to as the Manhattan distance or taxicab norm. 
Notably, the $L_1$ norm weighs all entries of the vector {\it{equally}} and is therefore less sensitive to outliers than the $L_2$ norm. 
Lasso supplements the regression (\ref{regression_nonlinear}) with a penalty term that consists of the $L_1$ norm multiplied by a penalty parameter $\alpha$,
\beq
  L(\vec{\theta}; \ten{F}) 
= \frac{1}{n_{\rm{data}}}
  \sum_{i=1}^{n_{\rm{data}}} 
 || \, \ten{P} (\ten{F}_i, \vec{\theta}) - \hat{\ten{P}}_i  \, ||^2 
+ \alpha \, || \, \vec{\theta} \, ||_1
 \rightarrow \mbox{min}
 \qquad
 \mbox{with}
 \qquad
 || \, \vec{\theta} \, ||_1 
= \sum_{i=1}^{n_{\rm{para}}}  
 | \, w_i \, |  \,.
\eeq
Its advantages are
{\it{enabling feature selection}} and {\it{inducing sparsity}} 
by reducing some weights exactly to zero which
effectively reduces model complexity
and improves interpretability;
{\it{mitigating overfitting}} by constraining the magnitude of the weights, which is especially important when data are limited or high-dimensional; and
{\it{providing predictive insights}} by identifying the most relevant weights \cite{hastie09}.
Its disadvantages are
{\it{introducing bias}}, which may result in underestimating certain coefficients and effects; and
{\it{focussing on selective effects}} while discarding others, 
especially in nuanced multi-effect situations when the weights are closely correlated.
Figure \ref{fig01} shows that, for the special case of 
$L_1$ regularization, the regularization term adopts a non-strictly-convex pyramid shape that promotes sparsity by reducing some weights exactly to zero. \\[6.pt]
\noindent
{\bf{\sffamily{L$_{\bf{1/2}}$ regularization or elastic net.}}} 
$L_{1/2}$ regularization, also known as elastic net, is a hybrid approach that seeks to combine the benefits of both $L_1$ and $L_2$ regularization \cite{zou05}. 
The elastic net supplements the regression (\ref{regression_nonlinear}) with two penalty terms in terms of the $L_2$ and $L_1$ norms multiplied by two independent penalty parameters $\alpha_2$ and $\alpha_1$, 
\beq
\begin{array}{l}
\D{L(\vec{\theta}; \ten{F}) 
= \frac{1}{n_{\rm{data}}}
  \sum_{i=1}^{n_{\rm{data}}} 
 || \, \ten{P} (\ten{F}_i, \vec{\theta}) - \hat{\ten{P}}_i  \, ||^2 
+ \alpha_2 \, || \, \vec{\theta} \, ||_2^2
+ \alpha_1 \, || \, \vec{\theta} \, ||_1
 \rightarrow \mbox{min}} \\[-8.pt]
\D{\hspace*{6.0cm}
 \mbox{with}
 \quad
 || \, \vec{\theta} \, ||_2^2 
= \sum_{i=1}^{n_{\rm{para}}}  
 | \, w_i \, |^2  
 \;\,
 \mbox{and}
 \;\,
 || \, \vec{\theta} \, ||_1 
= \sum_{i=1}^{n_{\rm{para}}} 
 | \, w_i \, |  \,.}
\end{array} 
\eeq
For $\alpha_1=0$ and $\alpha_2=0$, it recovers the classical ridge regression and lasso as special cases. For $\alpha_1 > 0$ and $\alpha_2>0$, $L_{1/2}$ regularization shares many features with $L_{p}$ regularization with $1<p<2$ and generates contours similar, but {\it{not identical}} to Figure \ref{fig01}, bottom left. However, in contrast to $L_{p}$ regularization with $1<p<2$ \cite{frank93}, $L_{1/2}$ regularization not only {\it{promotes stability}}, but also  {\it{induces sparsity}} while remaining convex \cite{hastie09}. 
Its disadvantage is its added {\it{computational complexity}}. 
Since $L_{1/2}$ regularization is {\it{not}} a special case of the $L_p$ regularization family, 
we will not consider it further throughout this study.\\[6.pt]
\noindent
{\bf{\sffamily{L$_{\bf{0.5}}$ regularization.}}} 
$L_p$ regularization with powers $0<p<1$ has become a popular tool in 
subset selection since it promotes sparsity more aggressively than simple $L_1$ regularization.
While $L_p$ norms are traditionally defined for powers larger than one, $p\ge1$, the concept of applying powers smaller than one, $0<p<1$, was introduced more than three decades ago in sparse regression of large systems \cite{frank93}. 
Notably, for powers smaller than one, the penalty term becomes non-convex and is no longer a norm in the classical sense.
For the special case of $p=0.5$, the penalty term uses the sum of the square roots of the vector components,
$|| \, \vec{\theta} \, ||_{0.5} = [\, \sum_{i=1}^{n_{\rm{para}}} \sqrt{|w_i|} \,]^2$,
and we can easily see that this construct no longer satisfies the triangle inequality.
$L_{0.5}$ regularization supplements the regression (\ref{regression_nonlinear}) with a penalty term that consists of this term, multiplied by a penalty parameter $\alpha$,
\beq
  L(\vec{\theta}; \ten{F}) 
= \frac{1}{n_{\rm{data}}}
  \sum_{i=1}^{n_{\rm{data}}} 
 || \, \ten{P} (\ten{F}_i, \vec{\theta}) - \hat{\ten{P}}_i  \, ||^2 
+ \alpha \, || \, \vec{\theta} \, ||_{0.5}^{0.5}
 \rightarrow \mbox{min}
 \qquad
 \mbox{with}
 \qquad
 || \, \vec{\theta} \, ||_{0.5}^{0.5} 
= \sum_{i=1}^{n_{\rm{para}}}  
  \sqrt{| \, w_i \, |}  \,.
\eeq
The advantages of $L_{0.5}$ regularization, or more generally of 
$L_{p}$ regularization with powers smaller than one, are
{\it{enhanced sparsity}} potentially leading to more parsimonious models; and
{\it{subset selection}} especially in high-dimensional datasets \cite{friedman12}.
Its disadvantages are its
{\it{computational complexity}} induced by its non-convex nature; and its
{\it{multiple local minima}} making subset selection complex and non-unique. 
Figure \ref{fig01} shows that, for the special cases of 
$L_{0.75}$, $L_{0.5}$, and $L_{0.25}$ regularization, the regularization term adopts an increasingly non-convex shape and promotes sparsity by more aggressively setting weights equal to zero. 
\clearpage
\noindent
{\bf{\sffamily{L$_{\bf{0}}$ regularization or subset selection.}}}  
$L_0$ regularization, is a form of subset selection that imposes a penalty on the number of non-zero parameters in a regression model. 
The origins of selecting a subset of relevant parameters date back to early efforts in regression modeling with the objective to discover parsimonious models with enhanced interpretation and prediction. However, formalizing this idea as an $L_0$ penalty method and connecting it rigorously to regularization has emerged more prominently with the advent of high-dimensional datasets \cite{frank93}.
The $L_0$ norm is commonly referred to as sparse norm and is not a norm in a strict mathematical sense. It refers to the pseudo-norm, 
$|| \, \vec{\theta} \, ||_0 = \sum_{i=1}^{n_{\rm{para}}} I (w_i \ne 0)$,
where $I(\circ)$ is the indicator function that is one if the condition inside the parenthesis is true and zero otherwise. As such, the $L_0$ norm counts the number of non-zero entries in a vector, which implies that this approach directly penalizes model complexity in terms of predictor inclusion. Notably, the $L_0$ norm is an {\it{explicit}}, {\it{discrete}} measure of sparsity. It is robust to outliers since it only counts the number of non-zero elements in the parameter vector and does not express preference for smaller or larger entries.
$L_{0}$ regularization supplements the regression (\ref{regression_nonlinear}) with a penalty term that consists of the $L_0$ norm, multiplied by a penalty parameter $\alpha$,
\beq
  L(\vec{\theta}; \ten{F}) 
= \frac{1}{n_{\rm{data}}}
  \sum_{i=1}^{n_{\rm{data}}} 
 || \, \ten{P} (\ten{F}_i, \vec{\theta}) - \hat{\ten{P}}_i  \, ||^2 
+ \alpha \, || \, \vec{\theta} \, ||_{0}
 \rightarrow \mbox{min}
 \qquad
 \mbox{with}
 \qquad
 || \, \vec{\theta} \, ||_0 
= \sum_{i=1}^{n_{\rm{para}}} 
 I (w_i \ne 0)  \,.
\eeq
Its advantages are its
{\it{conceptual simplicity}} by providing a direct mechanism for subset selection that directly penalizes non-zero parameters; 
{\it{reduced overfitting}} by promoting fewer non-zero parameters, in particular when data are limited; and 
{\it{enhanced model interpretability}} by focusing only on the relevant terms.
Its disadvantages are its
{\it{computational complexity}} that results from turning continuous model selection into an {\it{NP hard discrete combinatorial problem}} with $2^n$ possible parameter combinations, making it computationally intractable for problems with large parameter sets;  its 
{\it{non-convexity}} induced by the $L_0$ penalty term that leads to optimization challenges related to several local minima; and increased
{\it{instability}} by discovering models for which slight changes in the data can result in an entirely different parameter set. 
In the contour plot of Figure \ref{fig01}, $L_{0}$ regularization would correspond to two discrete planes along the two parameter axes. \\[6.pt]
\noindent
{\bf{\sffamily{Predictability and interpretability.}}}  
$L_p$ regularization is an intricate balance between predictability and interpretability:
For powers larger than one, $p>1$, $L_p$ regularization can improve {\it{predictability}}, increase robustness, prevent overfitting, and enhance generalization to new data by penalizing outliers and reducing extreme coefficients.
For powers equal to or smaller than one, $p\le1$, $L_p$ regularization, can improve {\it{interpretability}}, promote simpler models, and identify the most influential predictors by encouraging sparsity and forcing some coefficients exactly to zero.
Taken together, $L_p$ regularization is a trade-off between interpretability and predictability, between simplicity and accuracy, and between bias and variance.
Two hyperparameters, the power $p$ and the regularization strength $\alpha$, allow us to fine-tuning of this balance. Throughout this manuscript, we will provide a library of systematic examples that  illustrate the sensitivity of $L_p$ regularization with respect to these two hyperparameters.
\section{Neural networks} 
\label{neural_networks}
In this study, we adopt the concept of neural networks to perform regression in constitutive modeling with the objective to improve both {\it{predictability}} and {\it{interpretabilty}}. 
To demonstrate that our strategy generalizes to different types of neural networks, we compare two constitutive neural networks that have recently become popular in the context of automated model discovery \cite{linka23a,stpierre23}. 
Both networks are {\it{sparse neural networks}} by design, where sparsity is inspired by the underlying physics of hyperelasticity. In their input layer, they use characteristic features of the deformation gradient $\ten{F}$ to a priori satisfy the kinematic constraint of {\it{material objectivity}}, and acknowledge a characteristic isotropic and incompressible material behavior by satisfying the constraints of {\it{material symmetry}} and {\it{incompressibility}} \cite{linka21}. In their output layer, they learn a free energy function $\psi$ from which they derive the nominal stress $\ten{P}$ to a priori satisfy the dissipation inequality and, with it, {\it{thermodynamic consistency}} \cite{masi21}. In their hidden layers, both networks use a special set of custom-designed activation functions to a priori satisfy {\it{physical constraints}} \cite{asad22} and a particular network architecture to satisfy {\it{polyconvexity}} \cite{klein22,tac22}. \\[6.pt]
\noindent
{\bf{\sffamily{Invariant based and principal stretch based neural networks.}}}  
We explore two types of neural networks that use different types of activation functions to represent two different classes of constitutive models: invariant and principal stretch based \cite{holzapfel00book,ogden72}. 
Both networks are {\it{generalizations}} of popular constitutive models that include widely used hyperelastic models as special cases \cite{blatz62,mooney40,rivlin48,treloar48}. They are {\it{interpretable}} by design and their weights translate into physically meaningful parameters with physical units and a physical interpretation \cite{linka23}. 
Yet, there is a major difference between both models: 
The invariant model uses {\it{different functional forms}} for each activation function
and results in a {\it{nonlinear regression problem}},
whereas the principal stretch model uses the {\it{same functional form with different but fixed exponents}} and results in a {\it{linear regression problem}}. 
This has critical implications on the convexity of the objective function, and with it, on the nature of the solution. \\[6.pt]
\noindent
{\bf{\sffamily{Data.}}}  
We train our networks on both synthetic and real data from tension, compression, and shear tests. 
For the {\it{synthetic data}}, 
we generate stretch stress pairs 
for fixed parameters through forward simulation. 
Specifically, we calculate stresses
over a wide range of 
tensile stretches, 
compressive stretches, and 
shear strains
in ten equidistant increments,
resulting in three data sets
with eleven stretch-stress pairs each. 
For the {\it{real data}}, 
we extract stretch stress pairs 
from our previously published 
human brain experiments on 
5\,mm gray matter tissue cubes \cite{budday17,budday20}.
Specifically, we use the reported stresses
averaged over
$n=15$ tensile, 
$n=17$ compressive, and
$n=35$ shear experiments, 
in sixteen equidistant increments,
resulting in three data sets
with seventeen stretch-stress pairs each \cite{linka23a}. 
All data are available on our GitHub repository,
https://github.com/LivingMatterLab/CANN.
\subsection{Invariant based neural network}
\label{inv_network}
\begin{figure}[t]
\centering
\includegraphics[width=0.72\linewidth]{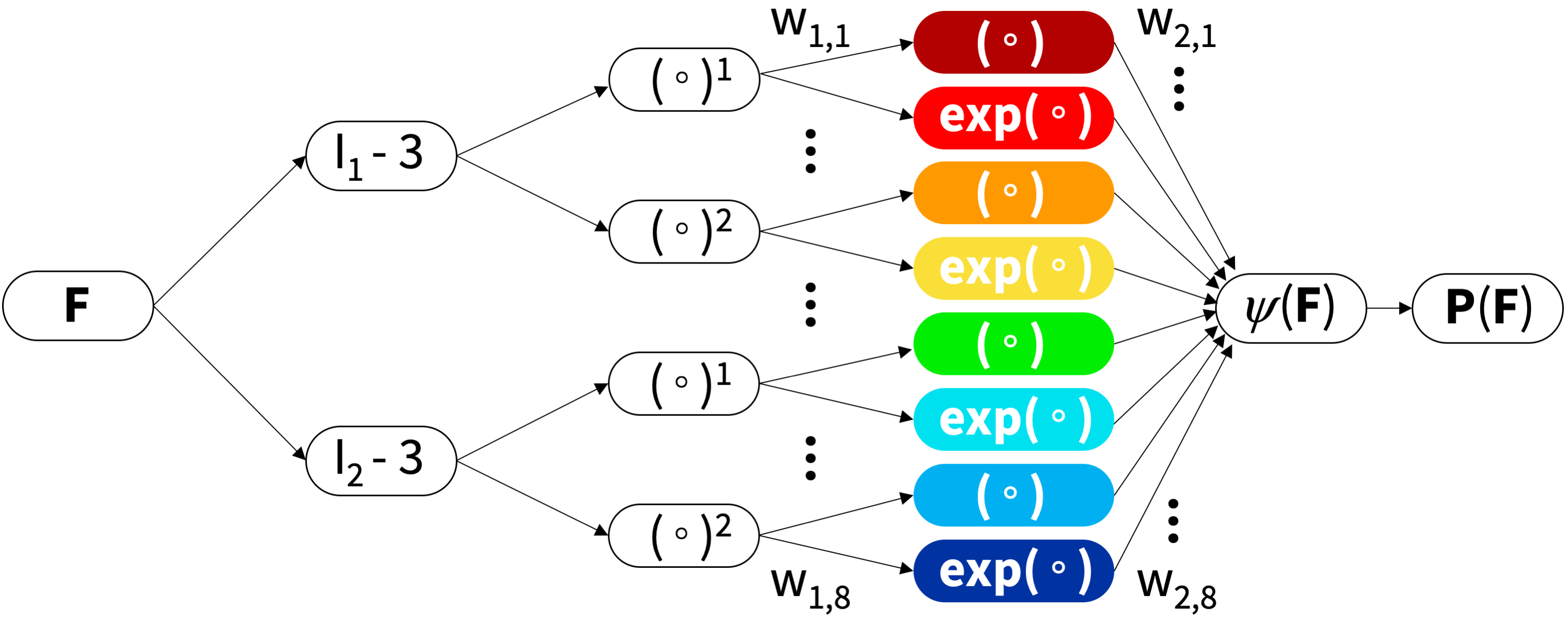}
\caption{{\bf{\sffamily{Invariant based neural network for automated model discovery.}}} 
The network takes the deformation gradient $\ten{F}$ as input and outputs the free energy function $\psi$ from which we calculate the stress 
$\ten{P} = \partial \psi / \partial \ten{F}$.
The network is invariant based, it first calculates the invariants
$I_1$ and $I_2$, and feeds them into its two hidden layers.
The first layer generates the first and second powers 
$(\circ)$ and $(\circ)^2$ of the invariants 
and the second layer applies the identity and exponential function 
$(\circ)$ and $\exp(\circ)$ to these powers.
The free energy function $\psi$ is a function of the eight color-coded terms.
During training, the network discovers the best model, of $2^8=256$ possible combinations of terms, to explain the experimental data $\hat{\ten{P}}$.}
\label{fig02}
\end{figure}

Invariant based constitutive neural networks take the deformation gradient $\ten{F}$ as input and predict the free energy function $\psi$ as output from which we calculate the stress $\ten{P} = \partial \psi / \partial \ten{F}$.
From the deformation gradient, they extract a set of invariants, 
in our example $I_1$ and $I_2$, 
and feed them into its two hidden layers \cite{linka23a,linka23b}. 
The first layer generates the powers of the invariants,
in our example the first and second $(\circ)$ and $(\circ)^2$,
and the second layer subjects these powers to specific functions,
in our example to the identity and exponential $(\circ)$ and $\exp(\circ)$.
The free energy function $\psi$ is a sum of the resulting eight terms.
Figure \ref{fig02} illustrates the invariant based constitutive neural network with the eight functional building blocks highlighted in color, where the hot red colors relate to the first invariant and the cold blue colors to the second. 
\\[4.pt]
During training, 
the network {\it{autonomously}} discovers the best model, 
out of $2^8=256$ possible combinations of terms,
and {\it{simultaneously}} learns its model parameters
$\vec{\theta} = \{\, w_{i,j} \,\}$.
It minimizes the loss function (\ref{regression_NN}),
the difference between the stress predicted by the model $\ten{P}(\ten{F}_i,\vec{\theta})$ and the experimentally measured stress $\hat{\ten{P}}_i$, divided by the number of data points used for training $n_{\rm{data}}$,
\beq
  L (\vec{\theta} ; \ten{F})
= \frac{1}{n_{\rm{data}}} 
  \sum_{i=1}^{n_{\rm{data}}}
|| \, \ten{P}(\ten{F}_i, \vec{\theta}) - \hat{\ten{P}}_i \, ||^2 
\rightarrow \mbox{min}\,.
\label{inv_loss}
\eeq
To ensure thermodynamic consistency, the network does not learn the stress directly \cite{ghaboussi91}, but rather derives it from the free energy function \cite{linka21}.
For our example in Figure \ref{fig02}, free energy function takes the following explicit representation,
\beq
\begin{array}{l@{\hspace*{0.1cm}}c@{\hspace*{0.1cm}}l@{\hspace*{0.1cm}}
              l@{\hspace*{0.1cm}}c@{\hspace*{0.1cm}}
              l@{\hspace*{0.1cm}}l@{\hspace*{0.1cm}}l@{\hspace*{0.04cm}}
              l@{\hspace*{0.1cm}}c@{\hspace*{0.1cm}}
              l@{\hspace*{0.1cm}}l@{\hspace*{0.1cm}}l@{\hspace*{0.0cm}}l}
    \psi(I_1,I_2)
&=& w_{2,1}w_{1,1} &[\,I_1 - 3\,]
&+& w_{2,2} & [ \, \exp (\,   w_{1,2} & [\, I_1 -3 \,]&)   - 1\,] \\
&+& w_{2,3}w_{1,3} &[\,I_1 - 3\,]^2
&+& w_{2,4} & [ \, \exp (\,   w_{1,4} & [\, I_1 -3 \,]^2&) - 1\,] \\
&+& w_{2,5}w_{1,5} &[\,I_2 - 3\,]
&+& w_{2,6} & [ \, \exp (\,   w_{1,6} & [\, I_2 -3 \,]&)   - 1\,] \\
&+& w_{2,7}w_{1,7} &[\,I_2 - 3\,]^2
&+& w_{2,8} & [ \, \exp (\,   w_{1,8} & [\, I_2 -3 \,]^2&)- 1\,] \,.
\label{inv_energy}
\end{array}
\eeq
with the following derivatives with respect to the invariants $I_1$ and $I_2$,
\beq
\begin{array}{ @{\hspace*{0.0cm}}
              l@{\hspace*{0.1cm}}l@{\hspace*{0.1cm}}r@{\hspace*{0.1cm}}
              l@{\hspace*{0.1cm}}l@{\hspace*{0.1cm}}l@{\hspace*{0.1cm}}
              l@{\hspace*{0.1cm}}l@{\hspace*{0.1cm}}l@{\hspace*{0.1cm}}
              l@{\hspace*{0.1cm}}l@{\hspace*{0.1cm}}l@{\hspace*{0.1cm}}
              l@{\hspace*{0.1cm}}l@{\hspace*{0.1cm}}l@{\hspace*{0.1cm}}
              l@{\hspace*{0.1cm}}c@{\hspace*{0.1cm}}
              l@{\hspace*{0.1cm}}l@{\hspace*{0.1cm}}l@{\hspace*{0.1cm}}
              l@{\hspace*{0.1cm}}l}
   \D{\frac{\partial \psi}{\partial I_1}}
&=& w_{2,1} w_{1,1}  
&+& w_{2,2} w_{1,2} \exp (\,   w_{1,2} [ I_1 -3 ]) 
&+&2\,[I_1 - 3][ w_{2,3}w_{1,3} 
&+& w_{2,4} w_{1,4} \exp (\,   w_{1,4} [ I_1 -3 ]^2)]\\[8.pt]
   \D{\frac{\partial \psi}{\partial I_2}}
&=& w_{2,5} w_{1,5}  
&+& w_{2,6} w_{1,6} \exp (\,   w_{1,6} [ I_2 -3 ]) 
&+&2\,[I_2 - 3][ w_{2,7}w_{1,7} 
&+& w_{2,8} w_{1,8} \exp (\,   w_{1,8} [ I_2 -3 ]^2)]
\label{inv_deriv}
\end{array}
\eeq
Using the second law of thermodynamics, we can derive the Piola stress,
$\ten{P} = {\partial \psi}/{\partial \ten{F}}$, 
as thermodynamically conjugate to the deformation gradient $\ten{F}$ \cite{holzapfel00book},
\beq
  \ten{P} 
= \frac{\partial \psi}{\partial I_1} 
  \cdot 
  \frac{\partial I_1}{\partial \ten{F}}
+ \frac{\partial \psi}{\partial I_2} 
  \cdot 
  \frac{\partial I_2}{\partial \ten{F}}
- p \, \ten{F}^{-{\rm{t}}} \,,
  \label{inv_stress01}
\eeq  
where the term $p \, \ten{F}^{-{\rm{t}}}$ ensures perfect incompressibility in terms of the pressure $p$ that we determine from the boundary conditions.   
For the network free energy (\ref{inv_energy}), the Piola stress is
\beq
\begin{array}{ @{\hspace*{0.0cm}}
              l@{\hspace*{0.1cm}}l@{\hspace*{0.1cm}}l@{\hspace*{0.0cm}}
              l@{\hspace*{0.1cm}}l@{\hspace*{0.1cm}}l@{\hspace*{0.0cm}}
              l@{\hspace*{0.0cm}}l@{\hspace*{0.1cm}}l@{\hspace*{0.1cm}}
              l@{\hspace*{0.0cm}}l@{\hspace*{0.1cm}}l@{\hspace*{0.1cm}}
              l@{\hspace*{0.0cm}}l@{\hspace*{0.1cm}}l@{\hspace*{0.1cm}}
              l@{\hspace*{0.1cm}}c@{\hspace*{0.1cm}}
              l@{\hspace*{0.1cm}}l@{\hspace*{0.1cm}}l@{\hspace*{0.0cm}}
              l@{\hspace*{0.1cm}}l}
   \ten{P}
&=&[
  & w_{2,1}   w_{1,1}   &
&+& w_{2,2} & w_{1,2} & \exp (\,   w_{1,2} & [\, I_1 -3 \,]&) \\
&+&2\,[\,I_1 - 3\,][& w_{2,3}w_{1,3} &
&+& w_{2,4} & w_{1,4} & \exp (\,   w_{1,4} & [\, I_1 -3 \,]^2&)]
&\D{\partial I_1}/{\partial \ten{F}}\\
&+&[
  & w_{2,5}   w_{1,5}   &
&+& w_{2,6} & w_{1,6} & \exp (\,   w_{1,6} & [\, I_2 -3 \,]&) \\
&+&2\,[\,I_2 - 3\,][& w_{2,7}w_{1,7} &
&+& w_{2,8} & w_{1,8}& \exp (\,   w_{1,8} & [\, I_2 -3 \,]^2&)]
&\D{{\partial I_2}/{\partial \ten{F}}
    - p \, \ten{F}^{-{\rm{t}}}} \,.
\label{inv_stress02}
\end{array}
\eeq
Notably, the Piola stress of the invariant based network (\ref{inv_stress02}) is a nonlinear function in the network weights $w_{i,j}$, which translates the loss function (\ref{inv_loss}) into a {\it{nonlinear regression}} problem, with possibly multiple local minima.
For this particular format, one of the first two weights of each row becomes redundant, and we can reduce the set of network parameters to twelve, 
$ \vec{\theta} = 
[\,(w_{1,1}w_{2,1}), w_{1,2}, w_{2,2}, (w_{1,3}w_{2,3}), w_{1,4}, w_{2,4} 
   (w_{1,5}w_{2,5}), w_{1,6}, w_{2,6}, (w_{1,7}w_{2,7}), w_{1,8}, w_{2,8} \,]$.
We train our invariant based network with tension, compression, and shear data 
and rewrite the loss function (\ref{inv_loss}) in terms of two contributions
that minimize the error 
between the normal and shear stresses predicted by the model, 
${P}_{11}(\lambda_i)$ and ${P}_{12}(\gamma_i)$,
and the data, 
$\hat{{P}}_{11,i}$ and $\hat{{P}}_{12,i}$,
where 
$n_{11}$ and $n_{12}$
denote the different stretch and shear levels 
$\lambda$ and $\gamma$,
\beq
  L (\vec{\theta} ; \lambda, \gamma)
= \frac{1}{n_{11}} \sum_{i=1}^{n_{11}}
|| \, {P}_{11}(\lambda_i) - \hat{{P}}_{11,i} \, ||^2 
+ \frac{1}{n_{12}} \sum_{i=1}^{n_{12}}
|| \, {P}_{12}(\gamma_i)  - \hat{{P}}_{12,i} \, ||^2 
\rightarrow \mbox{min}\,.
\label{inv_loss_P11_P12}
\eeq
To explore the effect of {\it{scaling}} of the three individual stress terms,
alternatively, we weigh all three experiments equally, 
and also train the network by minimizing the error 
between the normalized tensile, compressive, and shear stresses predicted by the model 
${P}_{\rm{ten}}(\lambda_i)$, 
${P}_{\rm{com}}(\lambda_i)$, and 
${P}_{\rm{shr}}(\gamma_i)$,
and the data
$\hat{{P}}_{{\rm{ten}},i}$,
$\hat{{P}}_{{\rm{com}},i}$, and
$\hat{{P}}_{{\rm{shr}},i}$, 
normalized by the maximum recorded tensile, compressive, and shear stresses,
$\hat{{P}}_{\rm{ten}}^{\rm{max}}$,
$\hat{{P}}_{\rm{com}}^{\rm{min}}$, and
$\hat{{P}}_{\rm{shr}}^{\rm{max}}$, 
where 
$n_{\rm{ten}}$,
$n_{\rm{com}}$, and
$n_{\rm{shr}}$ denote the different stretch and shear levels
$\lambda$ and $\gamma$,
\beq
  L (\vec{\theta} ; \lambda, \gamma)
= \frac{1}{n_{\rm{ten}}} \sum_{i=1}^{n_{\rm{ten}}}\!
  \left|\left| \, 
  \frac{{P}_{\rm{ten}}(\lambda_i) - \hat{{P}}_{{\rm{ten}},i}}
       {\hat{{P}}_{\rm{ten}}^{\rm{max}}}  \right|\right|^2 \!\!
+ \frac{1}{n_{\rm{com}}} \sum_{i=1}^{n_{\rm{com}}}\!
  \left|\left| \, 
  \frac{{P}_{\rm{com}}(\lambda_i) - \hat{{P}}_{{\rm{com}},i}}
       {\hat{{P}}_{\rm{com}}^{\rm{min}}}  \right|\right|^2 \!\!
+ \frac{1}{n_{\rm{shr}}} \sum_{i=1}^{n_{\rm{shr}}}\!
  \left|\left| \, 
  \frac{{P}_{\rm{shr}}(\gamma_i)  - \hat{{P}}_{{\rm{shr}},i}}
       {\hat{{P}}_{\rm{shr}}^{\rm{max}}}  \right|\right|^2 \!\!
  \rightarrow \mbox{min}\,.
\label{inv_loss_Pten_Pcom_Pshr}
\eeq
Below, we briefly derive the explicit analytical expressions for 
the Piola stresses
${P}_{11}(\lambda)$ in uniaxial tension and compression and 
${P}_{12}(\gamma)$  in simple shear, 
such that the tensile stress is 
$P_{\rm{ten}} = P_{11}$ for $\lambda >1$, 
the compressive stress is
$P_{\rm{com}} = P_{11}$ for $\lambda <1$, 
and the shear stress is
$P_{\rm{shr}} = P_{12}$ for all $\gamma$.\\[6.pt]   
\noindent{\bf{\sffamily{Uniaxial tension and compression.}}} 
For the special case of uniaxial tension and compression 
in terms of the stretch $\lambda$, with
$\lambda_1 = \lambda$ and 
$\lambda_2 = \lambda^{-1/2}$ and
$\lambda_3 = \lambda^{-1/2}$,
the invariants take the following form,
\beq
  I_1 =  \lambda^2 + \frac{2}{\lambda}
  \quad \mbox{and} \quad
  I_2 = 2\lambda  + \frac{1}{\lambda^2}
  \quad \mbox{and} \quad
  I_3 = 1  \,.
\eeq
Using equation (\ref{inv_stress01}) and the zero normal stress condition, 
$P_{22}=P_{33}=0$,
we obtain the following expression for the uniaxial stress stretch relation,
\beq
  P_{11} 
= 2 \, \left[ 
  \frac{\partial \psi}{\partial I_1}
+ \frac{1}{\lambda}
  \frac{\partial \psi}{\partial I_2}
  \right]
  \left[
  \lambda - \frac{1}{\lambda^2} 
  \right] \,,
\eeq
which translates into the following explicit expression between our network stress $P_{11}$ and the uniaxial stretch $\lambda$,
\beq
\begin{array}{ @{\hspace*{0.00cm}}l@{\hspace*{0.05cm}}
              l@{\hspace*{0.05cm}}r@{\hspace*{0.05cm}}r@{\hspace*{0.05cm}}
              l@{\hspace*{0.05cm}}l@{\hspace*{0.05cm}}l@{\hspace*{0.05cm}}
              l@{\hspace*{0.05cm}}l@{\hspace*{0.05cm}}l@{\hspace*{0.05cm}}
              l@{\hspace*{0.05cm}}l@{\hspace*{0.05cm}}l@{\hspace*{0.05cm}}
              l@{\hspace*{0.05cm}}l@{\hspace*{0.05cm}}l@{\hspace*{0.05cm}}
              l@{\hspace*{0.05cm}}c@{\hspace*{0.05cm}}
              l@{\hspace*{0.05cm}}l@{\hspace*{0.05cm}}l@{\hspace*{0.05cm}}
              l@{\hspace*{0.05cm}}l}
   P_{11}
&=&2\,[
  &[w_{2,1} w_{1,1}  
&+& w_{2,2} w_{1,2} \exp (\,   w_{1,2} [ I_1 -3 ]) 
&+&2\,[I_1 - 3][ w_{2,3}w_{1,3} 
&+& w_{2,4} w_{1,4} \exp (\,   w_{1,4} [ I_1 -3 ]^2)]\\
&+&\D{\frac{1}{\lambda}} 
  &[w_{2,5} w_{1,5}  
&+& w_{2,6} w_{1,6} \exp (\,   w_{1,6} [ I_2 -3 ]) 
&+&2\,[I_2 - 3][ w_{2,7}w_{1,7} 
&+& w_{2,8} w_{1,8} \exp (\,   w_{1,8} [ I_2 -3 ]^2)]]
   \, \D{\left[\lambda - \frac{1}{\lambda^2}\right]} \,.
\label{P11_inv}
\end{array}
\eeq
\noindent{\bf{\sffamily{Simple shear.}}} 
For the special case of simple shear,
in terms of the shear $\gamma$,
with
$ \lambda_1 
= \frac{1}{2}\,[\, + \gamma + \sqrt{\,4+\gamma^2} \,]$ and
$ \lambda_2 
= \frac{1}{2}\,[\, - \gamma + \sqrt{\,4+\gamma^2} \,]$ and
$\lambda_3 
= 1$, 
the invariants take the following form, 
\beq
  I_1 = 3 + \gamma^2
  \quad \mbox{and} \quad
  I_2 = 3 + \gamma^2
  \quad \mbox{and} \quad
  I_3 = 1 \,.
\eeq
Using equation (\ref{inv_stress01}), we obtain the following shear stress stretch relation,
\beq  
  P_{12} 
= 2\, \left[ 
  \frac{\partial \psi}{\partial I_1}
+ \frac{\partial \psi}{\partial I_2}
  \right]
  \gamma \, ,
\eeq
which translates into the following explicit expression between our network shear stress $P_{12}$ and the shear strain $\gamma$,
\beq
\begin{array}{ @{\hspace*{0.0cm}}
              l@{\hspace*{0.1cm}}l@{\hspace*{0.1cm}}r@{\hspace*{0.1cm}}
              l@{\hspace*{0.1cm}}l@{\hspace*{0.1cm}}l@{\hspace*{0.1cm}}
              l@{\hspace*{0.1cm}}l@{\hspace*{0.1cm}}l@{\hspace*{0.1cm}}
              l@{\hspace*{0.1cm}}l@{\hspace*{0.1cm}}l@{\hspace*{0.1cm}}
              l@{\hspace*{0.1cm}}l@{\hspace*{0.1cm}}l@{\hspace*{0.1cm}}
              l@{\hspace*{0.1cm}}c@{\hspace*{0.1cm}}
              l@{\hspace*{0.1cm}}l@{\hspace*{0.1cm}}l@{\hspace*{0.1cm}}
              l@{\hspace*{0.1cm}}l}
   P_{12}
&=&2 \, [
    w_{2,1} w_{1,1}  
&+& w_{2,2} w_{1,2} \exp (\,   w_{1,2} [ I_1 -3 ]) 
&+&2\,[I_1 - 3][ w_{2,3}w_{1,3} 
&+& w_{2,4} w_{1,4} \exp (\,   w_{1,4} [ I_1 -3 ]^2)\\[2.pt]
&&+ w_{2,5} w_{1,5}  
&+& w_{2,6} w_{1,6} \exp (\,   w_{1,6} [ I_2 -3 ]) 
&+&2\,[I_2 - 3][ w_{2,7}w_{1,7} 
&+& w_{2,8} w_{1,8} \exp (\,   w_{1,8} [ I_2 -3 ]^2)] \,\gamma \,.
\label{P12_inv}
\end{array}
\eeq
\begin{figure}[t]
\centering
\includegraphics[width=0.84\linewidth]{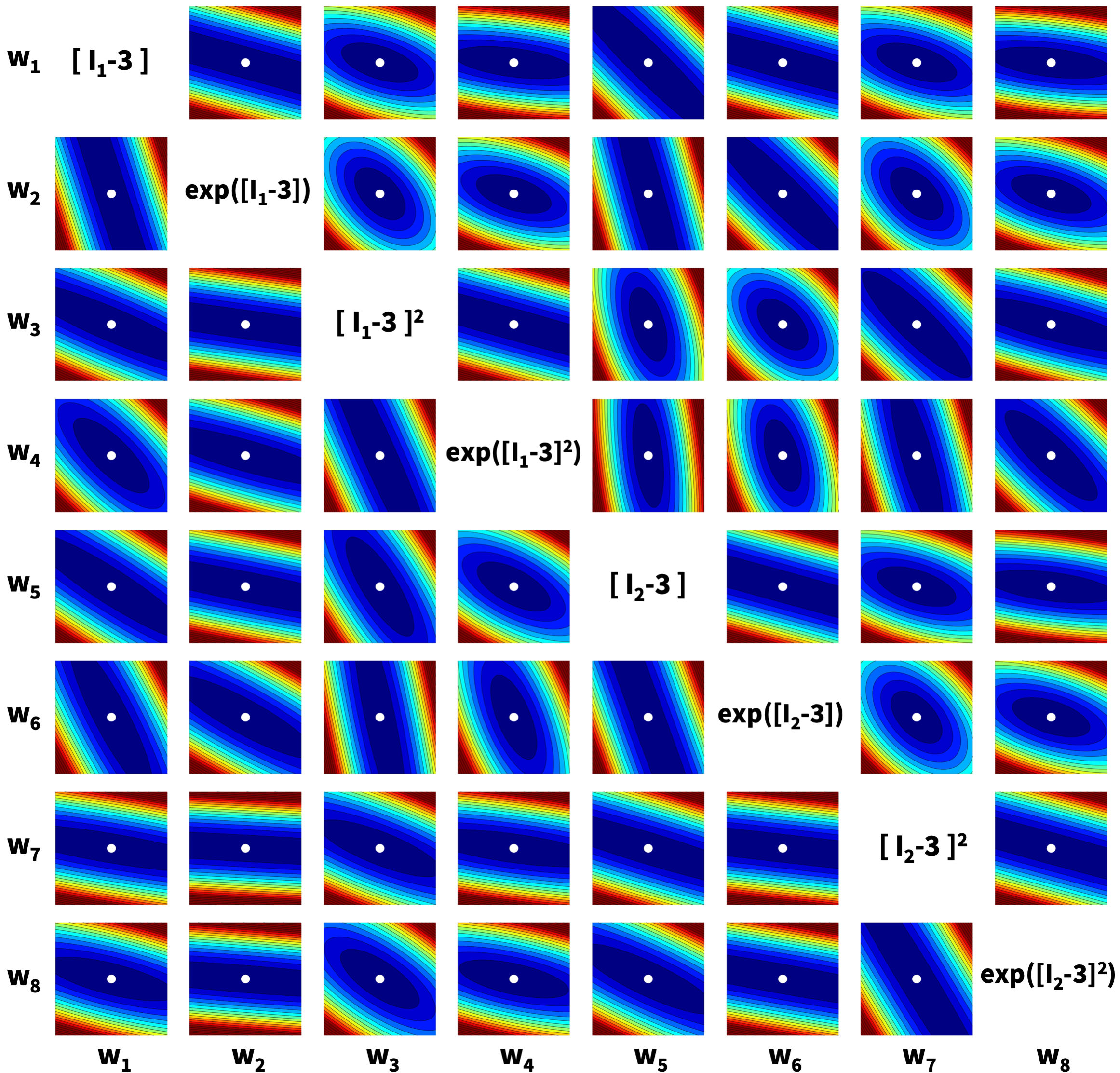}
\caption{{\bf{\sffamily{Loss functions of invariant based neural network.}}}
Contours of the loss function $L(\vec{\theta};\lambda,\gamma)$
for all 28 possible two-term models of the invariant based constitutive neural network in Figure \ref{fig02}.
The loss function is evaluated across  
tensile stretches $\lambda = [1.0,...,2.0]$,
compressive stretches $\lambda = [1.0,...,0.5]$, and
shear strains $\gamma = [0.0,...,0.5]$,
with network weights in the ranges 
$w_i = [0,...,2]$ and $w_j = [0,...,2]$.
The minimum of the loss function 
indicates the exact solution 
$w_i = 1$ and $w_j = 1$, represented through the white circle.
The lower triangle illustrates the non-normalized loss function (\ref{inv_loss_P11_P12}),
the upper triangle illustrates the normalized loss function (\ref{inv_loss_Pten_Pcom_Pshr}).
All loss functions are convex, with contours varying from 
ellipsoids to valleys with long ridges,
highlighting the collinearity of some $w_i$ and $w_j$ pairs.}
\label{fig03}
\end{figure}
Figure \ref{fig03} illustrates the contours of the loss function
$L(\vec{\theta};\lambda,\gamma)$
for all possible two-term models of the invariant based network in Figure \ref{fig02}. By combining any two terms of the model and setting all other weights equal to zero, we can generate 28 possible models. 
For these 28 combinations of two terms, 
we evaluate two versions of the loss function, 
non-normalized from equation (\ref{inv_loss_P11_P12}) 
and normalized from equation (\ref{inv_loss_Pten_Pcom_Pshr}),
using the invariant based definitions of the normal stress (\ref{P11_inv}) and shear stress (\ref{P12_inv}).
First, we generate synthetic data, 
$\hat{P}_{11,i}$ and $\hat{P}_{12,i}$, for   
tensile stretches of $\lambda = [1.0,...,2.0]$,
compressive stretches of $\lambda = [1.0,...,0.5]$, and
shear strains of $\gamma = [0.0,...,0.5]$,
in ten equidistant increments each,
assuming an exact solution
with fixed weights of the first layer,
$w_{1,1}=w_{1,3}=w_{1,5}=w_{1,7}=1.0$ and
$w_{1,2}=w_{1,4}=w_{1,6}=w_{1,8}=0.25$, 
and a pair of non-zero weights of the second layer,
$w_{2,i} = 1$ and $w_{2,j} = 1$, 
while fixing the remaining six weights of the second layer equal to zero.
This results in 28 training data sets of eleven
stretch-stress pairs each, for tension, compression, and shear. 
Second, we vary the two non-zero network weights in the ranges
$w_i = [0,..,2]$ and $w_j = [0,..,2]$. 
For each pair of weights,
we evaluate 
the normal and shear stresses
${P}_{11}(\lambda_i)$ and ${P}_{12}(\gamma_i)$
using equations
(\ref{P11_inv}) and (\ref{P12_inv}),
and extract the tensile, compressive, and shear stresses,
${P}_{\rm{ten_i}}(\lambda)$, ${P}_{\rm{com_i}}(\lambda)$, and ${P}_{\rm{shr_i}}(\gamma)$.
Third, 
we evaluate the {\it{non-normalized}} loss function (\ref{inv_loss_P11_P12}) 
as the mean squared error between the model stresses 
${P}_{11}(\lambda_i)$ and ${P}_{12}(\gamma_i)$
and the synthetically generated data stresses
$\hat{P}_{11,i}$ and $\hat{P}_{12,i}$,
and plot its contours for each pair of weights $w_i$ and $w_j$ in the lower triangle. 
Next, 
we evaluate the {\it{normalized}} loss function (\ref{inv_loss_Pten_Pcom_Pshr}) 
as the normalized mean squared error between the model stresses 
${P}_{\rm{ten}}(\lambda_i)$, ${P}_{\rm{com}}(\lambda_i)$, and ${P}_{\rm{shr}}(\gamma_i)$,
and the synthetically generated data stresses
$\hat{P}_{{\rm{ten}},i}$, $\hat{P}_{{\rm{com}},i}$ and $\hat{P}_{{\rm{shr}},i}$,
and plot its contours for each pair of weights $w_i$ and $w_j$ in the upper triangle. \\[6.pt]
Each loss function takes a minimum of 
$L(\vec{\theta};\lambda,\gamma) = 0$
for the {\it{exact solution}}, $w_i = 1$ and $w_j = 1$, 
indicated through the white circles. 
From this minimum, 
both versions of the loss function, non-normalized and normalized,
increase with both, decreasing and increasing weights $w_i$ and $w_j$, 
and remain {\it{convex}} within the entire domain, for all 28 two-term models.
Yet, the contours of the loss function vary significantly
for different pairs of weights $w_i$ and $w_j$,
indicating its sensitivity with respect to the individual terms:
Some pairs of weights generate loss functions of ellipsoidal shape, 
e.g., the 
$\{\,w_2, w_3\,\}$, $\{\,w_2, w_7\,\}$,
$\{\,w_3, w_6\,\}$, $\{\,w_6, w_7\,\}$ pairs
in the normalized upper triangle,
suggesting that in the studied stretch and shear range,  
these terms are {\it{non-collinear}},
and would represent a rich base for a potential constitutive model.
Other pairs of weights generate loss functions with long ridges parallel to the parameter axes, e.g., the 
$\{\,w_2, w_7\,\}$, $\{\,w_2, w_8\,\}$,
$\{\,w_6, w_7\,\}$, $\{\,w_6, w_8\,\}$ pairs
in the non-normalized lower triangle,
suggesting that these terms are almost {\it{collinear}} 
and not well suited as an independent base for a constitutive model. 
On average, 
the normalized pairs in the upper triangle seem to generate more convex loss functions than the non-normalized pairs in the lower triangle, 
suggesting that normalization helps to generate 
more convex loss functions,
a richer functional base, and 
a more robust solution overall.  
Notably, 
the $\{\,w_1, w_5\,\}$ model in the first row and fifth column and 
the $\{\,w_5, w_1\,\}$ model in the fifth row and first column
combine the linear terms in the first and second invariants, 
$[\,I_1-3\,]$ and $[\,I_2-3\,]$, 
and represent the popular Mooney Rivlin model for rubber-like materials \cite{mooney40,rivlin48}.
Overall, this simple example only illustrates the 28 two-term models out of a total set of all 256 possible models, only considers a limited stretch and shear range $\lambda$ and $\gamma$, and only screens a narrow window of parameter ranges $w_i$. 
Even within these limitations, the contours of loss functions are rather difficult to interpret, making it difficult to comprehend the full potential of the entire network, even though it only consists of eight distinct terms.   
\subsection{Principal stretch based neural network}
\label{lam_network}
\begin{figure}[t]
\centering
\includegraphics[width=0.72\linewidth]{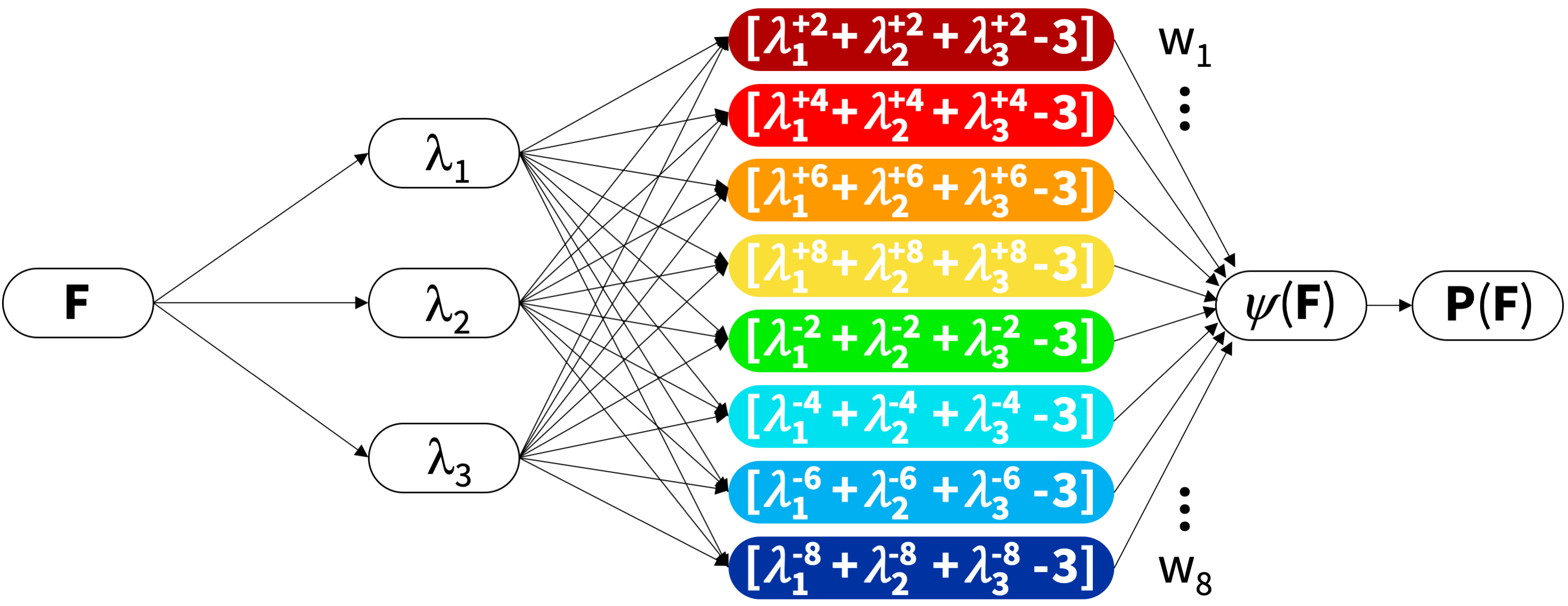}
\caption{{\bf{\sffamily{Principal stretch based neural network for automated model discovery.}}} 
The network takes the deformation gradient $\ten{F}$ as input and outputs the free energy function $\psi$ from which we calculate the stress 
$\ten{P} = \partial \psi / \partial \ten{F}$.
The network is principal stretch based, it first calculates the principal stretches
$\lambda_1$ and $\lambda_2$ and $\lambda_3$, and feeds them into its hidden layer.
The hidden layer applies eight different powers
$(\lambda_1^n+\lambda_2^n+\lambda_3^n-3)$
to these principal stretches.
The free energy function $\psi$ is a function of the eight color-coded terms.
During training, the network discovers the best model, of $2^8=256$ possible combinations of terms, to explain the experimental data $\hat{\ten{P}}$.}
\label{fig04}
\end{figure}

Principal stretch based constitutive neural networks take the deformation gradient $\ten{F}$ as input and predict the free energy function $\psi$ as output from which we calculate the stress $\ten{P} = \partial \psi / \partial \ten{F}$.
From the deformation gradient, they extract the principal stretches, 
$\lambda_1$, $\lambda_2$, and $\lambda_3$, 
and feed them into the hidden layer \cite{stpierre23,stpierre23a}.
The hidden layer applies eight different exponents
$(\lambda_1^n+\lambda_2^n+\lambda_3^n-3)$
to these stretches.
The free energy function $\psi$ is a sum of the resulting eight terms.
Figure \ref{fig04} illustrates the principal stretch based constitutive neural network with the eight functional building blocks highlighted in color, where the 
dark red and green terms are identical to the dark red and green terms of the invariant based network in Figure \ref{fig02}, while the other six terms are different. 
\\[6.pt]
During training, 
the network {\it{autonomously}} discovers the best model, 
out of $2^8=256$ possible combinations of terms,
and {\it{simultaneously}} learns its model parameters
$\vec{\theta} = \{\, w_{i} \,\}$.
It minimizes the loss function (\ref{regression_NN}),
the difference between the stress predicted by the model $\ten{P}(\ten{F}_i,\vec{\theta})$ and the experimentally measured stress $\hat{\ten{P}}_i$, divided by the number of data points used for training $n_{\rm{data}}$,
\beq
  L (\vec{\theta} ; \ten{F})
= \frac{1}{n_{\rm{data}}} 
  \sum_{i=1}^{n_{\rm{data}}}
|| \, \ten{P}(\ten{F}_i, \vec{\theta}) - \hat{\ten{P}}_i \, ||^2 
\rightarrow \mbox{min}\,.
\label{lam_loss}
\eeq
The free energy of the principal stretch based model takes the following explicit representation \cite{ogden72,valanis67},
$ \psi(\lambda_1,\lambda_2,\lambda_3)
= \sum_{i=1}^{n_{\rm{term}}}
  w_i \, 
  [\, \lambda_1^{\alpha_i} + \lambda_2^{\alpha_i} + \lambda_3^{\alpha_i} - 3 \,]$,
where the individual weights
$w_i = \mu_i/\alpha_i$
correspond to the shear moduli $\mu_i$ divided by the exponents $\alpha_i$.
For the eight-term model in Figure \ref{fig04}, 
we fix these exponents to
$\alpha = [\, +2, +4, +6, +8, -2, -4, -6 -8 \,]$,
such that the free energy becomes a sum of the following $n_{\rm{term}}=8$ terms,
\beq
\begin{array}{ @{\hspace*{0.0cm}}
              l@{\hspace*{0.1cm}}l@{\hspace*{0.1cm}}
              r@{\hspace*{0.1cm}}l@{\hspace*{0.1cm}}
              l@{\hspace*{0.1cm}}l@{\hspace*{0.1cm}}
              l@{\hspace*{0.1cm}}l@{\hspace*{0.1cm}}
              l@{\hspace*{0.1cm}}l@{\hspace*{0.1cm}}
              l@{\hspace*{0.1cm}}l@{\hspace*{0.1cm}}
              l@{\hspace*{0.1cm}}l@{\hspace*{0.1cm}}l}
    \psi(\lambda_1,\lambda_2,\lambda_3)
&=& \sum_{i=1}^3
  & w_{1} & [\,\lambda_i^{+2} - 1\,]
&+& w_{2} & [\,\lambda_i^{+4} - 1\,]
&+& w_{3} & [\,\lambda_i^{+6} - 1\,]
&+& w_{4} & [\,\lambda_i^{+8} - 1\,] \\
&&+&w_{5} & [\,\lambda_i^{-2} - 1\,]
&+& w_{6} & [\,\lambda_i^{-4} - 1\,]
&+& w_{7} & [\,\lambda_i^{-6} - 1\,]
&+& w_{8} & [\,\lambda_i^{-8} - 1\,]  \,.
\label{lam_energy}
\end{array}
\eeq
and its derivatives with respect to the principal stretches $\lambda_i$ takes the following form,
\beq
  \frac{\partial \psi}{\partial \lambda_i}
= 2\, w_{1} \lambda_i^{+1} 
+ 4\, w_{2} \lambda_i^{+3} 
+ 6\, w_{3} \lambda_i^{+5} 
+ 8\, w_{4} \lambda_i^{+7}
- 2\, w_{5} \lambda_i^{-3} 
- 4\, w_{6} \lambda_i^{-5} 
- 6\, w_{7} \lambda_i^{-7} 
- 8\, w_{8} \lambda_i^{-9} \,.
\eeq
Using the second law of thermodynamics, we can derive the Piola stress,
$\ten{P} = {\partial \psi}/{\partial \ten{F}}$, 
as thermodynamically conjugate to the deformation gradient $\ten{F}$ \cite{ogden72},
\beq
  \ten{P} 
= \sum_{i=1}^3
  \frac{\partial \psi}{\lambda_i} \vec{n}_i \otimes \vec{N}_i 
- p \, \ten{F}^{-{\rm{t}}}
\label{lam_stress01}
\eeq        
where  
$\vec{N}_i$ and $\vec{n}_i = \ten{F} \cdot \vec{N}_i$ are the eigenvectors in the undeformed and deformed configurations, 
and the term $p \, \ten{F}^{-{\rm{t}}}$ ensures perfect incompressibilty in terms of the pressure $p$ that we determine from the boundary conditions.   
For the network free energy (\ref{lam_energy}), the Piola stress is
\beq
\begin{array}{ @{\hspace*{0.0cm}} 
              l@{\hspace*{0.1cm}}l@{\hspace*{0.1cm}}
              r@{\hspace*{0.1cm}}l@{\hspace*{0.1cm}}
              l@{\hspace*{0.1cm}}l@{\hspace*{0.1cm}}
              l@{\hspace*{0.1cm}}l@{\hspace*{0.1cm}}
              l@{\hspace*{0.1cm}}l@{\hspace*{0.1cm}}
              l@{\hspace*{0.1cm}}l@{\hspace*{0.1cm}}
              l@{\hspace*{0.1cm}}l@{\hspace*{0.1cm}}
              l@{\hspace*{0.1cm}}l@{\hspace*{0.1cm}}
              l@{\hspace*{0.1cm}}l@{\hspace*{0.1cm}}l}
    \ten{P}
&=& \sum_{i=1}^3 \,[
  & 2 &w_{1} & [\,\lambda_i^{+1} - 1\,]
&+& 4 &w_{2} & [\,\lambda_i^{+3} - 1\,]
&+& 6 &w_{3} & [\,\lambda_i^{+5} - 1\,]
&+& 8 &w_{4} & [\,\lambda_i^{+7} - 1\,] \\
&&-&2 &w_{5} & [\,\lambda_i^{-3} - 1\,]
&-& 4 &w_{6} & [\,\lambda_i^{-5} - 1\,]
&-& 6 &w_{7} & [\,\lambda_i^{-7} - 1\,]
&-& 8 &w_{8} & [\,\lambda_i^{-9} - 1\,] \,] \,
   \vec{n}_i \otimes \vec{N}_i - p \, \ten{F}^{-{\rm{t}}}\,,
\label{lam_stress02}
\end{array}
\eeq
parameterized in terms of eight network weights,
$\vec{\theta} = [\,w_1, w_2, w_3, w_4, w_5, w_6, w_7, w_8 \,]$.
Notably, the Piola stress of the principal stretch based network (\ref{lam_stress02}) with fixed exponents is a linear function in the network weights $w_{i}$, which translates the loss function (\ref{lam_loss}) into a {\it{linear regression}} problem, with a single unique global minimum \cite{ogden04}.
We train our principal stretch based network 
with tension, compression, and shear data 
and rewrite the loss function (\ref{lam_loss}) in terms of two contributions
that minimize the error 
between the tensile and compressive stresses predicted by the model 
${P}_{11}(\lambda_i)$ and the data $\hat{{P}}_{11,i}$, and
between the shear stresses predicted by the model 
${P}_{12}(\gamma_i)$ and the data $\hat{{P}}_{12,i}$,
where we include data from 
$n_{11}$ different stretch levels $\lambda$ and
$n_{12}$ different shear levels $\gamma$,
\beq
  L (\vec{\theta} ; \lambda, \gamma)
= \frac{1}{n_{11}} \sum_{i=1}^{n_{11}}
|| \, {P}_{11}(\lambda_i) - \hat{{P}}_{11,i} \, ||^2 
+ \frac{1}{n_{12}} \sum_{i=1}^{n_{12}}
|| \, {P}_{12}(\gamma_i)  - \hat{{P}}_{12,i} \, ||^2 
\rightarrow \mbox{min}\,.
\label{lam_loss_P11_P12}
\eeq
For comparison, similar to the invariant based network in the Section \ref{inv_network}, 
we also train the network by minimizing the error 
between the normalized tensile, compressive, and shear stresses predicted by the model 
${P}_{\rm{ten}}(\lambda_i)$, 
${P}_{\rm{com}}(\lambda_i)$, and 
${P}_{\rm{shr}}(\gamma_i)$,
and the data
$\hat{{P}}_{{\rm{ten}},i}$,
$\hat{{P}}_{{\rm{com}},i}$, and
$\hat{{P}}_{{\rm{shr}},i}$, 
normalized by the maximum recorded tensile, compressive, and shear stresses,
$\hat{{P}}_{\rm{ten}}^{\rm{max}}$,
$\hat{{P}}_{\rm{com}}^{\rm{min}}$, and
$\hat{{P}}_{\rm{shr}}^{\rm{max}}$, 
where 
$n_{\rm{ten}}$,
$n_{\rm{com}}$, and
$n_{\rm{shr}}$ denote the different stretch and shear levels
$\lambda$ and $\gamma$,
\beq
  L (\vec{\theta} ; \lambda, \gamma)
= \frac{1}{n_{\rm{ten}}} \sum_{i=1}^{n_{\rm{ten}}}\!
  \left|\left| \, 
  \frac{{P}_{\rm{ten}}(\lambda_i) - \hat{{P}}_{{\rm{ten}},i}}
       {\hat{{P}}_{\rm{ten}}^{\rm{max}}}  \right|\right|^2 \!\!
+ \frac{1}{n_{\rm{com}}} \sum_{i=1}^{n_{\rm{com}}}\!
  \left|\left| \, 
  \frac{{P}_{\rm{com}}(\lambda_i) - \hat{{P}}_{{\rm{com}},i}}
       {\hat{{P}}_{\rm{com}}^{\rm{min}}}  \right|\right|^2 \!\!
+ \frac{1}{n_{\rm{shr}}} \sum_{i=1}^{n_{\rm{shr}}}\!
  \left|\left| \, 
  \frac{{P}_{\rm{shr}}(\gamma_i)  - \hat{{P}}_{{\rm{shr}},i}}
       {\hat{{P}}_{\rm{shr}}^{\rm{max}}}  \right|\right|^2 \!\!
  \rightarrow \mbox{min}\,.
\label{lam_loss_Pten_Pcom_Pshr}
\eeq
Below, we briefly derive the explicit analytical expressions for the
Piola stresses
${P}_{11}(\lambda)$ in uniaxial tension and compression and 
${P}_{12}(\gamma)$  in simple shear, 
such that the tensile stress is 
$P_{\rm{ten}} = P_{11}$ for $\lambda >1$, 
the compressive stress is
$P_{\rm{com}} = P_{11}$ for $\lambda <1$, 
and the shear stress is
$P_{\rm{shr}} = P_{12}$ for all $\gamma$.\\[6.pt]        
\noindent{\bf{\sffamily{Uniaxial tension and compression.}}} 
For the special case of uniaxial tension and compression in terms of the stretch $\lambda$, the principal stretches are
\beq
\lambda_1 = \lambda
\qquad \mbox{and} \qquad
\lambda_2 = \lambda^{-1/2}
\qquad \mbox{and} \qquad
\lambda_3 = \lambda^{-1/2} \,.
\eeq
Using equation (\ref{lam_stress01}) and the zero normal stress condition, 
$P_{22}=P_{33}=0$,
we obtain the following expression for the uniaxial stress stretch relation,
\beq
  P_{11}
= \frac{1}{\lambda} \, \sum_{i=1}^{n_{\rm{term}}}
  \alpha_i \, w_i \, [\, \lambda^{\alpha_i} - \lambda^{-\alpha_i/2} \,] \,,
\eeq
which translates into the following explicit expression between our network stress $P_{11}$ and the uniaxial stretch $\lambda$,
\beq
\begin{array}{ @{\hspace*{0.0cm}}
              l@{\hspace*{0.1cm}}l@{\hspace*{0.1cm}}
              r@{\hspace*{0.1cm}}l@{\hspace*{0.1cm}}
              l@{\hspace*{0.1cm}}l@{\hspace*{0.1cm}}
              l@{\hspace*{0.1cm}}l@{\hspace*{0.1cm}}
              l@{\hspace*{0.1cm}}l@{\hspace*{0.1cm}}
              l@{\hspace*{0.1cm}}l@{\hspace*{0.1cm}}
              l@{\hspace*{0.1cm}}l@{\hspace*{0.1cm}}l}
    P_{11}
&=& \D{\frac{1}{\lambda}} \,[\, 
  & 2\, w_{1} & [\,\lambda^{+2} - \lambda^{-1}\,]
&+& 4\, w_{2} & [\,\lambda^{+4} - \lambda^{-2}\,]
&+& 6\, w_{3} & [\,\lambda^{+6} - \lambda^{-3}\,]
&+& 8\, w_{4} & [\,\lambda^{+8} - \lambda^{-4}\,] \\
&&-&2\, w_{5} & [\,\lambda^{-2} - \lambda^{+1}\,]
&-& 4\, w_{6} & [\,\lambda^{-4} - \lambda^{+2}\,]
&-& 6\, w_{7} & [\,\lambda^{-6} - \lambda^{+3}\,]
&-& 8\, w_{8} & [\,\lambda^{-8} - \lambda^{+4}\,] \,] \,.
\label{P11_lam}
\end{array}
\eeq
\noindent{\bf{\sffamily{Simple shear.}}} 
For the special case of simple shear, in terms of the shear 
$\gamma$, we obtain the principal stretches,
\beq
  \lambda_1 
= \frac{\gamma + \sqrt{4+\gamma^2}}{2} 
  \quad \mbox{and} \quad
  \lambda_2 
= \frac{-\gamma + \sqrt{4+\gamma^2}}{2} 
= \frac{1}{\lambda_1}
  \quad \mbox{and} \quad
  \lambda_3 
= 1 \,.
\eeq
Using equation (\ref{lam_stress01}), we obtain the following expression for the shear stress stretch relation, 
\beq
  P_{12}
= \frac{1}{1+\lambda^2} \, \sum_{i=1}^{n_{\rm{term}}}
  \alpha_i \, w_i \, [\, \lambda^{\alpha_i+1} - \lambda^{1-\alpha_i} \,]
  \qquad
  \mbox{with}
  \qquad
  \lambda  
= \frac{1}{2} \, \left[\, \gamma + \sqrt{4+\gamma^2} \,\right]  
= \lambda_1 = \frac{1}{\lambda_2} \,,
\eeq
which translates into the following explicit expression between our network shear stress $P_{12}$ and shear strain $\gamma$,
\beq
\begin{array}{ @{\hspace*{0.0cm}}
              l@{\hspace*{0.1cm}}l@{\hspace*{0.1cm}}
              r@{\hspace*{0.1cm}}l@{\hspace*{0.1cm}}
              l@{\hspace*{0.1cm}}l@{\hspace*{0.1cm}}
              l@{\hspace*{0.1cm}}l@{\hspace*{0.1cm}}
              l@{\hspace*{0.1cm}}l@{\hspace*{0.1cm}}
              l@{\hspace*{0.1cm}}l@{\hspace*{0.1cm}}
              l@{\hspace*{0.1cm}}l@{\hspace*{0.1cm}}l}
    P_{12}
&=& \D{\frac{1}{1+\lambda^2}} \,[\, 
  & 2\, w_{1} & [\,\lambda^{+3} - \lambda^{-1}\,]
&+& 4\, w_{2} & [\,\lambda^{+5} - \lambda^{-3}\,]
&+& 6\, w_{3} & [\,\lambda^{+7} - \lambda^{-5}\,]
&+& 8\, w_{4} & [\,\lambda^{+9} - \lambda^{-7}\,] \\
&&-&2\, w_{5} & [\,\lambda^{-1} - \lambda^{+3}\,]
&-& 4\, w_{6} & [\,\lambda^{-3} - \lambda^{+5}\,]
&-& 6\, w_{7} & [\,\lambda^{-5} - \lambda^{+7}\,]
&-& 8\, w_{8} & [\,\lambda^{-7} - \lambda^{+9}\,] \,] \,.
\label{P12_lam}
\end{array}
\eeq
\begin{figure}[t]
\centering
\includegraphics[width=0.84\linewidth]{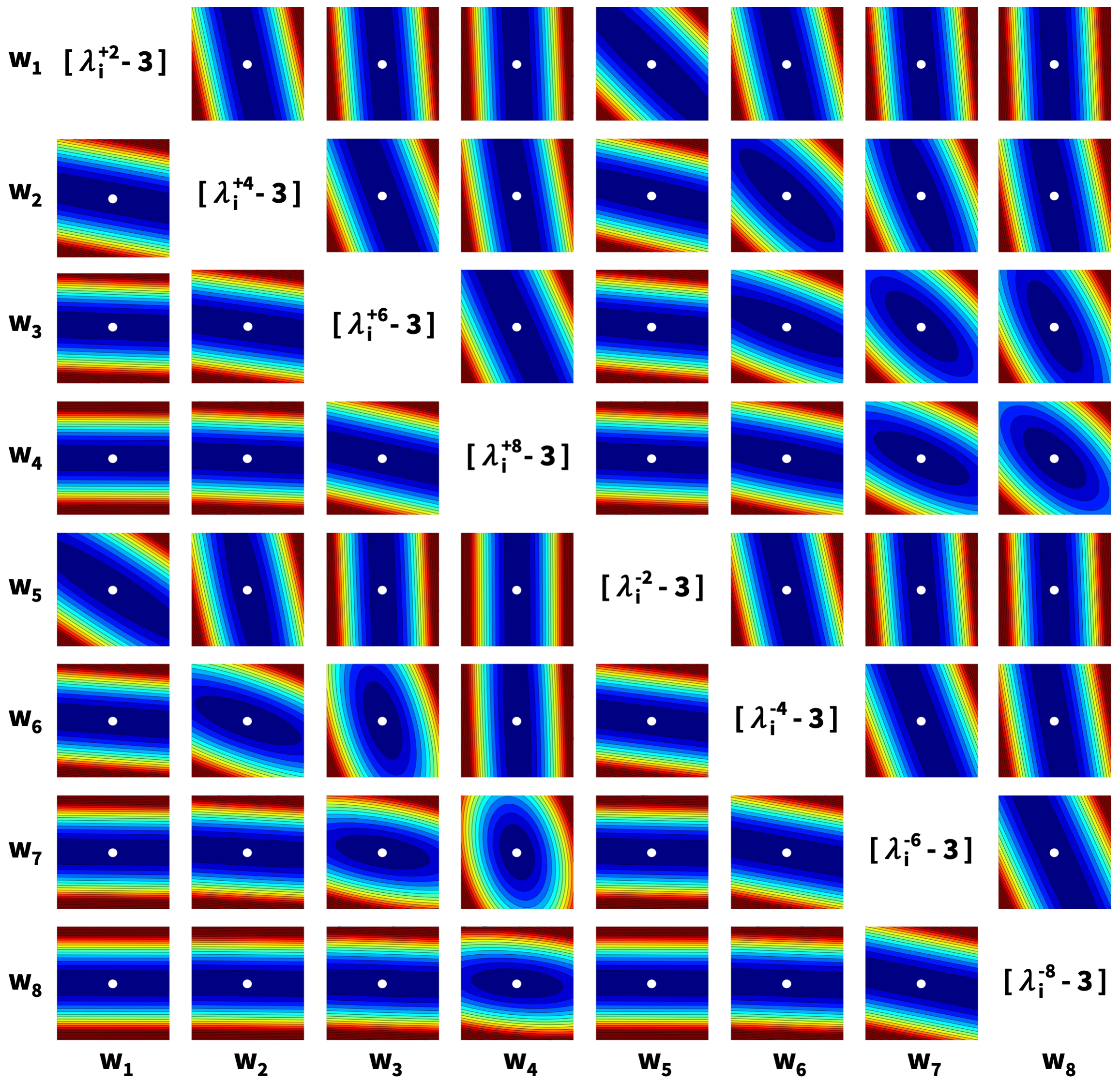}
\caption{{\bf{\sffamily{Loss functions of principal stretch based neural network.}}}
Contours of the loss function $L(\vec{\theta};\lambda,\gamma)$
for all 28 possible two-term models of the principal stretch based constitutive neural network in Figure \ref{fig04}.
The loss function is evaluated across  
tensile stretches $\lambda = [1.0,...,2.0]$,
compressive stretches $\lambda = [1.0,...,0.5]$, and
shear strains $\gamma = [0.0,...,0.5]$,
with network weights in the ranges 
$w_i = [0,...,2]$ and $w_j = [0,...,2]$.
The minimum of the loss function 
indicates the exact solution 
$w_i = 1$ and $w_j = 1$, represented through the white circle. 
The lower triangle illustrates the non-normalized loss function (\ref{lam_loss_P11_P12}),
the upper triangle illustrates the normalized loss function (\ref{lam_loss_Pten_Pcom_Pshr}).
All loss functions are convex, with contours varying from 
a few ellipsoids to many valleys with long ridges,
highlighting the collinearity of many $w_i$ and $w_j$ pairs.}
\label{fig05}
\end{figure}
Figure \ref{fig05} illustrates the contours of the loss function 
$L(\vec{\theta};\lambda,\gamma)$ for all possible two-term models of the principal stretch based network in Figure \ref{fig04}. By combining any two terms of the model and setting all other second-layer weights equal to zero, we can generate 28 possible models. 
For these 28 combinations of two terms, 
we evaluate two versions of the loss function,  
non-normalized from equation (\ref{lam_loss_P11_P12}) 
and normalized from equation (\ref{lam_loss_Pten_Pcom_Pshr}),
following the method described in Section \ref{inv_network}, 
but now using the principal stretch based definitions of the normal stress (\ref{P11_lam}) and shear stress (\ref{P12_lam}).
Similar to Section \ref{inv_network},
we plot the {\it{non-normalized}} loss function in the lower triangle and the {\it{normalized}} loss function in the upper triangle. 
Again, by design, 
all loss functions take a minimum of 
$L(\vec{\theta};\lambda,\gamma) = 0$
for the {\it{exact solution}}, $w_i = 1$ and $w_j = 1$, 
indicated through the white circles. 
From this minimum, both versions of the loss function,
non-normalized and normalized,
increase with both decreasing and increasing weights $w_i$ and $w_j$, and remain {\it{convex}} within the entire domain, for all 28 two-term models. 
Notably, in contrast to the loss function contours of the invariant based model in Figure \ref{fig03}, the contours of the principal stretch based model in Figure \ref{fig05} display less variation for different pairs of weights $w_i$ and $w_j$: 
Only a few pairs of weights generate loss functions of ellipsoidal shape, e.g., 
the 
$\{\,w_3, w_6\,\}$, $\{\,w_4, w_7\,\}$ pairs
in the non-normalized lower triangle, or the
$\{\,w_2, w_6\,\}$, $\{\,w_3, w_7\,\}$, $\{\,w_4, w_8\,\}$ pairs
in the normalized upper and non-normalized lower triangles, 
suggesting that, in the studied stretch and shear range,  
only a few pairs of terms are {\it{non-collinear}}
and would represent a solid base for a potential constitutive model. 
Most pairs of weights generate loss functions with long ridges parallel to the parameter axes, 
suggesting that many terms are almost {\it{collinear}} 
and not well suited as a functional base for a constitutive model. 
In contrast to the invariant based network, normalization does not seem to fix this issue, both the upper and lower triangle display this collinearity.  
Notably, 
the $\{\,w_1, w_5\,\}$ model in the first row and fifth column and 
the $\{\,w_5, w_1\,\}$ model in the fifth row and first column
combine the positive and negative second powers of the principal stretches,
$[\,\lambda_1^{+2}+\lambda_2^{+2}+\lambda_3^{+2}-3\,]$ and 
$[\,\lambda_1^{-2}+\lambda_2^{-2}+\lambda_3^{-2}-3\,]$, and
represent the popular Mooney Rivlin model \cite{mooney40,rivlin48}, 
which is identical to the 
$\{\,w_1, w_5\,\}$ and $\{\,w_5, w_1\,\}$ models of the invariant based model in Figure \ref{fig03}.
Overall, while these contours are difficult to interpret, we can compare them directly to Figure \ref{fig03} and realize that, within the studied stretch and shear range $\lambda$ and $\gamma$, and parameter window $w_i$, the invariant based network seems to represent a much broader spectrum of functions than the principal stretch based network for which the functional base seems to be generally more narrow and almost collinear. 
We also note that the loss function is highly {\it{sensitive to normalization}}: 
For both networks, the normalized loss functions 
(\ref{inv_loss_Pten_Pcom_Pshr}) and (\ref{lam_loss_Pten_Pcom_Pshr}) 
tend to generate more convex shapes than the non-normalized loss functions
(\ref{inv_loss_P11_P12}) and (\ref{lam_loss_P11_P12}), 
which is why we will focus on the maximum-stress normalized loss functions 
(\ref{inv_loss_Pten_Pcom_Pshr}) and (\ref{lam_loss_Pten_Pcom_Pshr})
in all following examples. 
\section{L$\!_{\mbox{\normalsize{p}}}$ Regularized Neural Networks}
\label{Lp_networks} 
We now integrate the concepts of 
$L_p$ regularization from Section \ref{Lp_regularization} and 
constitutive neural network modeling from Section \ref{neural_networks} 
and explore the resulting regression in view of {\it{predictability}} and {\it{interpretability}}. 
Specifically, we supplement the loss function of the constitutive neural network with a penalty term of $L_p$ type,
\beq
  L(\vec{\theta}; \ten{F}) 
= \frac{1}{n_{\rm{data}}}
  \sum_{i=1}^{n_{\rm{data}}} 
 || \, \ten{P} (\ten{F}_i, \vec{\theta}) - \hat{\ten{P}}_i  \, ||^2 
+ \alpha \, || \, \vec{\theta} \, ||_p^p
  \rightarrow \mbox{min}
  \quad
  \mbox{with}
  \quad
  || \, \vec{\theta} \, ||_p^p 
= \sum_{i=1}^{n_{\rm{para}}} 
  |\, w_i \,|^p \,.
\label{NN_Lp_loss}
\eeq 
The loss function minimizes the error between the model stress that we derive from the free energy of the neural network,
$\ten{P}(\ten{F}_i;\vec{\theta}) = \partial \psi / \partial \ten{F}$,
and the experimentally measured stress ${\hat{\ten{P}}}_i$ divided by the number of data points $n_{\rm{data}}$, penalized by the $L_p$ norm, 
$|| \, \vec{\theta} \, ||_p^p  = \sum_{i=1}^{n_{\rm{para}}} |\, w_i \,|^p$,
of the parameter vector 
$\vec{\theta} = \{\,w_i\,\}$ made up of the network weights $w_i$, multiplied by the penalty parameter $\alpha \ge 0$.
Specifically, we use tension, compression, and shear data and specify the stress error as the normalized difference
between the tensile, compressive, and shear stresses predicted by the neural network, 
${P}_{\rm{ten}}(\lambda_i)$, 
${P}_{\rm{com}}(\lambda_i)$, and 
${P}_{\rm{shr}}(\gamma_i)$,
and the data,
$\hat{{P}}_{{\rm{ten}},i}$,
$\hat{{P}}_{{\rm{com}},i}$, and
$\hat{{P}}_{{\rm{shr}},i}$, 
at 
$n_{\rm{ten}}$,
$n_{\rm{com}}$, and
$n_{\rm{shr}}$ stretch and shear levels $\lambda$ and $\gamma$,
\beq
\begin{array}{l@{\hspace*{0.2cm}}c@{\hspace*{0.2cm}}l}
 \D{L (\vec{\theta} ; \lambda, \gamma)}
&\D{=}
&\D{\frac{1}{n_{\rm{ten}}} \sum_{i=1}^{n_{\rm{ten}}}
  \left|\left| \, 
  \frac{{P}_{\rm{ten}}(\lambda_i) - \hat{{P}}_{{\rm{ten}},i}}
       {\hat{{P}}_{\rm{ten}}^{\rm{max}}}  \right|\right|^2 
+ \frac{1}{n_{\rm{com}}} \sum_{i=1}^{n_{\rm{com}}}
  \left|\left| \, 
  \frac{{P}_{\rm{com}}(\lambda_i) - \hat{{P}}_{{\rm{com}},i}}
       {\hat{{P}}_{\rm{com}}^{\rm{min}}}  \right|\right|^2 }\\
&\D{+}
&\D{\frac{1}{n_{\rm{shr}}} \sum_{i=1}^{n_{\rm{shr}}}
  \left|\left| \, 
  \frac{{P}_{\rm{shr}}(\gamma_i)  - \hat{{P}}_{{\rm{shr}},i}}
       {\hat{{P}}_{\rm{shr}}^{\rm{max}}}  \right|\right|^2 
+ \alpha \, || \, \vec{\theta} \, ||_p^p
  \rightarrow \mbox{min}\,.}
\end{array}
\label{NN_Lp_loss_Pten_Pcom_Pshr}
\eeq
In the following, we systematically explore the sensitivity of the loss function (\ref{NN_Lp_loss_Pten_Pcom_Pshr}) with respect to the two hyperparameters of the $L_p$ regularization, the power $p$ and the penalty parameter $\alpha$. For illustrative purposes, we first focus on a simplified two-term model, the Mooney Rivlin model that is shared between both neural networks, before we explore both $L_p$ regularized complete eight-term networks.\\[6.pt]
\noindent{\bf{\sffamily{L$\!_{\mbox{\normalsize{p}}}$ regularized 
Mooney Rivlin model.}}} 
The Mooney Rivlin model \cite{mooney40,rivlin48} is a two-term constitutive model that is located right at the intersection of the invariant based neural network in Figure \ref{fig02} and the principal stretch based network in Figure \ref{fig04}. Notably, it is the {\it{only}} model, for which both networks coincide. 
It uses the dark red term,
$[\,I_1-3\,] = [\, \lambda_1^{+2}+\lambda_2^{+2}+\lambda_3^{+2} - 3 \,]$, 
and the green term,
$[\,I_1-2\,] = [\, \lambda_2^{-2}+\lambda_2^{-2}+\lambda_3^{-2} - 3 \,]$,
of both neural networks, and weighs them by the network weights,
$w_{1,1} \, w_{2,1} = w_1$ and 
$w_{1,5} \, w_{2,5} = w_5$, 
while all other network weights are identical to zero,
\beq
\begin{array}{l@{\hspace*{0.2cm}}c@{\hspace*{0.2cm}}l}
 \D{\psi (I_1,I_2)}
&\D{=} 
&\D{w_{1,1} \, w_{2,1} \, [\,I_1-3\,]
+ w_{1,5} \, w_{2,5} \, [\,I_2-3\,] \qquad \mbox{and}} \\
 \D{\psi (\lambda_1,\lambda_2,\lambda_3)}
&\D{=}
&\D{w_1 \, [\, \lambda_1^{+2}+\lambda_2^{+2}+\lambda_3^{+2} - 3 \,]
+ w_5 \, [\, \lambda_1^{-2}+\lambda_2^{-2}+\lambda_3^{-2} - 3 \,] \,.}
\end{array}
\label{MR_energy}
\eeq
This implies that the activation of any other weight will make the invariant and principal stretch based networks drift away from one another. The Mooney Rivlin model in equation (\ref{MR_energy}) includes 
the one-term dark red Neo Hooke model \cite{treloar48} with 
$[\,I_1-3\,] = [\, \lambda_1^{+2}+\lambda_2^{+2}+\lambda_3^{+2} - 3 \,]$ and 
the one-term green Blatz Ko model \cite{blatz62} with 
$[\,I_1-2\,] = [\, \lambda_1^{-2}+\lambda_2^{-2}+\lambda_3^{-2} - 3 \,]$ 
as special cases.
For the Mooney Rivlin model, the $L_p$ regularized loss function from equation (\ref{NN_Lp_loss_Pten_Pcom_Pshr}) specifies to 
\beq
\begin{array}{l@{\hspace*{0.2cm}}c@{\hspace*{0.2cm}}l}
 \D{L (w_1,w_5 ; \lambda, \gamma)}
&\D{=}
&\D{\frac{1}{n_{\rm{ten}}} \sum_{i=1}^{n_{\rm{ten}}}
  \left|\left|  
  \frac{{P}_{\rm{ten}}(\lambda_i) - \hat{{P}}_{{\rm{ten}},i}}
       {\hat{{P}}_{\rm{ten}}^{\rm{max}}}  \right|\right|^2 
+ \frac{1}{n_{\rm{com}}}   \sum_{i=1}^{n_{\rm{com}}}\!
  \left|\left| 
  \frac{{P}_{\rm{com}}(\lambda_i) - \hat{{P}}_{{\rm{com}},i}}
       {\hat{{P}}_{\rm{com}}^{\rm{min}}}  \right|\right|^2}\\
&\D{+} 
&\D{\frac{1}{n_{\rm{shr}}} \sum_{i=1}^{n_{\rm{shr}}}
  \left|\left|  
  \frac{{P}_{\rm{shr}}(\gamma_i)  - \hat{{P}}_{{\rm{shr}},i}}
       {\hat{{P}}_{\rm{shr}}^{\rm{max}}}  \right|\right|^2       
+ \alpha_p \,  
  [ w_1^p + w_5^p ]
\rightarrow \mbox{min} \,,}
\end{array}
\label{MR_loss}
\eeq
with the Mooney Rivlin stresses 
in tension, 
$P_{\rm{ten}} = P_{11}$ for $\lambda >1$, 
and compression,
$P_{\rm{com}} = P_{11}$ for $\lambda <1$, 
from equations (\ref{P11_inv}) and (\ref{P11_lam}),
and in shear,
$P_{\rm{shr}} = P_{12}$ for all $\gamma$.
from equations (\ref{P12_inv}) and (\ref{P12_lam}),
\beq
  P_{11}
= 2\, [ \lambda - {1}/{\lambda^2} ]
      [ w_1 + w_5 / \lambda ]   
\qquad \mbox{and} \qquad
  P_{12}
= 2\,\gamma \, [ w_1 + w_5 ] \,.
\label{MR_stress}
\eeq
Notably, the uniaxial stress $P_{11}$ and shear stress $P_{12}$
of the Mooney Rivlin model (\ref{MR_stress}) are linear functions in the network weights $w_1$ and $w_5$, which translates the 
neural network loss,
$  \sum_{i=1}^{n_{\rm{ten}}}
|| [{P}_{\rm{ten}}(\lambda_i) - \hat{{P}}_{{\rm{ten}},i}]/{P}_{\rm{ten}}^{\rm{max}} ||^2 / n_{\rm{ten}} 
+  \sum_{i=1}^{n_{\rm{com}}}
|| [{P}_{\rm{com}}(\lambda_i) - \hat{{P}}_{{\rm{com}},i}]/{P}_{\rm{com}}^{\rm{min}} ||^2 / n_{\rm{com}}
+  \sum_{i=1}^{n_{\rm{shr}}}
|| [{P}_{\rm{shr}}(\lambda_i) - \hat{{P}}_{{\rm{shr}},i}]/{P}_{\rm{shr}}^{\rm{max}}  ||^2 / n_{\rm{shr}}$,
of the loss function (\ref{MR_loss}) into a linear regression problem, with a single unique global minimum.
\begin{figure}[t]
\centering
\includegraphics[width=0.72\linewidth]{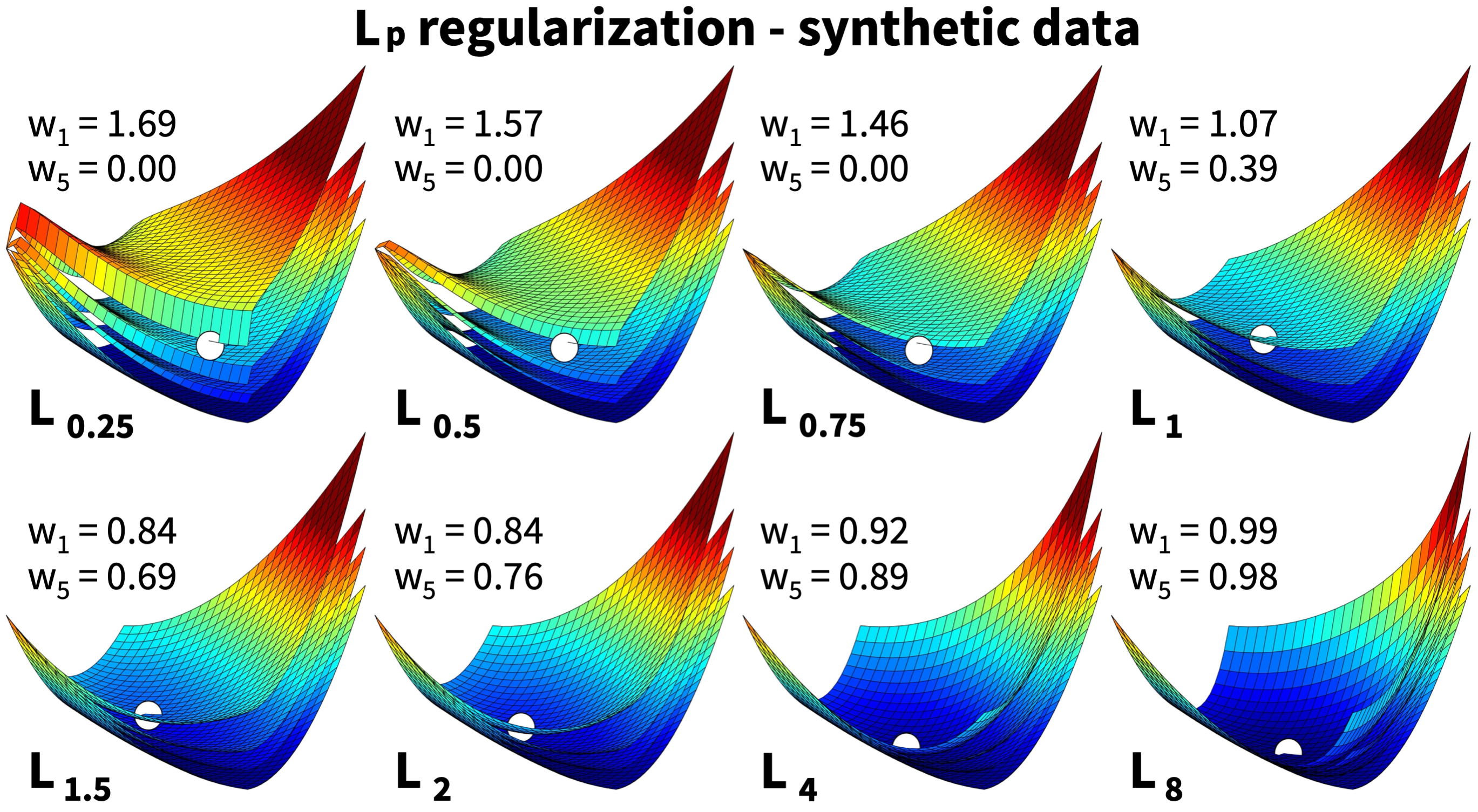}
\caption{{\bf{\sffamily{Loss functions of Lp regularized Mooney Rivlin model for synthetic data.}}} 
Contours of the $L_p$ regularized loss function,
$L(w_1,w_5;\lambda,\gamma)$,
for the two-term Mooney Rivlin model with varying powers,
$p = [0.25, 0.5, 0.75, 1, 1.5, 2, 4, 8]$,
evaluated for the two parameters, $w_1$ and $w_5$,
for synthetic data from tension, compression, and shear tests.
For $p \le 1$, top row, with the special case of 
$L_1$ regularization or lasso in the fourth column,
$L_{\rm{p}}$ regularization promotes sparsity 
by training $w_1=0$ exactly to zero, 
but the loss function is no longer strictly convex 
and has multiple local minima.
For $p  >  1$, bottom row, with the special case of 
$L_2$ regularization or ridge regression in the second column, 
$L_{\rm{p}}$ regularization promotes stability, 
retains both non-weights, $w_1>0$ and $w_5>0$, 
and maintains a convex loss function with a single global minimum.}
\label{fig06}
\end{figure}
Figure \ref{fig06} illustrates the contours of the $L_p$ regularized loss function 
$L(w_1,w_5;\lambda,\gamma)$ for the two-term Mooney Rivlin model,
with varying powers,
$p = [0.25, 0.5, 0.75, 1, 1.5, 2, 4, 8]$,
evaluated for the two parameters $w_1$ and $w_5$,
using synthetic data from tension, compression, and shear tests.
The loss function consists of the neural network loss,
$  \sum_{i=1}^{n_{\rm{ten}}}
|| [{P}_{\rm{ten}}(\lambda_i) - \hat{{P}}_{{\rm{ten}},i}]/{P}_{\rm{ten}}^{\rm{max}} ||^2 / n_{\rm{ten}} 
+  \sum_{i=1}^{n_{\rm{com}}}
|| [{P}_{\rm{com}}(\lambda_i) - \hat{{P}}_{{\rm{com}},i}]/{P}_{\rm{com}}^{\rm{min}} ||^2 / n_{\rm{com}}
+  \sum_{i=1}^{n_{\rm{shr}}}
|| [{P}_{\rm{shr}}(\lambda_i) - \hat{{P}}_{{\rm{shr}},i}]/{P}_{\rm{shr}}^{\rm{max}}  ||^2 / n_{\rm{shr}}$,
illustrated in the first row and fifth column of 
Figures \ref{fig03} and Figures \ref{fig05}; 
supplemented by the $L_p$ regularization,
$\alpha_p \, [\, |\,w_1\,|^p + |\,w_5\,|^p  \,]$,
illustrated in Figure \ref{fig01}.
For all eight graphs in Figure~\ref{fig06}, we evaluate the loss function (\ref{MR_loss}) using the Mooney Rivlin stresses (\ref{MR_stress}).
First, we generate synthetic data, 
$\hat{P}_{\rm{ten}}$, 
$\hat{P}_{\rm{com}}$,  
$\hat{P}_{\rm{shr}}$
for   
tensile stretches of $\lambda = [1.0,...,2.0]$,
compressive stretches of $\lambda = [1.0,...,0.5]$, and
shear strains of $\gamma = [0.0,...,0.5]$,
in ten equidistant increments each,
assuming an exact solution with
$w_{1,1} \, w_{2,1} = w_1 = 1$ and 
$w_{1,5} \, w_{2,5} = w_5 = 1$, 
while fixing the remaining weights equal to zero.
This results in the training data sets of eleven
stretch-stress pairs for tension, compression, and shear. 
Second, we vary the two Mooney Rivlin network weights in the ranges
$w_1 = [0,..,2]$ and $w_5 = [0,..,2]$, 
and evaluate the tensile, compressive, and shear model stresses,
${P}_{\rm{ten}}(\lambda_i)$, 
${P}_{\rm{com}}(\lambda_i)$,  
${P}_{\rm{shr}}(\gamma_i)$
using equations (\ref{MR_stress}).
Third, we evaluate the loss function (\ref{MR_loss}) 
as the normalized mean squared error between the model stresses 
${P}_{\rm{ten}}(\lambda_i)$, 
${P}_{\rm{com}}(\lambda_i)$,  
${P}_{\rm{shr}}(\gamma_i)$
and the synthetically generated data stresses
$\hat{P}_{\rm{ten}}$, 
$\hat{P}_{\rm{com}}$,  
$\hat{P}_{\rm{shr}}$,
supplemented by the $L_p$ regularization
$\alpha_p \, [\, |\,w_1\,|^p + |\,w_5\,|^p \,]$ 
for the eight different powers, 
$p = [0.25, 0.5, 0.75, 1, 1.5, 2, 4, 8]$.
As these powers $p$ increase by two orders of magnitude, 
fixing the second hyperparameter $\alpha$ to one and the same value for all eight examples
would increasingly emphasize the $L_p$ regularization 
over minimizing the actual network loss, and
generate increasingly biased results. 
Instead, for each power $p$, we select the penalty parameter $\alpha$ such that the maximum value of the loss function within the screened parameter window, in the dark red upper right corner, 
at $w_1=2$ and $w_5=2$, 
consists of equal contributions by the network term and the regularization term. 
This results in eight different penalty parameters,
$\alpha = [\, 4.69, 3.94, 3.31, 2,79, 1.97, 1,39, 0.35, 0.02 \,]$.
For each set of hyperparameters $\{p,\alpha\}$, 
we increase the penalty parameter in four increments, 
indicated through the four hyperplanes in each graph. 
Similar to the non-regularized loss functions of the invariant and principal stretch based networks in Figures \ref{fig03} and \ref{fig05}, we highlight the 
minimum of the last of these four loss functions
$L(w_i,w_j;\lambda,\gamma)$
through a white sphere. 
Importantly, in contrast to the {\it{non-regularized}} loss functions
in Figures \ref{fig03} and \ref{fig05}, 
the {\it{regularized}} loss function in Figure \ref{fig06} no longer has a minimum of 
$L(\vec{\theta};\lambda,\gamma)=0$ at $w_1 = 1$ and $w_5 = 1$.
Instead, the minimum of the loss function and its location in the $\{w_1,w_5\}$-space are now functions for the two hyperparameters $p$ and $\alpha$.
For the eight powers and penalty parameters we used in this example, the minima of the loss function at the location of the white sphere become
${\rm{min}} (L) = [\, 5.83, 5.56, 5.24, 4.82, 3.25, 2.22, 0.57, 0.04 \,]$,
and their varying locations in the $\{w_1,w_5\}$-space are indicated through the white spheres in Figure \ref{fig06}. \\[6.pt]
Figure \ref{fig06} reveals several interesting features of the $L_p$ regularized Mooney Rivlin model:
Most notably, the regularized loss function is highly sensitive to the power $p$ and varies significantly for $p$ {\it{below}} and {\it{above}} one, as we conclude from the different shapes in the first and second rows. 
For $p \le 1$, in the top row, with the special case of 
$L_1$ regularization or lasso in the fourth column,
$L_{\rm{p}}$ regularization promotes sparsity 
by training one of the weights exactly to zero, in this case $w_5 = 0$, 
while the other weight remains positive, $w_1>0$.
Importantly, for $p < 1$, the loss function is no longer convex 
and has two local minima, one at $w_1=0$ and one at $w_5=0$.
Notably, for a too small power, e.g., for $p < 0.25$, we observe a drastic regularization with sharp-contoured gradients towards the parameter planes,
and the model loses robustness.
For $p  >  1$, in the bottom row, with the special case of 
$L_2$ regularization or ridge regression in the second column, 
$L_{\rm{p}}$ regularization promotes stability and
retains both non-zero weights, $w_1>0$ and $w_5>0$.
The loss function remains convex with a single global minimum.
Increasing the penalty parameter $\alpha$ amplifies these effects and moves the regularized minimum further away from the non-regularized minimum. 
Taken together, 
while a regularization across a continuous spectrum of powers $p$ provides a lot of flexibility, 
the discovered weights $w_1$ and $w_5$ are highly sensitive to the selection of the two hyperparameters $p$ and $\alpha$:
While the power $p$ acts as a switch between {\it{sparsity}} and {\it{robustness}}, the penalty parameter $\alpha$ induces a trade-off between {\it{regularization}} and {\it{bias}}. 
\begin{figure}[t]
\centering
\includegraphics[width=0.84\linewidth]{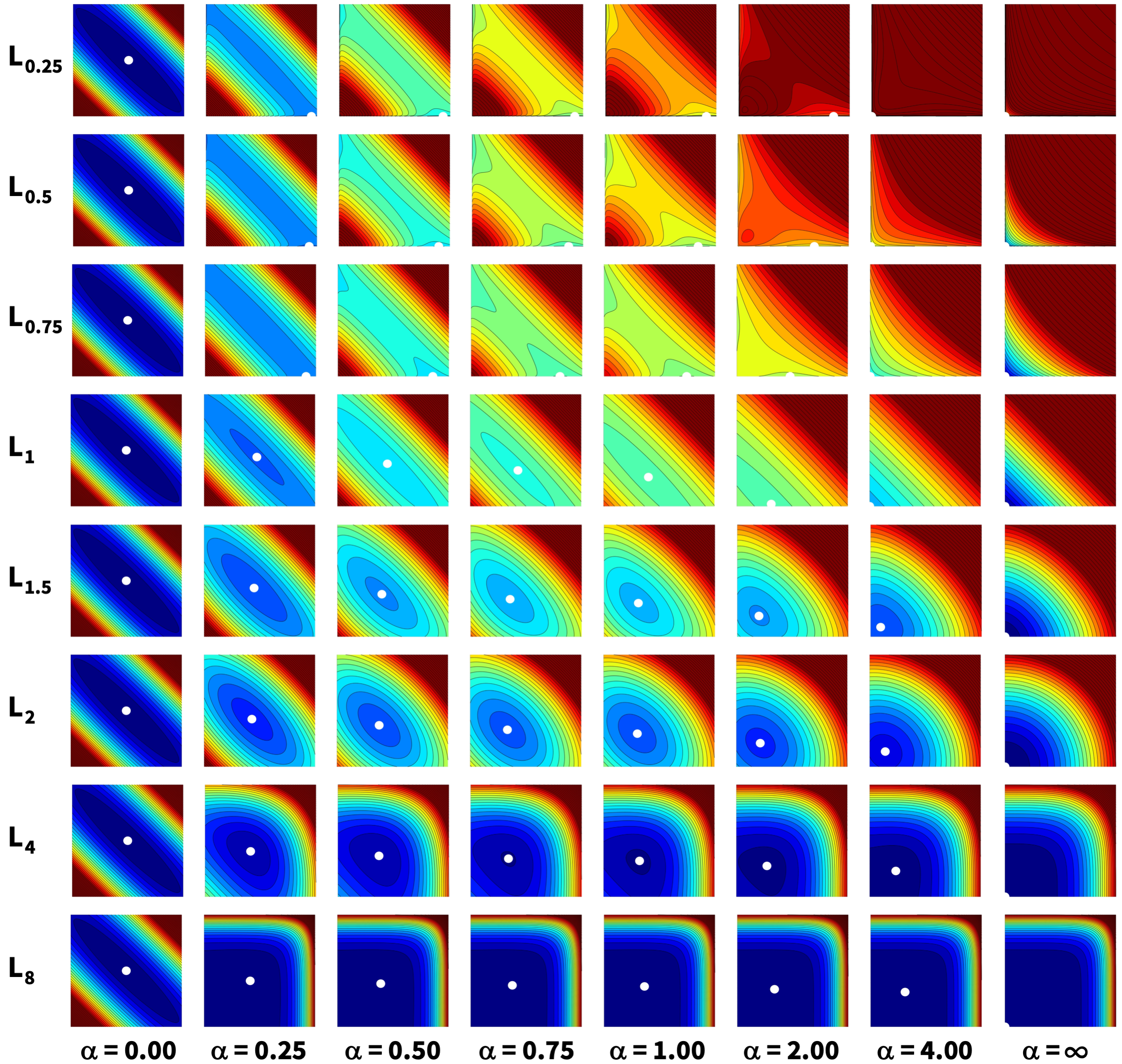}
\caption{{\bf{\sffamily{Loss functions of Lp regularized Mooney Rivlin model for synthetic data.}}} 
Contours of the $L_p$ regularized loss function,
$L(w_1,w_5;\lambda,\gamma)$,
for the two-term Mooney Rivlin model with varying powers,
$p = [0.25, 0.5, 0.75, 1, 1.5, 2, 4, 8]$, and penalty parameters
$\alpha = [0,0.25,0.50,0.75,1,2,4,\infty]$,
evaluated for the two parameters, $w_1$ and $w_5$,
for synthetic data from tension, compression, and shear tests.
Without regularization, left column, with $\alpha = 0$,
the minimum of the loss function 
is identical to the exact solution 
$w_1 = 1$ and $w_5 = 1$, represented through the white circle. 
With infinite regularization, right column, with $\alpha = \infty$,
the loss function is identical to the $L_p$ regularization term and the contours are identical to Figure \ref{fig01}.
For $p < 1$, with increasing $\alpha$, 
the loss function gradually loses convexity,
the minimum first moves towards $w_1 \ge 1$ and $w_5 = 0$,
and then towards $w_1 \rightarrow 0$ and $w_5 = 0$.
For $p > 1$, with increasing $\alpha$, 
the loss function always remains convex,
both weights always remain active, $w_1 \ge 0$ and $w_5 \ge 0$,
as the minimum moves gradually towards $w_1 \rightarrow 0$ and $w_5 \rightarrow 0$.}
\label{fig07}
\end{figure}\\[6.pt]
Figure \ref{fig07} illustrates the contours of the $L_p$ regularized loss function 
$L(w_1,w_5;\lambda,\gamma)$ for the two-term Mooney Rivlin model,
with varying powers,
$p = [0.25, 0.5, 0.75, 1, 1.5, 2, 4, 8]$,
and penalty parameters,
$\alpha = [0,0.25,0.50,0.75,1,2,4,\infty]$,
evaluated for the two parameters, $w_1$ and $w_5$,
using synthetic data from tension, compression, and shear tests.
For all 64 contour plots, we evaluate the normalized loss function (\ref{MR_loss}) following the method of Figure \ref{fig06}, but now by varying both hyperparameters, $p$ and $\alpha$. 
Without regularization, left column, with $\alpha = 0$,
all eight contour plots are identical to the non-regularized Mooney Rivlin loss function in the first row and fifth column of Figures \ref{fig03} and \ref{fig05}.
Its minimum is identical to the exact solution, 
$w_1 = 1$ and $w_5 = 1$, represented through the white circles. 
With infinite regularization, right column, with $\alpha = \infty$,
all eight contour plots are a two-dimensional projection of the $L_p$ regularization contours in Figure \ref{fig01}.
For $p \le 1$, in the four top rows,
with increasing $\alpha$, from left to right,
the loss function gradually loses strict convexity,
the minimum first moves towards $w_1 \ge 1$ and $w_5 = 0$,
and then towards $w_1 \rightarrow 0$ and $w_5 = 0$.
For $p > 1$, in the four bottom rows,
with increasing $\alpha$, from left to right,
the loss function always remains convex,
both weights always remain active, $w_1 \ge 0$ and $w_5 \ge 0$,
and move closer together as the minimum gradually moves towards zero, $w_1 \rightarrow 0$ and $w_5 \rightarrow 0$.
\\[6.pt]
Figure \ref{fig07} confirms our observations from Figure \ref{fig06} and provides additional insights into the $L_p$ regularized Mooney Rivlin model:
The regularized loss function is highly sensitive to both hyperparameters, $p$ and $\alpha$. 
Decreasing the power to or below one, $p \le 1$, 
{\it{increases interpretability}} by promoting sparsity as a subset of weights become exactly zero;
smaller powers $p$ and larger penalty parameters $\alpha$
promote sparsity more drastically and 
generate increasingly less convex loss functions. 
Increasing the power above one, $p > 1$, 
{\it{increases predictability}} by promoting robustness as the loss function becomes increasingly convex;
larger powers $p$ and larger penalty parameters $\alpha$
promote robustness more drastically and 
generate increasingly more convex loss functions. 
These observations confirm the general notion that 
$L_p$ regularization is an intricate balance 
between predictability and interpretability and
between regularization and bias
that requires a careful selection of the appropriate values for the hyperparameters $p$ and $\alpha$.
\begin{figure}[t]
\centering
\includegraphics[width=0.72\linewidth]{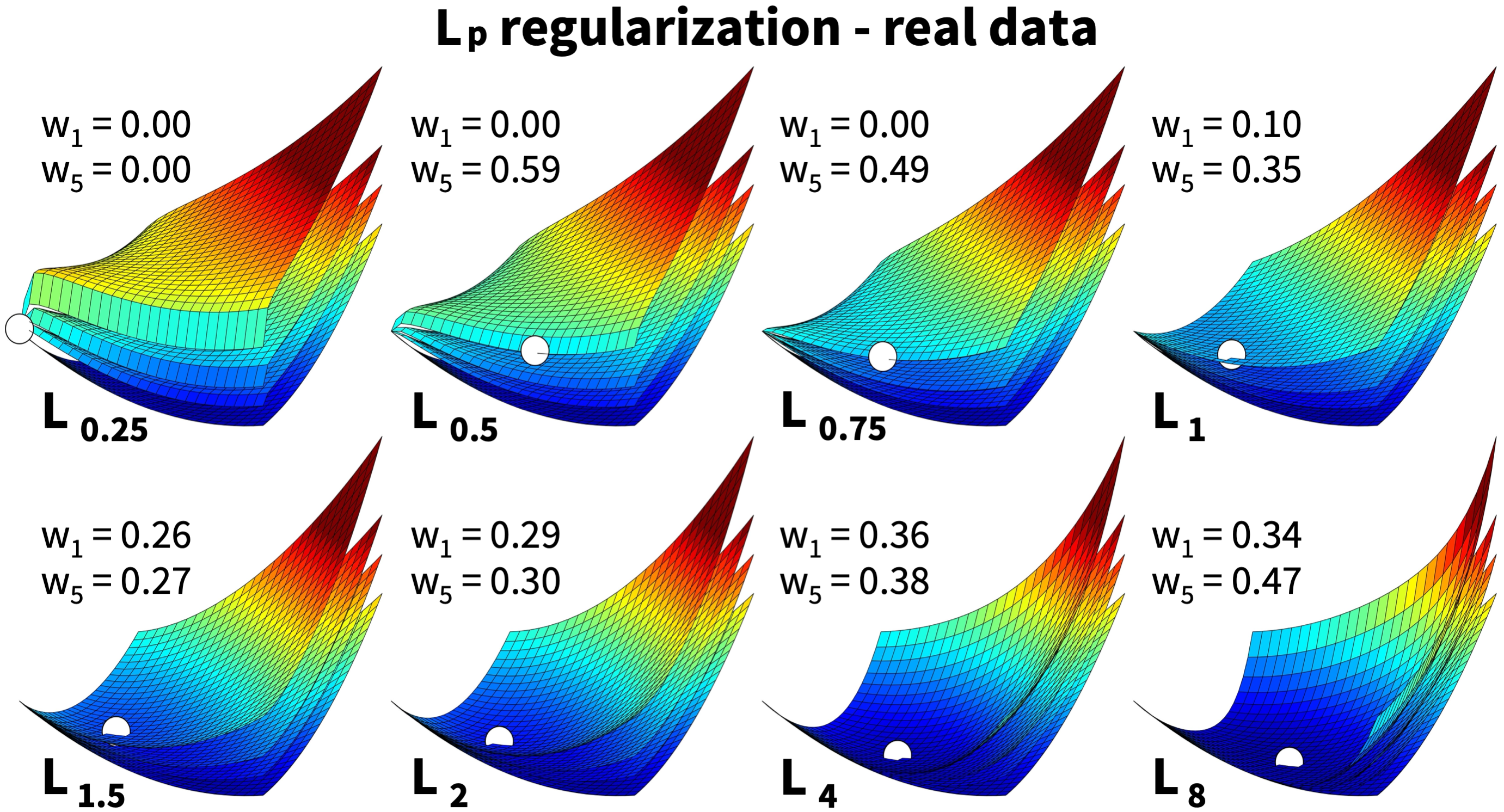}
\caption{{\bf{\sffamily{Loss functions of Lp regularized Mooney Rivlin model for real data.}}} 
Contours of the $L_p$ regularized loss function,
$L(w_1,w_5;\lambda,\gamma)$,
for the two-term Mooney Rivlin model with varying powers,
$p = [0.25, 0.5, 0.75, 1, 1.5, 2, 4, 8]$,
evaluated for the two parameters, $w_1$ and $w_5$,
for real data from tension, compression, and shear tests of human brain.
For $p \le 1$, top row, with the special case of 
$L_1$ regularization or lasso in the fourth column,
$L_{\rm{p}}$ regularization promotes sparsity 
by training $w_1=0$ exactly to zero, 
but the loss function is no longer strictly convex 
and has multiple local minima.
For $p  >  1$, bottom row, with the special case of 
$L_2$ regularization or ridge regression in the second column, 
$L_{\rm{p}}$ regularization promotes stability, 
retains both non-weights, $w_1>0$ and $w_5>0$, 
and maintains a convex loss function with a single global minimum.}
\label{fig08}
\end{figure}\\[6.pt]
Figure \ref{fig08} illustrates the contours of the $L_p$ regularized loss function 
$L(w_1,w_5;\lambda,\gamma)$ for the two-term Mooney Rivlin model,
with varying powers $p$,
evaluated for the two parameters $w_1$ and $w_5$,
but now using real data from tension, compression, and shear tests of human brain \cite{budday17}.
For all eight graphs, 
we evaluate the loss function (\ref{MR_loss}) 
as the normalized mean squared error between the model 
${P}_{\rm{ten}}(\lambda_i)$, 
${P}_{\rm{com}}(\lambda_i)$,  
${P}_{\rm{shr}}(\gamma_i)$, 
and the data
$\hat{P}_{\rm{ten}}$, 
$\hat{P}_{\rm{com}}$,  
$\hat{P}_{\rm{shr}}$,
for   
tensile stretches of $\lambda = [1.0,...,1.1]$,
compressive stretches of $\lambda = [0.9,...,1.0]$, and
shear strains of $\gamma = [0.0,...,0.2]$,
in 16 equidistant increments each \cite{linka23a},
for varying  Mooney Rivlin network weights in the ranges
$w_1 = [0,..,1]$ and $w_5 = [0,..,1]$,
and apply $L_p$ regularization,
$\alpha_p \, [\, |\,w_1\,|^p + |\,w_5\,|^p \,]$, 
for eight different powers, 
$p = [0.25, 0.5, 0.75, 1, 1.5, 2, 4, 8]$.
We select a penalty parameter 
$\alpha^{\rm{max}}=0.6585$, 
such that the maximum value of the loss function within the screened parameter window, in the dark red upper right corner, 
at $w_1=1$ and $w_5=1$, 
consists of equal contributions by the network term and the regularization term. 
For each set of hyperparameters $\{p,\alpha\}$, 
we increase the penalty parameter in four increments, 
$\alpha = [0.00, 0.25,0.50,1.00] \alpha^{\rm{max}}$,
indicated through the four hyperplanes in each graph, 
and highlight the minimum of the last of these four loss functions
$L(w_i,w_j;\lambda,\gamma)$
through a white sphere. 
Similar to Figure~\ref{fig06} based on synthetic data, the minimum of the loss function and its location in the $\{w_1,w_5\}$-space are functions for the two hyperparameters $p$ and $\alpha$.
For the plain non-regularized loss function, the minimum of the loss function is $0.0713$ and 
its weights are $w_1=0.00$ and $w_5=0.84$.
For the eight powers, the minima of the loss function at the location of the white sphere are
${\rm{min}} (L) = [\, 0.67, 0.63, 0.56, 0.50, 0.34, 0.24, 0.11, 0.08 \,]$,
and their varying locations in the $\{w_1,w_5\}$-space are indicated through the white spheres in Figure \ref{fig08}. \\[6.pt]
Figure \ref{fig08} reveals several interesting differences between the loss function for {\it{synthetic data}} in Figure \ref{fig06} and for {\it{real data}}, in this case from human brain experiments, in Figure \ref{fig08}. 
Most importantly, for the synthetic data, we assumed an {\it{exact}} minimum at $w_1=1$ and $w_5=1$, where the loss function is exactly zero for the non-$L_p$-regularized model, and takes the value of the regularization term 
$\alpha \, ||\, \vec{\theta} \,||_p^p$ otherwise. 
For the real data, we no longer know a priori where the exact minimum is and it is no longer exactly zero, since the Mooney Rivlin model is not exact for the real data. 
From screening the parameter plane, find the minimum loss at $w_1=0.00$ and $w_5=0.84$. 
Strikingly, this suggests that 
the one-parameter Blatz Ko model \cite{blatz62} with  
$\psi = w_5 \, [\,I_2-3\,]$ and
$P = w_5 \,\partial I_2 / \partial \ten{P} - p\,\ten{P}^{-t}$
is better suited to describe the experimental data than the two-parameter Mooney Rivlin \cite{mooney40,rivlin48} model. 
However, we can clearly see the negative effect of {\it{over-regularization}} with too large penalty parameters $\alpha$:
For $p < 1$, in the top row, the minimum of the loss function remains on the $w_1=0.00$ axis, but the Blatz Ko parameter is drastically reduced from its non-regularized value of $w_5=0.84$ to $w_5=0.59$, 
$w_5=0.49$, and even $w_5=0$.
For $p \ge 1$, in the bottom row, the minimum of the loss function even moves away from the $w_1=0.00$ axis, and both parameters become activated at a similar magnitude between
$w_1=[0.26,...,0.36]$ and $w_5=[0.27,...,0.47]$.
Taken together, 
the discovered weights $w_1$ and $w_5$ are highly {\it{sensitive to over-regularization}} for extreme ranges of the hyperparameters $p$ and $\alpha$:
Extreme penalty parameters induce increased {\it{bias}} as the loss function increasingly focuses on minimizing the penalty term rather than the regularization problem itself. 
\subsection{Lp regularized invariant based neural network}
\label{inv_Lp_network}
Similar to the previous example, we explore the effects of $L_p$ regularization with respect to the two hyperparameters $p$ and $\alpha$, 
but now
for the full eight-term invariant based network \cite{linka23a},
instead of the two-term Mooney Rivlin model \cite{mooney40,rivlin48}, and
for training on real instead of synthetic data.
We use tension, compression, and shear data from human brain tests \cite{budday17},
over 
a tensile range of $\lambda=[1.0,...,1.1]$,
a compressive range of $\lambda = [0.9,...,1.0]$, and
a shear range of $\gamma = [0.0,...,0.2]$,
sampled in 16 equidistant increments each, 
averaged over anywhere between $n=15$ and $n=35$ specimen \cite{linka23a}. 
We train the invariant based neural network from Figure \ref{fig02} in Section \ref{inv_network} and minimize the loss function from equation (\ref{NN_Lp_loss_Pten_Pcom_Pshr}) with the stress definitions (\ref{P11_inv}) and (\ref{P12_inv}) with three different powers,
$p = [0.5, 1.0, 2.0]$,
and four different penalty parameters,
$\alpha = [0.000, 0.001, 0.010, 0.100]$.
We use the Adam optimizer, a robust adaptive algorithm for stochastic gradient-based first-order optimization \cite{kingma14}.  
\begin{figure}[t]
\centering
\includegraphics[width=0.84\linewidth]{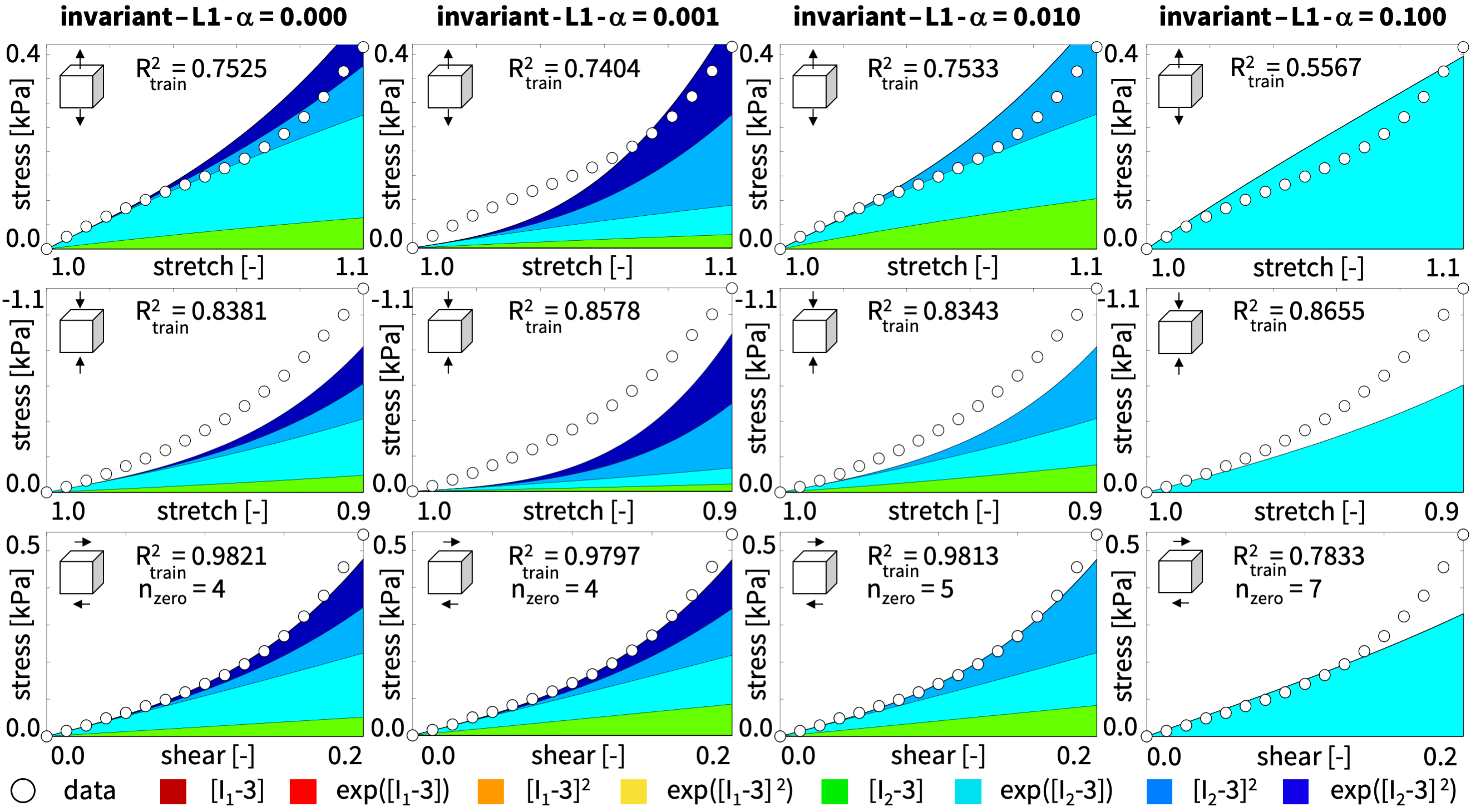}
\caption{{\bf{\sffamily{Discovered models of L1 regularized invariant based network.}}} 
Nominal stress as a function of stretch or shear strain for the invariant based neural network with $L_1$ regularization for varying 
penalty parameters $\alpha = [\, 0.000, 0.001, 0.010, 0.100 \,]$, 
trained with human gray matter tension, compression, and shear data. 
Circles represent the experimental data. 
Color-coded regions represent the discovered model terms. 
Coefficients of determination ${\text{R}}^2$ indicate the goodness of fit.}
\label{fig09}
\centering
\includegraphics[width=0.84\linewidth]{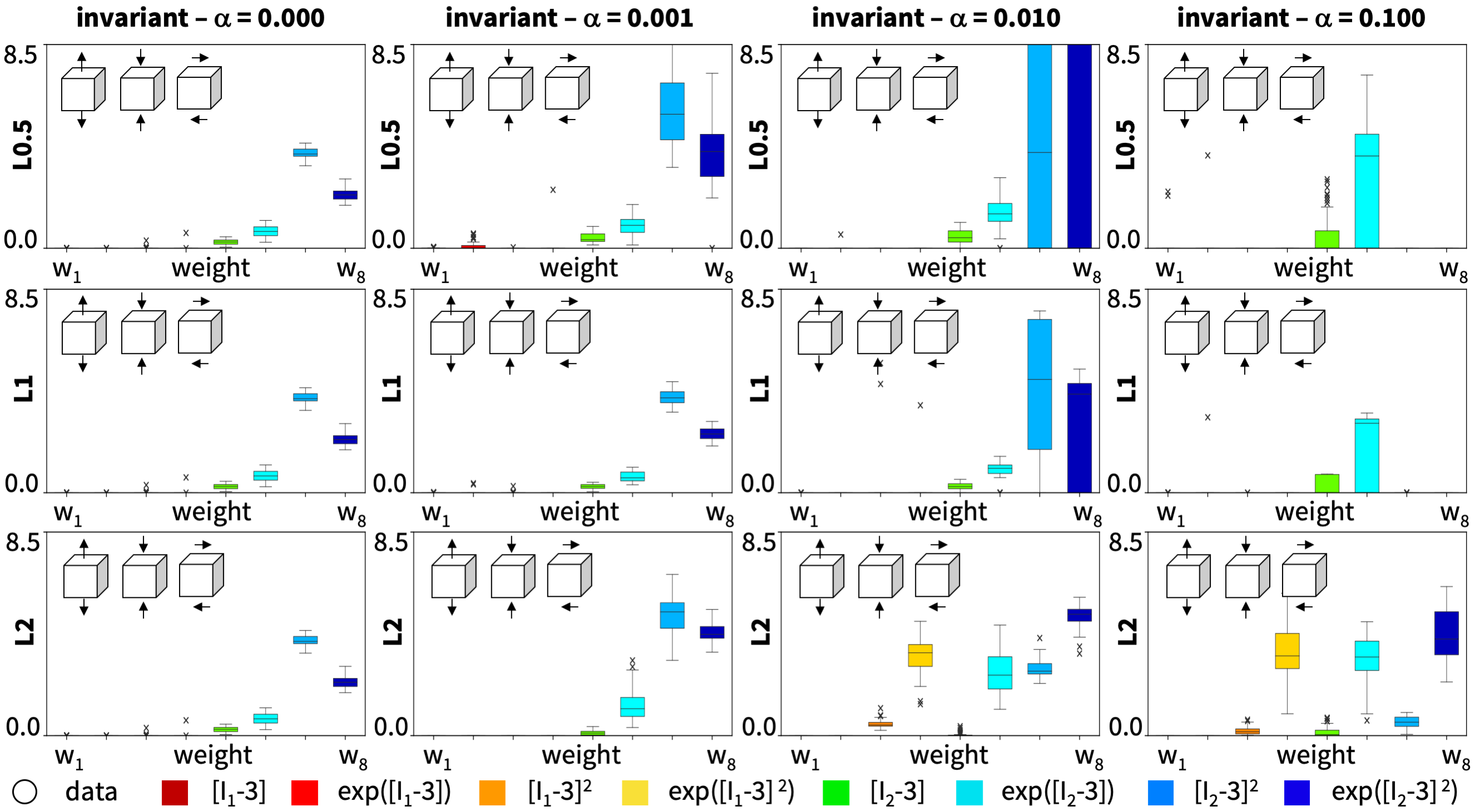}
\caption{{\bf{\sffamily{Discovered models of Lp regularized invariant based network.}}} 
Distribution of discovered weights for the invariant based neural network with $L_p$ regularization 
for varying powers $p=[\,0.5, 1.0, 2.0]$ and 
penalty parameters $\alpha = [\, 0.000, 0.001, 0.010, 0.100 \,]$. 
Colored boxes indicate the relevance of the eight model terms, with means and standard deviations from $n=100$ realizations with varying initializations of the network weights.}
\label{fig10}
\end{figure}\\[6.pt]
Figure \ref{fig09} summarizes our four discovered models in terms of the nominal stress as a function of stretch or shear strain, with the penalty parameter $\alpha$ increasing from left to right. 
The circles represent the experimental data \cite{budday17}.
The color-coded regions represent the stress contributions of the eight model terms according to Figure \ref{fig02}.
The coefficients of determination ${\text{R}}^2$ quantify the goodness of fit.
Overall, the $L_p$ regularized invariant based network trains solidly and provides a good fit of the data. 
Without regularization, in the left column, 
the network discovers four non-zero terms, 
all in terms of the second invariant, $[\,I_2-3\,]$, 
indicated through cold green-to-blue colors, 
\[
  \psi
= w_{5} \,   [\,I_2 - 3\,]
+ w_{2,6} \, [ \, \exp (\,   w_{1,6} \, [\, I_2 -3 \,] )   - 1\,] 
+ w_{7} \,   [\,I_2 - 3\,]^2
+ w_{2,8} \, [ \, \exp (\,   w_{1,8} \, [\, I_2 -3 \,]^2 ) - 1\,] \,,
\]
with stiffness-like parameters 
$w_{5}  =0.129$\,kPa,
$w_{2,6}=0.358$\,kPa,
$w_{7}  =3.840$\,kPa, and
$w_{2,8}=1.406$\,kPa 
and exponential weights,
$w_{1,6}=1.152$ and
$w_{1,8}=2.891$, 
and its stress takes the following form,
$ \ten{P}
= [ w_{5} + w_{2,6} w_{1,6} \exp (\,w_{1,6} [\, I_2 -3 \,]) 
+ 2\,[\,I_2 - 3\,]
  [ w_{7} + w_{2,8} w_{1,8} \exp (\,w_{1,8} [\, I_2 -3 \,]^2)]]
  \partial I_2/\partial \ten{F}
- p \, \ten{F}^{-{\rm{t}}}$.
As the penalty parameter increases, from left to right, 
the number of non-zero terms decreases.
With a penalty parameter $\alpha=0.010$, in the third column, 
the network discovers three non-zero terms, 
all in terms of the second invariant, $[\,I_2-3\,]$, 
indicated through cold green-to-light-blue colors, 
\[
  \psi
= w_{5} \,   [\,I_2 - 3\,]
+ w_{2,6} \, [ \, \exp (\,   w_{1,6} \, [\, I_2 -3 \,] )   - 1\,] 
+ w_{7} \,   [\,I_2 - 3\,]^2 \,,
\]
with stiffness-like parameters 
$w_{5}  =0.231$\,kPa,
$w_{2,6}=1.443$\,kPa, and
$w_{7}  =7.364$\,kPa, 
and the exponential weight,
$w_{1,6}=5.102$,
and its stress takes the following form,
$ \ten{P}
= [ w_{5} + w_{2,6} w_{1,6} \exp (\,w_{1,6} [\, I_2 -3 \,]) 
+ 2\,[\,I_2 - 3\,]
  w_{7} ]
  \partial I_2/\partial \ten{F}
- p \, \ten{F}^{-{\rm{t}}}$.
For the largest penalty parameter, in the right column, 
the network discovers a single non-zero term,
the turquoise linear exponential term of the second invariant,
\[
  \psi
= w_{2,6} \, [\, \exp(\,w_{1,6}\,[\,I_2-3\,])-1\,] \,,
\]
with the stiffness-like parameter $w_{2,6}=0.2462$\,kPa 
and the exponential weight        $w_{1,6}=2.9937$, 
and its stress takes the following form,
$ \ten{P} 
= w_{2,6}\,w_{1,6} \, \exp(\,w_{1,6}\,[\,I_2-3\,]) \, \partial I_2 / \partial \ten{F} 
- p\,\ten{F}^{-\scas{t}}$.
While Figure \ref{fig09} provides great visual insights into the performance of $L_1$ regularization with varying penalty parameters, it only represents a {\it{snapshot of model discovery}} in the eight-dimensional parameter space of the network. Subset selection and model discovery are not only sensitive to the initialization of the parameter vector 
$\vec{\theta} = \{\,w_i\,\}$, but also to the stochastic nature of the Adam optimizer. This implies that different runs may produce different results. This raises the question how {\it{reproducible}} and {\it{robust}} the results in Figure \ref{fig09} are for varying initial conditions and training runs. \\[6.pt]
Figure \ref{fig10} summarizes the discovered weights for the invariant based network with $L_p$ regularization 
for varying powers $p=[\,0.5, 1.0, 2.0]$ and 
penalty parameters $\alpha = [\, 0.000, 0.001, 0.010, 0.100 \,]$. 
For all twelve combinations of the two hyperparameters, we perform a total of $n=100$ training runs each, with varying initial conditions for the network weights $w_i = \{\,w_1,...,w_8\,\}$, such that each of the four models in Figure \ref{fig09} is the result of one of the $L_1$ regularized training runs in the middle row. The colored boxes in Figure \ref{fig10} indicate the relevance of the eight model terms, with their means and standard deviations.
Interestingly, the $L_{0.5}$ and $L_{0.1}$ regularizations in the first and second rows perform qualitatively similarly: They both start with four dominant terms, all in terms of the second invariant. Except for a small number of outliers, they both converge to two dominant one-term models, the green $[\,I_2-3\,]$ and the turquoise $\rm{exp}([\,I_2-3\,])$ models, while all other weights train to zero. 
The fact that both networks {\it{alternate}} between these two terms 
is a result of the {\it{non-convex}} nature 
of the underlying nonlinear regression problem 
associated with the invariant based network and 
indicates the existence of {\it{multiple local minima}}.
Instead, the $L_{2}$ regularization in the bottom row 
converges to a model that consistently 
trains the 
dark red $[\,I_1-3\,]$ and
red $\rm{exp}([\,I_1-3\,])$ terms to zero,
and maintains six non-zero terms, 
of which the
yellow $\rm{exp}([\,I_1-3\,]^2)$,
turquoise $\rm{exp}([\,I_2-3\,])$, and 
dark blue $\rm{exp}([\,I_2-3\,]^2)$ terms are dominant.
\begin{figure}[t]
\centering
\includegraphics[width=0.72\linewidth]{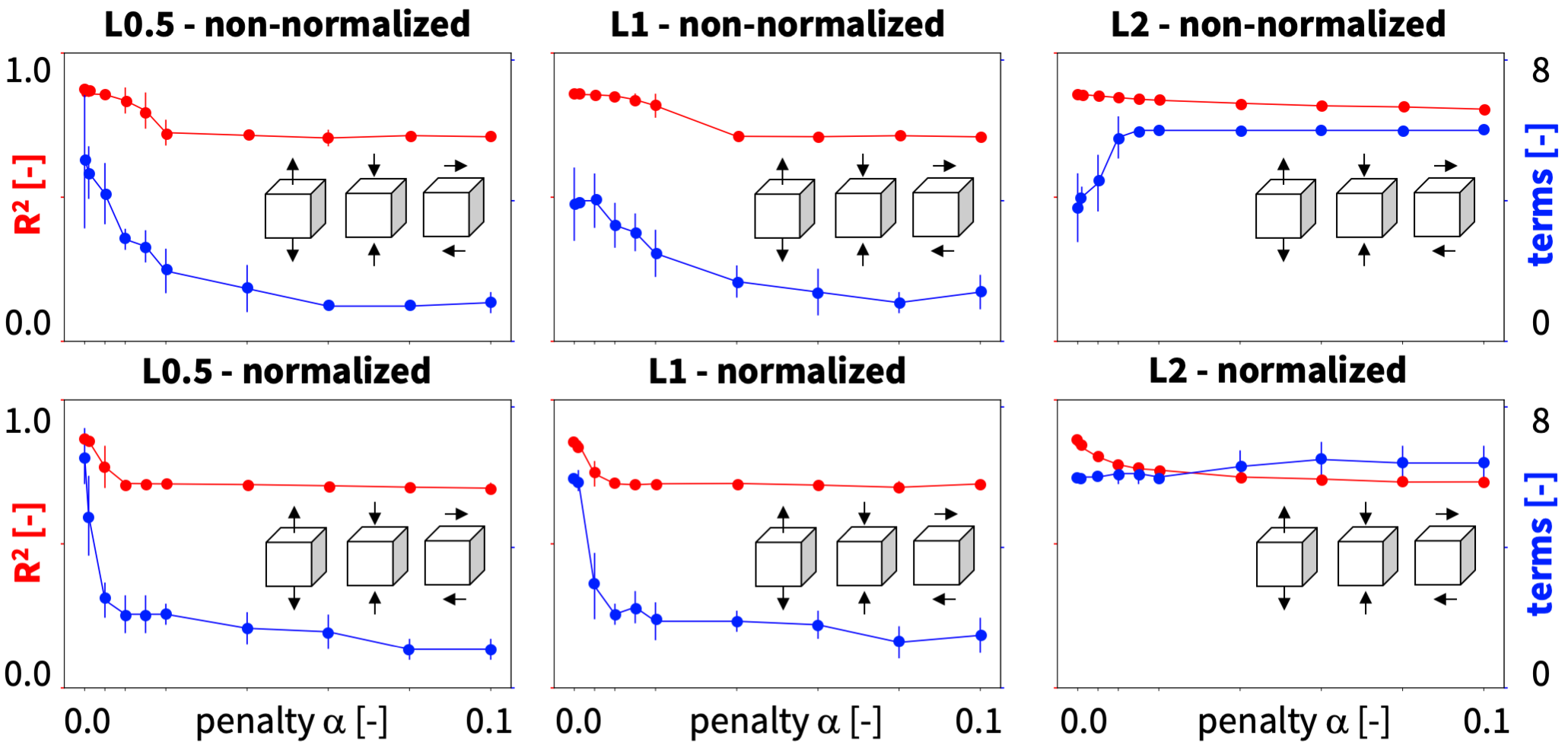}
\caption{{\bf{\sffamily{Convergence of Lp regularized invariant based network.}}} 
Goodness of fit and number of terms for the invariant based neural network with $L_p$ regularization 
for varying powers $p=[\,0.5, 1.0, 2.0 \,]$ and 
penalty parameters $\alpha = [\, 0.000, 0.001, 0.005, 0.010, 0.015, 0.020, 0.040, 0.060, 0.080, 0.100 \,]$. 
Top row uses a non-normalized regularization in terms of the weights $w_i$,
bottom row uses  a normalized regularization in terms of the $L_0$-normalized weights
$w_i\,/\,w_{i,L_0}$. 
Red dots indicate the coefficient of determination $R^2$,
blue dots indicate the number of terms, with means and standard deviations from $n=10$ realizations.}
\label{fig11}
\end{figure}\\[6.pt]
Figure \ref{fig11} summarizes the convergence of the $L_p$ regularized invariant based network in terms of the goodness of fit and number of terms, 
for varying powers $p=[\,0.5, 1.0, 2.0 \,]$ and 
penalty parameters $\alpha = [\, 0.000, 0.001, 0.005, 0.010, 0.015, 0.020, 0.040, 0.060, 0.080, 0.100 \,]$. 
Red dots indicate the coefficient of determination $R^2$,
blue dots indicate the number of terms, with means and standard deviations from $n=10$ realizations.
A known shortcoming of the $L_p$ regularization is that it introduces bias and moves the solution away from the minimum of the network loss towards the minimum of the regularization loss.  
This is particularly critical for our network in which all weights have a different meaning and potentially also a different magnitude. 
To quantify the effects of this potential limitation, 
Figure \ref{fig11} compares the non-normalized regularization,
$ || \, \vec{\theta} \, ||_p^p 
= \sum_{i=1}^{n_{\rm{para}}} |\, w_i \,|^p$, 
in terms of the weights $w_i$ that we have used throughout this study 
against a normalized regularization,
$ || \, \vec{\theta} \, ||_p^p 
= \sum_{i=1}^{n_{\rm{para}}} |\, w_i \,/\, w_{i,L_0} \,|^p$ , 
in terms of the $L_0$-normalized weights
$w_i\,/\,w_{i,L_0}$. 
Here, $w_{i,L_0}$ are the weights of the one-term models from the diagonal in Table \ref{tab01}. \\[6.pt]
Figure \ref{fig11} confirms that regularization is a trade-off between 
{\it{error}} and {\it{complexity}}, or similarly, between
the {\it{goodness of fit}} and the {\it{number of terms}}. 
While the $L_{0.5}$ and $L_1$ regularizations behave qualitatively similarly and promote sparsity by reducing the number of non-zero terms to one, the $L_2$ promotes robustness by maintaining a large subset of six non-zero terms. 
The $L_1$ regularization is less aggressive than the $L_{0.5}$ regularization and requires larger penalty parameters $\alpha$ to achieve a similar sparseness, which could induce a larger bias, away from minimum of the network loss towards the minimum of the regularization loss. 
Normalizing the $L_p$ penalty term by using the $L_0$-normalized weights
$w_i\,/\,w_{i,L_0}$ instead of the non-normalized weights $w_i$ accelerates the positive effects of regularization, especially in the small-penalty-parameter regime, and could provide a viable solution to reduce regularization-induced bias. Ultimately, in the large-penalty-parameter regime, the non-normalized and normalized regularizations converge towards a similar goodness of fit and number of terms. \\[6.pt]
Taken together,
our results confirm the general notion that $L_p$ regularization 
{\it{increases interpretability}} 
for powers equal to or below one, $p \le 1$, 
by promoting sparsity as a subset of weights train exactly to zero; and
{\it{increases predictability}}
for powers larger than one, $p > 1$, 
by promoting robustness as a unique subset of weights emerges as dominant.
Larger penalty parameters $\alpha$ amplify these trends at the price of an increased bias, which we can reduce, at least in part, by normalizing the network weights in the the penalty term. 
\\[6.pt]
\noindent{\bf{\sffamily{L$_{\tens{0}}$ regularized 
invariant based neural network.}}} 
For comparison, we explore the effects of $L_0$ regularization using the same 
tension, compression, and shear data from human brain tests as in the previous example \cite{budday17}. 
We train the invariant based neural network from Figure \ref{fig02} in Section \ref{inv_network} and minimize the loss function from equation (\ref{NN_Lp_loss_Pten_Pcom_Pshr}) with the stress definitions (\ref{P11_inv}) and (\ref{P12_inv}), but now use a penalty term,
$\alpha \, || \, \vec{\theta} \, ||_0$, with the $L_0$ norm
$|| \, \vec{\theta} \, ||_0 = \sum_{i=1}^{n_{\rm{para}}} I(w_i \ne 0)$,
to penalize the total number of non-zero terms in the model. 
In essence, $L_0$ regularization turns network training into a 
{\it{discrete combinatorial problem}} with $2^8-1=255$ possible models, 
8 with a single term, 
28 with two,
56 with three,
70 with four,
56 with five,
28 with six,
8 with seven, and
1 with all eight terms. 
For illustrative purposes, we focus on the eight {\it{one-term}} and 28 {\it{two-term models}} 
that follow by explicitly setting the other seven and six terms of the network to zero. 
\begin{table}[h]
\centering
\caption{{\bf{\sffamily{L$_{\tens{0}}$ regularized 
invariant based neural network.}}} 
Weights and remaining loss of the one- and two-term models of the $L_0$ regularized 
invariant based neural network. 
The diagonale summarizes the discovered one-term models penalized by $\alpha$,
the off-diagonale the two-term models penalized by $2\alpha$. 
Best-in-class models are highlighted in bold.}
\vspace*{0.2cm} 
\footnotesize
\renewcommand{\arraystretch}{0.9}
\label{tab01}
\hspace*{-0.4cm} 
\begin{tabular}{|c||c|c|c|c|c|c|c|c|} \hline               
      & & & & & & & & \\ \\ [-16.pt]
      & {\bf{\sffamily{w}}}$_{\bf{1}}$ 
      & \makecell{{\bf{\sffamily{w}}}$_{\bf{1,2}}$, {\bf{\sffamily{w}}}$_{\bf{2,2}}$}
      & {\bf{\sffamily{w}}}$_{\bf{3}}$ 
      & \makecell{{\bf{\sffamily{w}}}$_{\bf{1,4}}$, {\bf{\sffamily{w}}}$_{\bf{2,4}}$}
      & {\bf{\sffamily{w}}}$_{\bf{5}}$ 
      & \makecell{{\bf{\sffamily{w}}}$_{\bf{1,6}}$, {\bf{\sffamily{w}}}$_{\bf{2,6}}$}
      & {\bf{\sffamily{w}}}$_{\bf{7}}$ 
      & \makecell{{\bf{\sffamily{w}}}$_{\bf{1,8}}$, {\bf{\sffamily{w}}}$_{\bf{2,8}}$}  \\ [2.pt] \hline \hline
  \makecell{{\bf{\sffamily{w}}}$_{\bf{1}}$}
& \makecell{0.796} 
& \makecell{0.237 \\ 0.918, 0.600}
& \makecell{0.400 \\10.048}
& \makecell{0.403 \\ 3.666, 2.718}
& \makecell{0.000 \\ 0.840}
& \makecell{0.000 \\ 0.957, 0.865}
& \makecell{0.330 \\12.545}
& \makecell{0.330 \\ 3.810, 3.286} \\ \hline
 {\bf{\sffamily{loss}}}
& 0.092 + $\alpha$ & 0.090 +2$\alpha$ & 0.060 +2$\alpha$ & 0.060 +2$\alpha$
& 0.071 +2$\alpha$ & 0.069 +2$\alpha$ & 0.040 +2$\alpha$ & 0.040 +2$\alpha$ \\ \hline \hline 
  \makecell{\!\!{\bf{\sffamily{w}}}$_{\bf{1,2}}$,{\bf{\sffamily{w}}}$_{\bf{2,2}}$\!\!\!}
& \makecell{0.918, 0.600 \\ 0.237}
& \makecell{1.076, 0.727 }
& \makecell{0.980, 0.410 \\9.811}
& \makecell{1.219, 0.329 \\4.167, 2.342}
& \makecell{0.369, 0.000 \\0.841}
& \makecell{0.558, 0.000 \\3.186, 0.250}
& \makecell{1.095, 0.294 \\12.547}
& \makecell{0.822, 0.407 \\4.089, 2.993} \\ \hline
 {\bf{\sffamily{loss}}}
& 0.090 +2$\alpha$ & 0.089 + $\alpha$ & 0.060 +2$\alpha$ & 0.060 +2$\alpha$ 
& 0.071 +2$\alpha$ & 0.063 +2$\alpha$ & 0.040 +2$\alpha$ & 0.040 +2$\alpha$ \\ \hline \hline
  \makecell{{\bf{\sffamily{w}}}$_{\bf{3}}$} 
& \makecell{10.048 \\ 0.400} 
& \makecell{9.811 \\ 0.980, 0.410} 
& \makecell{18.348}
& \makecell{9.388\\3.011, 2.977}
& \makecell{8.151\\0.507} 
& \makecell{8.173\\0.876, 0.569} 
& \makecell{8.216\\10.916} 
& \makecell{8.178\\3.193, 3.422 } \\ \hline
 {\bf{\sffamily{loss}}}
& 0.060 +2$\alpha$ & 0.060 +2$\alpha$ & 0.086 + $\alpha$ & 0.086 +2$\alpha$ 
& 0.049 +2$\alpha$ & 0.049 +2$\alpha$ & 0.069 +2$\alpha$ & 0.070  \\ \hline \hline
  \makecell{\!\!{\bf{\sffamily{w}}}$_{\bf{1,4}}$, {\bf{\sffamily{w}}}$_{\bf{2,4}}$\!\!\!}
& \makecell{3.666, 2.718\\0.403} 
& \makecell{4.167, 2.342\\1.219, 0.329} 
& \makecell{3.011, 2.977\\9.388} 
& \makecell{4.305, 4.244} 
& \makecell{3.134, 2.672\\0.497} 
& \makecell{3.014, 2.669\\1.266, 0.394} 
& \makecell{3.210, 2.559\\10.896} 
& \makecell{2.973, 2.737\\3.149, 3.477} \\ \hline
 {\bf{\sffamily{loss}}}
& 0.060 +2$\alpha$ & 0.060 +2$\alpha$ & 0.086 +2$\alpha$ & 0.086 + $\alpha$ 
& 0.049 +2$\alpha$ & 0.048 +2$\alpha$ & 0.070 +2$\alpha$& 0.070  +2$\alpha$ \\ \hline \hline
  \makecell{{\bf{\sffamily{w}}}$_{\bf{5}}$} 
& \makecell{0.840\\0.000}
& \makecell{0.841\\0.369, 0.000} 
& \makecell{0.507\\8.151}
& \makecell{0.497\\3.134, 2.672}
& \makecell{\bf{0.840}}
& \makecell{0.234\\0.747, 0.803}
& \makecell{\bf{0.406}\\ \bf{11.178}}
& \makecell{\bf{0.411}\\ \bf{3.644, 3.030}} \\ \hline
 {\bf{\sffamily{loss}}}
& 0.071 +2$\alpha$ & 0.071 +2$\alpha$ & 0.049 +2$\alpha$ & 0.049 +2$\alpha$ 
& \bf{0.071 +} $\vec{\alpha}$ & 0.070 +2$\alpha$ & {\bf{0.033 +2}}$\vec{\alpha}$ & {\bf{0.033 +2}}$\vec{\alpha}$ \\ \hline \hline
  \makecell{\!\!{\bf{\sffamily{w}}}$_{\bf{1,6}}$,{\bf{\sffamily{w}}}$_{\bf{2,6}}$\!\!\!}
& \makecell{0.957, 0.865\\0.000}
& \makecell{3.186, 0.250\\0.558, 0.000} 
& \makecell{0.876, 0.569\\8.173} 
& \makecell{1.266, 0.394\\3.014, 2.669} 
& \makecell{0.747, 0.803\\0.234} 
& \makecell{\bf{0.898, 0.923}} 
& \makecell{\bf{1.022, 0.396}\\ \bf{11.014}}
& \makecell{\bf{0.839, 0.484}\\ \bf{3.427, 3.208}} \\ \hline
 {\bf{\sffamily{loss}}}
& 0.069 +2$\alpha$ & 0.063 +2$\alpha$ & 0.049 +2$\alpha$ & 0.048 +2$\alpha$ 
& 0.070 +2$\alpha$ & \bf{0.069 +} $\vec{\alpha}$ & {\bf{0.033 +2$\vec{\alpha}$}} & {\bf{0.033 +2$\vec{\alpha}$}} \\ \hline \hline
  \makecell{{\bf{\sffamily{w}}}$_{\bf{7}}$} 
& \makecell{12.545\\0.330}
& \makecell{12.547\\1.095, 0.294}
& \makecell{10.916\\8.216}
& \makecell{10.896\\3.210, 2.559} 
& \makecell{\bf{11.178}\\ \bf{0.406}} 
& \makecell{\bf{11.014}\\ \bf{1.022, 0.396}} 
& \makecell{\bf{19.599}}
& \makecell{ 9.520\\3.565, 2.834} \\ \hline
 {\bf{\sffamily{loss}}}
& 0.040 +2$\alpha$ & 0.040 +2$\alpha$ & 0.069 +2$\alpha$ & 0.070 +2$\alpha$ 
& {\bf{0.033 +2}}$\vec{\alpha}$ & {\bf{0.033 +2}}$\vec{\alpha}$ & \bf{0.059 +} $\vec{\alpha}$ & 0.059 +2$\alpha$ \\ \hline \hline
  \makecell{\!\!{\bf{\sffamily{w}}}$_{\bf{1,8}}$, {\bf{\sffamily{w}}}$_{\bf{2,8}}$\!\!\!}
& \makecell{3.810, 3.286\\0.330}
& \makecell{4.089, 2.993\\0.822, 0.407} 
& \makecell{3.193, 3.422\\8.178} 
& \makecell{3.149, 3.477\\2.973, 2.737} 
& \makecell{\bf{3.644, 3.030}\\ \bf{0.411}} 
& \makecell{\bf{3.427, 3.208}\\ \bf{0.839, 0.484}}
& \makecell{3.565, 2.834\\9.520} 
& \makecell{\bf{4.560, 4.280}} \\ \hline
 {\bf{\sffamily{loss}}}
& 0.040 +2$\alpha$ & 0.040 +2$\alpha$ & 0.070 +2$\alpha$ & 0.070 +2$\alpha$ 
& {\bf{0.033 +2}}$\vec{\alpha}$ & {\bf{0.033 +2}}$\vec{\alpha}$ & 0.059 +2$\alpha$ & {\bf{0.060}} + $\vec{\alpha}$ \\ \hline 
\end{tabular}
\end{table}
\begin{figure}[t]
\centering
\includegraphics[width=0.84\linewidth]{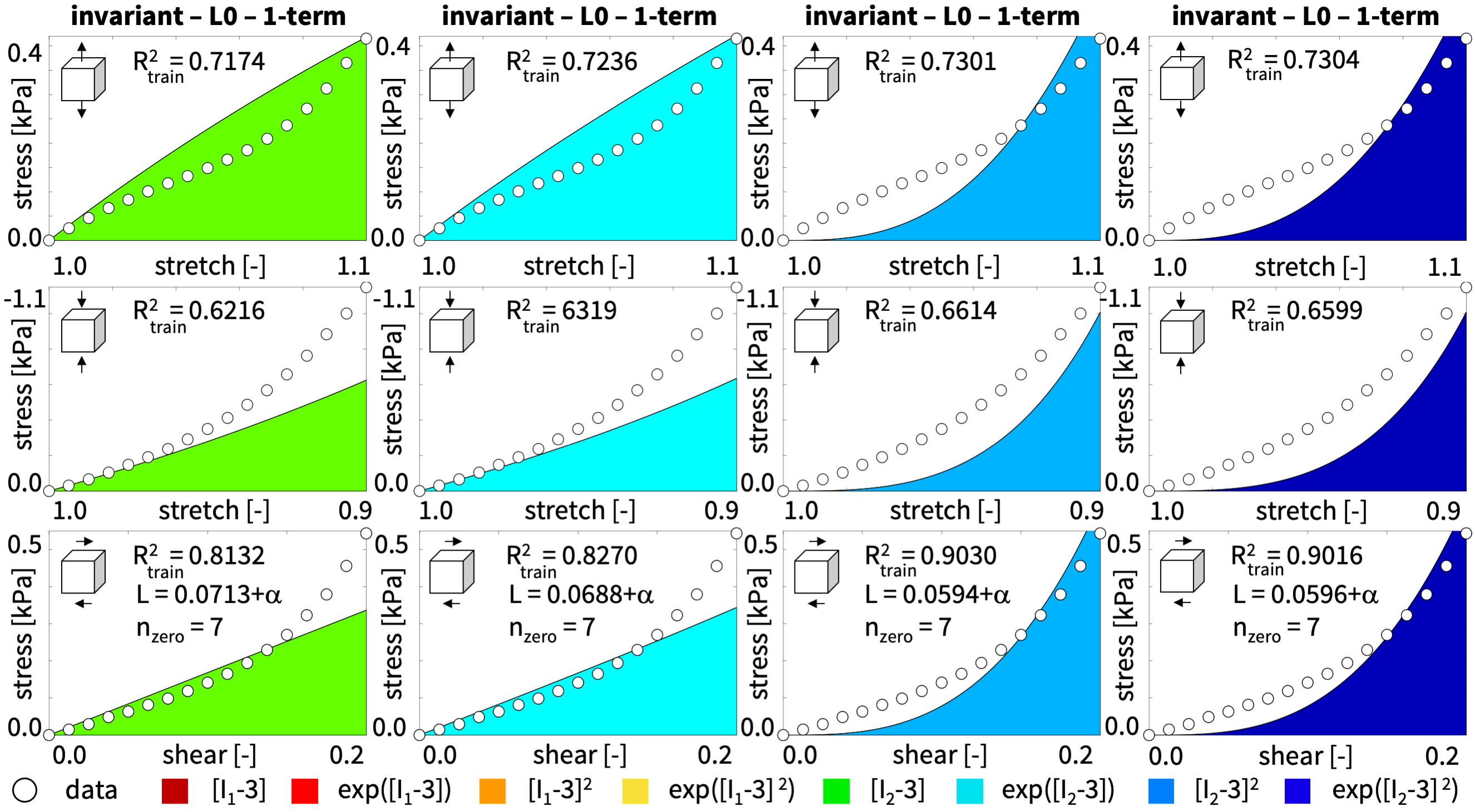}
\caption{{\bf{\sffamily{Discovered best-in-class one-term models of L$_0$ regularized invariant based network.}}} 
Nominal stress as a function of stretch or shear strain for the invariant based constitutive neural network with $L_0$ regularization, 
trained with human gray matter tension, compression, and shear data. 
Circles represent the experimental data. 
Color-coded regions represent the discovered model terms. 
Coefficients of determination ${\text{R}}^2$ indicate the goodness of fit for each individual test;
remaining loss $L$ indicates the quality of the overall fit.}
\label{fig12}
\vspace*{0.25cm}
\includegraphics[width=0.84\linewidth]{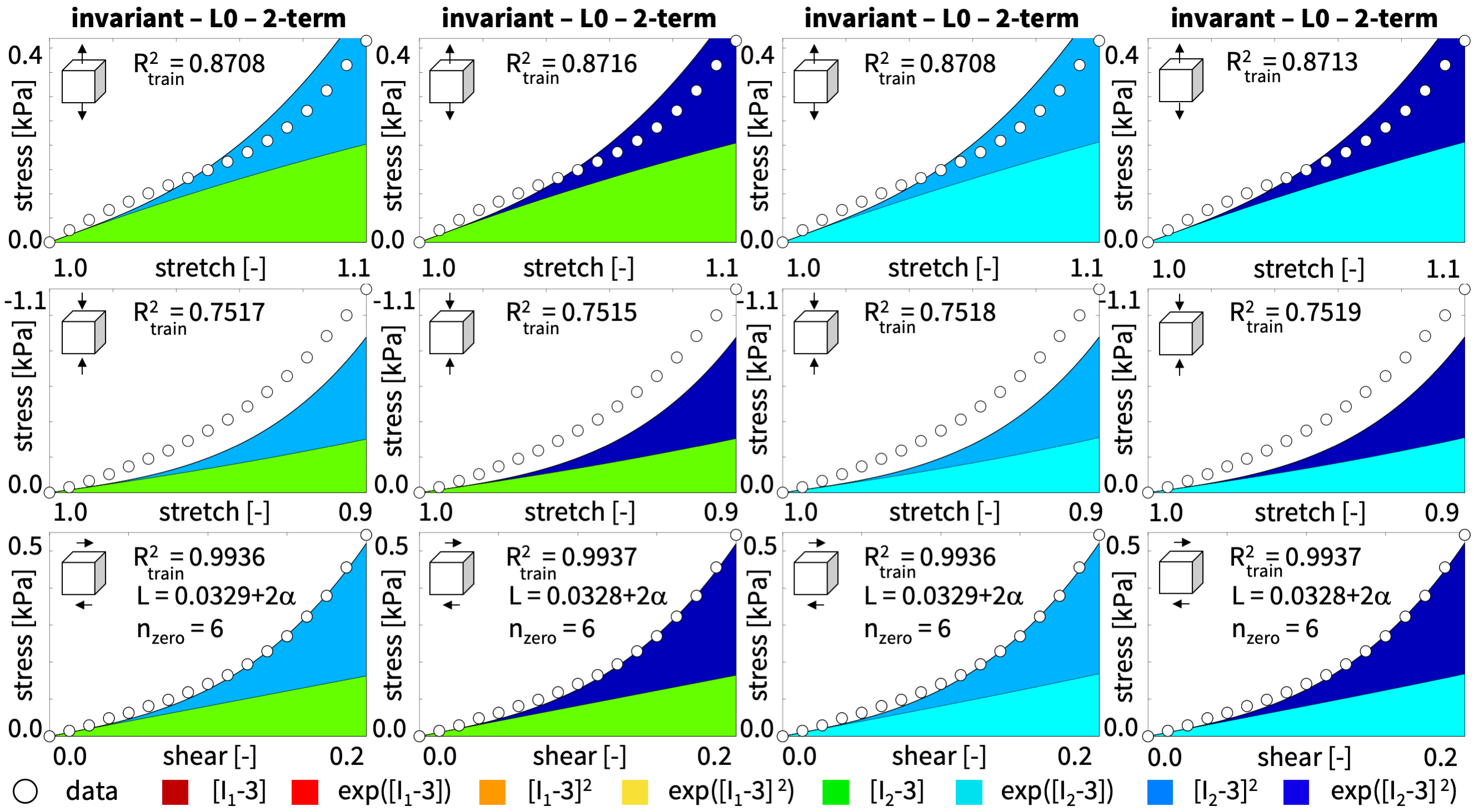}
\caption{{\bf{\sffamily{Discovered best-in-class two-term models of L$_0$ regularized invariant based network.}}} 
Nominal stress as a function of stretch or shear strain for the invariant based constitutive neural network with $L_0$ regularization, 
trained with human gray matter tension, compression, and shear data. 
Circles represent the experimental data. 
Color-coded regions represent the discovered model terms. 
Coefficients of determination ${\text{R}}^2$ indicate the goodness of fit for each individual test;
remaining loss $L$ indicates the quality of the overall fit.}
\label{fig13}
\end{figure}\\[6.pt]
Table \ref{tab01} summarizes the weights and remaining losses of the one- and two-term models of the $L_0$ regularized invariant based neural network. 
The diagonal summarizes the discovered one-term models, the off-diagonal the two-term models. 
The $L_0$ regularization penalizes the one-term models by $\alpha$ and the two-term models by $2\alpha$.
The boldface cells highlight the four {\it{best-in-class models}} of each category. 
Figures \ref{fig12} and \ref{fig13} illustrate the stress-stretch and stress-shear plots of these four one- and two-term models. 
Interestingly, all best-in-class models are models in terms of the second invariant $I_2$ indicated through the cold green-to-blue colors. None of the eight best models includes the first invariant $I_1$ indicated through the warm red-to-yellow colors. 
This finding contradicts the common practice of using primarily the first invariant, 
for example, in popular and widely used neo Hooke model.
Strikingly, the classical Hooke model \cite{treloar48}
represented through the dark red term in both networks
with 
$\psi 
= w_1 \, [\,I_1-3\,]$,
a stiffness-like parameter $w_1=0.7964$\,kPa,
a shear modulus $\mu = 2\,w_1 = 1.5928$\,kPa,
and a remaining loss of $0.0918+\alpha$ 
has the largest remaining loss and performs {\it{the worst}} of all one-term models. 
Similarly, the Demiray model \cite{demiray72} 
represented through the red term 
with 
$\psi 
= w_{2,2} \, [\,\exp (\, w_{1,2}\, [\,I_1-3\,] \,) -1\,]$,
a stiffness-like parameter $w_{2,2}=0.7265$\,kPa,
an exponent $w_{1,2}=1.0763$, 
and a remaining loss of $0.0894+\alpha$,
and the Holzapfel type model \cite{holzapfel00}
represented through the yellow term 
with 
$\psi 
= w_{2,4} \, [\,\exp (\, w_{1,4}\, [\,I_1-3\,]^2 \,) -1\,]$,
a stiffness-like parameter $w_{2,4}=4.2436$\,kPa, 
an exponent $w_{1,4}=4.3048$, 
and a remaining loss of $0.0863+\alpha$,
also perform worse than all one-term second-invariant models.
Yet, these results agree well with our previous observations that the second invariant is better suited to represent the behavior of brain tissue than the first invariant \cite{linka23a}. 
The best-in-class one-term model with the lowest remaining loss is the model with the light blue quadratic term of the second invariant,
\[
  \psi 
= w_7 \, [\,I_2-3\,]^2 \,,
\]
with the stiffness-like parameter $w_7 = 19.5994$\,kPa
for which the stress takes the following form,
$ \ten{P} 
= 2 \, [\,I_2-3\,] \, w_7 \, \partial I_2 / \partial \ten{F} 
- p\,\ten{F}^{-\scas{t}}$.
The best-in-class two-term model is the model with 
the turquoise linear exponential and 
the dark blue quadratic exponential terms of the second invariant,
\[
  \psi
= w_{2,6} \, [\, \exp(\,w_{1,6}\,[\,I_2-3\,])-1\,]
+ w_{2,8} \, [\, \exp(\,w_{1,8}\,[\,I_2-3\,])^2-1\,] \,,
\]
with the stiffness-like parameters $w_{2,6}=0.4835$\,kPa and $w_{2,8}=3.2080$\,kPa,
and the exponential weights        $w_{1,6}=0.8393$ and $w_{1,8}=3.4273$, 
for which the stress takes the following form,
$ \ten{P} 
= [\,                 w_{2,6}\,w_{1,6} \, \exp(\,w_{1,6}\,[\,I_2-3\,])
+ 2 \, [\,I_2-3\,] \, w_{2,8}\,w_{2,8} \, \exp(\,w_{1,8}\,[\,I_2-3\,])^2
  ]\, \partial I_2 / \partial \ten{F} 
- p\,\ten{F}^{-\scas{t}}$.
For this simple example, 
the remaining loss of the best one-term model is 
$0.0594+\alpha$ and 
the remaining loss of the best two-term model is 
$0.0328+2\alpha$.
This implies that, 
for penalty parameters $\alpha < 0.0266$,
the $L_0$ regularization would favor the two-term model, while 
for penalty parameters $\alpha \ge 0.0266$,
the $L_0$ regularization would favor the one-term model.\
These simple considerations highlight the importance of the penalty parameter $\alpha$, which explicitly acts as a {\it{discrete switch}} between the number of terms we want to include in our model.\\[6.pt]
Taken together, this example illustrates the
discrete nature of the $L_0$ regularization as a {\it{discrete combinatorial problem}} that becomes increasingly expensive as the number of model terms increases. 
Our results emphasize the sensitivity of the $L_0$ regularization with respect to the penalty parameter $\alpha$ and highlight the trade-off between {\it{bias}} and {\it{variance}}: Increasing the penalty parameter increases bias, reduces variance, and decreases model complexity as the total number of non-zero terms decreases towards one.
\subsection{Lp regularized principal stretch based neural network}
\label{lam_Lp_network}
Similar to the previous example, we explore the effects of $L_p$ regularization with respect to the two hyperparameters $p$ and $\alpha$, 
but now
for the full principal stretch based network with all eight terms \cite{linka23a},
for training on tension, compression, and shear data from human brain tests \cite{budday17}.
We train the principal based neural network from Figure \ref{fig04} in Section \ref{lam_network} and minimize the loss function from equation (\ref{NN_Lp_loss_Pten_Pcom_Pshr}) with the stress definitions (\ref{P11_lam}) and (\ref{P12_lam}) with three different powers,
$p = [0.5, 1.0, 2.0]$,
and four different penalty parameters,
$\alpha = [0.000, 0.001, 0.010, 0.100]$
using the Adam optimizer \cite{kingma14}.  
\begin{figure}[t]
\centering
\includegraphics[width=0.84\linewidth]{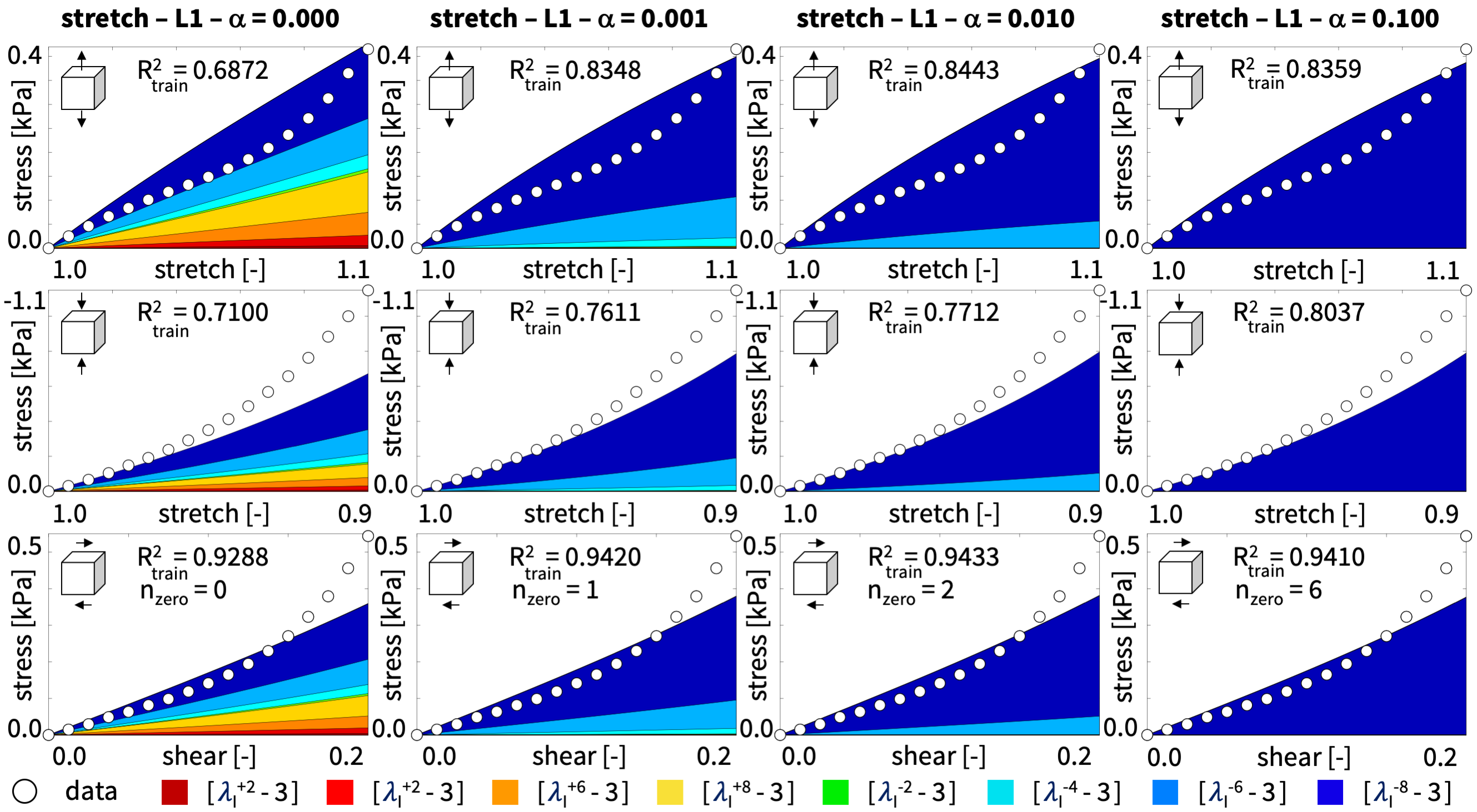}
\caption{{\bf{\sffamily{Discovered models of L1 regularized principal stretch based network.}}} 
Nominal stress as a function of stretch or shear strain for the principal stretch based neural network with $L_1$ regularization for varying 
penalty parameters $\alpha = [\, 0, 0.1, 0.001, 0.0001 \,]$, 
trained with human gray matter tension, compression, and shear data. 
Circles represent the experimental data. 
Color-coded regions represent the stress contributions of the eight model terms. 
Coefficients of determination ${\text{R}}^2$ indicate the goodness of fit.}
\label{fig14}
\centering
\includegraphics[width=0.84\linewidth]{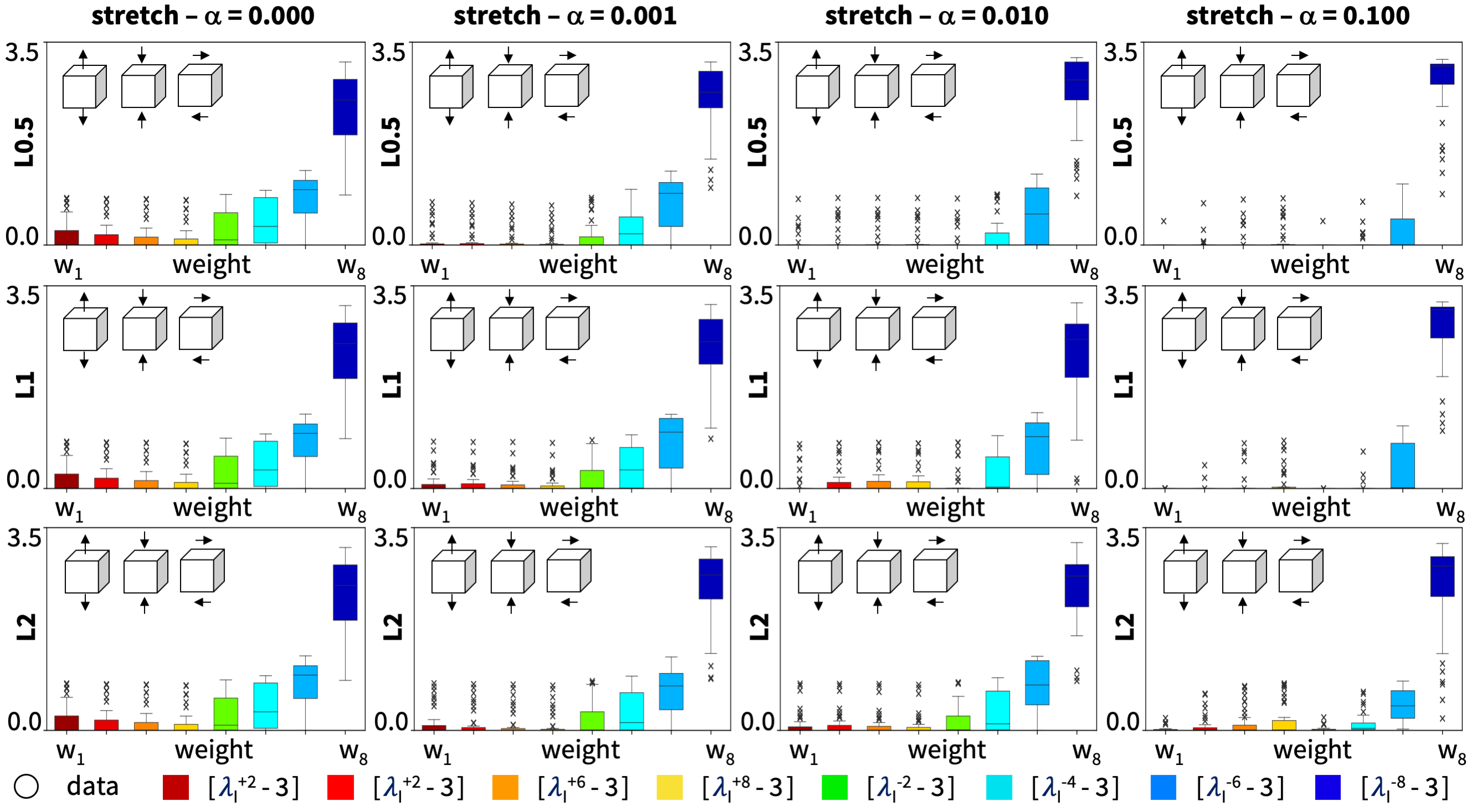}
\caption{{\bf{\sffamily{Discovered models of Lp regularized principal stretch based network.}}} 
Distribution of discovered weights for the principal stretch based neural network with $L_p$ regularization 
for varying powers $p=[\,0.5, 1.0, 2.0]$ and 
penalty parameters $\alpha = [\, 0.000, 0.001, 0.010, 0.100 \,]$. 
Colored boxes indicate the relevance of the eight model terms, with means and standard deviations from $n=100$ realizations with varying initializations of the network weights.}
\label{fig15}
\end{figure}\\[6.pt]
Figure \ref{fig14} summarizes our four discovered models in terms of the nominal stress as a function of stretch or shear strain, with the penalty parameter $\alpha$ increasing from left to right. 
The circles represent the experimental data \cite{budday17}.
The color-coded regions represent the stress contributions of the eight model terms according to Figure \ref{fig04}.
The coefficients of determination ${\text{R}}^2$ quantify the goodness of fit.
Similar to the invariant based network in Section \ref{inv_Lp_network},
the $L_p$ regularized principal stretch based network trains solidly and provides a good fit of the data. 
Without regularization, in the left column, the network discovers all eight non-zero terms. 
As the penalty parameter increases, from left to right, the number of non-zero terms decreases.
For the largest penalty parameter, in the right column, the network discovers a single dominant term, the dark blue $[\,\lambda_i^{-8}-3\,]$ term,
\[
\psi = w_8 \, \mbox{$\sum_{i=3}^3$} \, [\, \lambda_i^{-8} -1 \,]
\]
with a stiffness-like parameter
$w_8 = 0.0534$\,kPa and a stress
$\ten{P} 
= -8\, w_8 \, \mbox{$\sum_{i=3}^3$} \, [\, \lambda_i^{-9} -1 \,] \,
\vec{n}_i \otimes \vec{N}_i - p\,\ten{F}^{-\scas{t}}$.\\[6.pt]
Figure \ref{fig15} summarizes the discovered weights for the principal stretch based network with $L_p$ regularization 
for varying powers $p=[\,0.5, 1.0, 2.0]$ and 
penalty parameters $\alpha = [\, 0.000, 0.001, 0.010, 0.100 \,]$. 
For all twelve combinations of the two hyperparameters, we perform a total of $n=100$ training runs each, with varying initial conditions for the network weights $w_i = \{\,w_1,...,w_8\,\}$, such that each of the four models in Figure \ref{fig14} is the result of one of the $L_1$ regularized training runs in the middle row. The colored boxes in Figure \ref{fig15} indicate the relevance of the eight model terms, with their means and standard deviations.
In contrast to the invariant based network in Section \ref{inv_Lp_network}, the principal stretch based network consistently discovers similar terms across all three regularizations, with a clear preference for the dark blue $[\,\lambda_i^{-8}-3\,]$ term.
The fact that all networks robustly discover {\it{similar}} terms 
is a result of the {\it{convex}} nature 
of the underlying linear regression problem 
associated with the principal stretch based network and
indicates the existence of a single {\it{unique global minimum}}.
However, the $L_{0.5}$ and $L_{1}$ regularized networks gradually drop more non-zero terms as the penalty parameter increases, while the 
$L_{2}$ regularized network maintains all eight terms. 
As we had already anticipated from comparing Figures \ref{fig03} and \ref{fig05}, the functional base of the principal stretch based network is more collinear than the base of the invariant based network, which result in a more gradual shift of the active weights, from $w_1$ towards $w_8$, as the penalty parameter increases. 
The fact that all three regularizations converge to the boundary of our domain, 
the dark blue $[\,\lambda_i^{-8}-3\,]$ term with the minimum exponent of minus eight, suggests that the true best fit might lay outside the current parameter range, with even smaller exponents. This agrees well with previous studies\cite{budday17,stpierre23} that have discovered one-term Ogden models with exponents of
$[\,\lambda_i^{-18}-3\,]$ and
$[\,\lambda_i^{-19}-3\,]$.\\[6.pt]
Taken together, this example illustrates that model discovery with $L_p$ regularization {\it{generalizes}} well to different network types, irrespective of whether the terms are invariant or principal stretch based. 
Comparing both network types reveals that the method is sensitive to the {\it{nonlinear}} vs {\it{linear}} nature of the underlying regression problem: 
The nonlinear invariant based network alternates between different dominant terms associated with multiple local minima, while the
linear principal stretch based network consistently discovers similar terms associated with a single unique global minimum. 
\section{Conclusion and recommendations} 
\label{conclusion}
$L_p$ regularization is a powerful technology to finetune the training process of a neural network. In automated model discovery, it provides the critical missing piece of the puzzle that enables a controlled down-selection of the discovered terms and focus on the most important features of the model while putting less emphasis on minor effects. By promoting sparsity of the parameter vector, $L_p$ regularization inherently improves interpretability and provides valuable insights into the underlying nature of the data. Importantly, $L_p$ regularization introduces two hyperparameters: the power $p$ by which it penalizes the individual model parameters, and the penalty parameter $\alpha$ by which it scales the relative importance of the regularization loss in comparison to the neural network loss. Both parameters enable a precise control of model discovery from data and it is crucial to understand their mathematical subtleties, computational implications, and physical effects. Here we reviewed the mathematics and computation of the most common representatives of the $L_p$ family, and demonstrated their features in terms of two classes of constitutive neural networks, invariant and principal stretch based, trained with both, synthetic and real data. Training with synthetic data proved to be robust and stable, and generally provides excellent metrics for quality control since we know the exact solution. However, it remains a toy problem that fails to reveal the true usefulness in practical real-world applications. Training with real data was algorithmically robust, but challenging, since we know nothing about the exact solution.
Our study uses neural networks as a tool for linear and nonlinear regression. We acknowledge that our results can be interpreted and well understood without resorting neural networks and generalize naturally to other regression techniques including symbolic regression, genetic programming, or system identification. \\[6.pt]
For conciseness, we have limited the scope of the present review: 
First, we only considered small networks with no more than eight terms, but point out that the automated model discovery generalizes well to isotropic networks with 12 terms \cite{linka23a}, transversely isotropic networks with 16 terms \cite{linka23b}, two-fiber family networks with 16 terms \cite{peirlinck23}, and orthotropic networks with 32 terms.
Second, we trained on all available data and did not investigate splitting the data into train and test sets, which we have done in our previous work \cite{linka23a,stpierre23}.
Third, we did not explicitly study the effects of controlled noise, but point out that Figures \ref{fig08} to \ref{fig15} are all based on real experimental data with real natural noise.
Fourth, we did not further explore hybrid top down approaches like SINDy \cite{brunton16}, since the nonlinear nature of our optimization problem does not guarantee that we easily find the global minimum \cite{zhang19}, from which we could initiate a sequential thresholded least squares down-selection; 
Finally, we have not yet investigated the effects of $L_p$ regularization on uncertainty quantification, something we are currently exploring in a separate Bayesian approach.\\[6.pt]
We would like to share the most important lessons we have learnt throughout this study:\\[6.pt]
\noindent{\bf{\sffamily{Normalize first!}}} 
We cannot overstate the importance of normalizing. Clearly, while normalizing is less of an issue in linear regression, it is critical in nonlinear regression. This holds for both the training data, illustrated in Figures \ref{fig03} and \ref{fig05}, and the weights, illustrated in Figure \ref{fig11}. The loss function typically contains several terms of different magnitude that compete during minimization. It proves important to normalize by the number of data sets in each category, the magnitude of the tensile, compressive, and shear stresses, and the magnitude of the weights to balance the impact of the individual contributions.\\[6.pt]
\noindent{\bf{\sffamily{L$_{\bf{0}}$ regularization is the most honest member of the Lp family.}}} 
$L_0$ regularization or subset selection is honest, transparent, and unbiased. Its penalty parameter $\alpha$ acts as a {\it{direct switch}} to select the desired number of terms. It is the {\it{only}} member of the $L_p$ family that explicitly controls the balance between the number of terms and the goodness of fit, illustrated in Figure \ref{fig11}. While $L_0$ regularization across the entire network translates into an expensive {\it{NP hard discrete combinatorial problem}} of the order of $2^n$, we recommend to begin any discovery by running an $L_0$ regularization for all possible one- and two-term models to determine the best-in-class models of each category and identify the dominant terms, similar to Figures \ref{fig12} and \ref{fig13} and Table \ref{tab01}. Importantly, in nonlinear regression, the best-in-class $n$-term model may actually {\it{not}} be a subset of the best-in-class $(n+1)$-term model, and {\it{successively removing terms}} like in iterative pruning \cite{han15} or sequential thresholding least squares \cite{brunton16} might not be a viable solution. Instead, running $L_0$ regularization for all possible one- and two-term models provides a quick first insight into the nature and hierarchy of the best-in-class models \cite{peirlinck23a}. From this initial first glimpse, we can proceed by {\it{successively adding terms}}. In addition, from the best-in-class one-term models, we
can use the discovered weights $w_{i,L_0}$ to {\it{initialize}} the weights for higher order runs and to {\it{normalize}} the weights in the regularization term, 
$\alpha \, ||\, \vec{\theta} \,||_p^p 
= \sum_i^{n_{\rm{para}}} |\, w_i / w_{i,L_0} \,|^p$. \\[6.pt]
\noindent{\bf{\sffamily{L$_{\bf{1}}$ regularization is powerful for subset selection, but needs large penalty parameters to be effective.}}} 
$L_1$ regularization or lasso promotes sparsity by reducing a large subset of parameters exactly to zero. Notably, for all the examples in our study, this down-selection required quite large penalty parameters $\alpha$--often on the order of one--to work effectively. This not only affects the magnitude of the discovered parameters, but often also the discovered model itself. For example, for $\alpha=0.1$, the $n=100$ independent realizations of the $L_1$ regularization in Figure \ref{fig10} alternate between the green and turquoise one-term models, while the unbiased plain $L_0$ regularization in Figure \ref{fig12} and Table \ref{tab01} ranks these two models clearly behind the blue and dark blue one-term models. To identify regularization-induced bias, we recommend to always compare the results of the $L_1$ regularization against the best-in-class low-term models of the $L_0$ regularization. This comparison is simple and inexpensive, and provides valuable insights into the magnitude of selection bias and the aggressiveness of the $L_1$ regularization.\\[6.pt]
\noindent{\bf{\sffamily{L$_{\bf{0.5}}$ regularization promotes sparsity for small penalty parameters, but suffers from multiple local minima.}}} 
$L_{0.5}$ regularization addresses the shortcomings of the classical $L_1$ regularization by down-selecting more aggressively, requiring smaller penalty parameters, and introducing less bias. While $L_{0.5}$ regularization works well in practice, it is computationally challenging. Its non-convexity introduces multiple local minima, indicated through the first rows in Figures \ref{fig06}, \ref{fig07} and \ref{fig08}, and through the green and turquoise one-term models in Figure \ref{fig10}, and the blue and dark blue one-term models in Figure \ref{fig15}. To avoid getting stuck in a local minimum, we highly recommend exploring different initialization strategies for the network weights. Specifically, we were able to robustly identify multiple local minima by initializing the weights with the $L_0$ regularized weights $w_{i,L_0}$. Alternatively, we could gradually ramp up the effect of regularization by starting with a penalty parameter $\alpha=0$ and smoothly increase it to a desired strength, essentially by moving from left to right in Figure \ref{fig07}.
For quality control, we recommend comparing the remaining loss of each converged run against the remaining $L_0$ regularized baseline loss as reported in Table \ref{tab01}.\\[6.pt]
\noindent{\bf{\sffamily{L$_{\bf{2}}$ regularization promotes stability, but is not suited for subset selection.}}} 
$L_2$ regularization, by design, is not suited to reduce a subset of terms exactly to zero. Instead, it maintains all terms as indicated in the bottom rows of Figures \ref{fig10} and \ref{fig15}, each for $n=100$ runs. From Figures \ref{fig06}, \ref{fig07}, and \ref{fig08} we conclude that, for increasing penalty parameters $\alpha$, $L_2$ regularization reduces outliers by {\it{first}} bringing the weights closer together and {\it{then}} collectively reducing them toward zero. Clearly, $L_2$ regularization improves convexity, which makes model discovery more robust and more stable. However, it not only fails to down-select the number of terms, but also strongly biases the solution away from the minimum of the pure network loss towards the minimum of the regularization loss. We do {\it{not}} recommend using $L_2$ regularization, or any other member of the $L_p$ regularization family with powers larger than one, $p>1$, to increase sparsity and improve interpretability. \\[-16.pt]
\begin{table}[h]
\centering
\caption{{\bf{\sffamily{Lp regularization.}}} 
Comparison of special cases, advantages, disadvantages, and references.}
\vspace*{0.2cm} 
\footnotesize
\renewcommand{\arraystretch}{0.9}
\label{tab02}
\hspace*{-2.1cm}
\begin{tabular}{|l||l|l|l|l||l|}
\hline              
      & \makecell{\bf{\sffamily{algorithm}}} 
      & \makecell{\bf{\sffamily{regularization}}} 
      & \makecell{\bf{\sffamily{advantages}}} 
      & \makecell{\bf{\sffamily{disadvantages}}} 
      & \makecell{\bf{\sffamily{refs}}} \\  \hline \hline
        {\bf{\sffamily{L$_{\bf{0}}$}}}
      & \makecell{\bf{\sffamily{subset selection}}}  
      & \makecell{$\alpha \, || \vec{\theta} ||_0$ \\
        $||\vec{\theta} \, ||_0 = \sum_{i}  I(w_i\ne0) $}    
      & \makecell[l]{
        $\bullet $ penalizes number of non-zero terms \\
        $\bullet $ term count is inherently unbiased  \\
        $\bullet $ conceptually simple and honest\\
        $\bullet $ {\it{promotes sparsity}} \\
        $\bullet $ improves interpretability \\
        $\bullet $ valuable insight for one or two terms}
      & \makecell[l]{
        $\bullet $ solves {\it{discrete combinatorial problem}}  \\        
        $\bullet $ computationally {\it{expensive}}, ${\mathcal{O}}(2^n)$  \\        
        $\bullet $ but manageable for one- or two-terms}
      & \cite{frank93}  \\  \hline
        {\bf{\sffamily{L$_{\bf{0.5}}$}}}
      & \makecell{compromise\\
                  between $L_0$ and $L_1$}   
      & \makecell{$\alpha \, || \vec{\theta} ||_{0.5}^{0.5}$ \\
        $||\vec{\theta} \, ||_{0.5}^{0.5} = \sum_{i} \sqrt{| \, w_i \, |}$}    
      & \makecell[l]{
        $\bullet $ improved efficiency compared to $L_0$  \\
        $\bullet $ improved sparsity compared to $L_1$ \\
        $\bullet $ reduces some parameters exactly to zero \\
        $\bullet $ works even for smaller $\alpha$ and less bias}
      & \makecell[l]{
        $\bullet $ non-convex, multiple local minima   \\
        $\bullet $ increased computational complexity}
      & \cite{frank93}  \\  \hline
        {\bf{\sffamily{L$_{\bf{1}}$}}}
      & \makecell{{\bf{\sffamily{lasso}}}\\
        least absolute \\shrinkage and \\selection operator}   
      & \makecell{$\alpha \, || \vec{\theta} ||_1$ \\
        $||\vec{\theta} \, ||_1 = \sum_{i} | \, w_i \, |$}    
      & \makecell[l]{
        $\bullet $ weighs all components equally \\
        $\bullet $ less sensitive to outliers than $L_2$\\
        $\bullet $ reduces some parameters exactly to zero \\
        $\bullet $ {\it{promotes sparsity}} \\
        $\bullet $ improves interpretability}
      & \makecell[l]{
        $\bullet $ not strictly convex, {\it{local minima}}  \\
        $\bullet $ emphasizes selective effects \\
        $\bullet $ introduces bias, {\it{inaccurate for large $\alpha$}} }
      & \cite{tibshirani96}  \\  \hline
        {\bf{\sffamily{L$_{\bf{1/2}}$}}}
      & \makecell{\bf{\sffamily{elastic net}}\\
                  compromise\\
                  between L1 and L2}
      & \makecell{$\alpha_1 \, || \vec{\theta} ||_1
                  +\alpha_2 \, || \vec{\theta} ||_2^2$ \\
        $||\vec{\theta} \, ||_1 = \sum_{i} | \, w_i \, |\;$ \\
        $||\vec{\theta} \, ||_2^2 = \sum_{i} | \, w_i \, |^2$}
      & \makecell[l]{
        $\bullet $ improved stability compared to $L_1$  \\
        $\bullet $ improved sparsity compared to $L_2$}
      & \makecell[l]{
        $\bullet $ increased computational complexity}
      & \cite{zou05}  \\  \hline
        {\bf{\sffamily{L$_{\bf{2}}$}}}
      & \makecell{\bf{\sffamily{ridge regression}}}
      & \makecell{$\alpha_2 \, || \vec{\theta} ||_2^2$ \\
        $||\vec{\theta} \, ||_2^2 = \sum_{i} | \, w_i \, |^2$}
      & \makecell[l]{
        $\bullet $ uses components squared \\
        $\bullet $ reduces outliers, {\it{improves predictability}} \\
        $\bullet $ increases robustness  \\
        $\bullet $ promotes stability }
      & \makecell[l]{
        $\bullet $ introduces bias  \\
        $\bullet $ moves parameters towards each other  \\
        $\bullet $ reduces but maintains {\it{all}} parameters  \\
        $\bullet $ does {\it{not}} promote sparsity}      
      & \cite{hoerl70} \\  \hline
\end{tabular} 
\vspace*{-0.4cm}
\end{table} 
\clearpage
Table \ref{tab02} provides a side-by-side comparison of the $L_p$ regularizations we explored throughout this study along with their advantages, disadvantages, and references. \\[6.pt]
\noindent{\bf{\sffamily{Densifying instead of sparsifying.}}} 
The {\it{closure problem}} is a common challenge in both fluid and solid mechanics. It refers to the difficulty of fully specifying the constitutive equations that relate stresses and strains and characterize the material behavior. 
In fluid mechanics, the closure problem is closely related to turbulence modeling, 
where it approximates intricate interactions between different scales, and can be well represented through {\it{polynomials}} \cite{brunton16}. 
In solid mechanics, the closure problem characterizes complex material behaviors at the microscopic scale, and is traditionally often represented through a combination of {\it{polynomials}} \cite{flaschel21}, {\it{exponentials}} \cite{demiray72,holzapfel00}, {\it{logarithms}} \cite{gent96}, and {\it{powers}} \cite{ogden72,valanis67}.
In the context of model discovery, assuming perfect data, polynomial models translate into a convex linear optimization problem with a single unique global minimum, while exponential, logarithmic, or power models translate into a non-convex nonlinear optimization problem with possibly multiple local minima. 
For convex discovery problems with a unique global minimum, inducing sparsity has been well established through a top down approach in which we first calculate a dense parameter vector at the global minimum, and then {\it{sparsify the parameter vector}} by sequentially thresholding and removing the least relevant terms \cite{brunton16,zhang19,flaschel21,wang21}.
For non-convex discovery problems with multiple local minima, this approach is infeasible since different initial conditions may result in different solutions with non-unique  parameter vectors \cite{ogden04}. 
Instead of trying to sparsify a dense parameter vector, we recommend to gradually {\it{densify the parameter vector}} from scratch. 
This bottom up approach iteratively solves the discrete combinatorial problem and densifies the parameter vector by sequentially adding the most relevant terms \cite{nikolov22}.
Importantly, instead of solving the NP hard discrete combinatorial problem associated with a complete $L_0$ regularization that screens all possible combinations of terms at ${\mathcal{O}}(2^n)$, 
we recommend to gradually add terms, starting with the 
best-in-class one-term model at ${\mathcal{O}}(n)$, 
adding a second term at ${\mathcal{O}}(n)$, 
and repeating addition until the incremental improvement of the overall loss function meets a user-defined convergence criterion.
At most, this algorithm involves ${\mathcal{O}}(n^2)$ evaluations of the loss function 
to land on a fully populated dense parameter vector.
Importantly, for non-convex model discovery problems, 
this algorithm--while cost effective and well-rationalized--is not guaranteed to converge to the global minimum. 
Instead of successively adding up to $n$ terms, 
for practical purposes, 
it is often sufficient to limit the number of desirable terms 
to one, two, three or four, 
and identify the {\it{best-in-class model}} of each class, 
which requires a discrete comparison of 
$(\, 8! / (n! (8-n)!) \,)$ 
discrete models, in our case 8, 28, 56, or 70 \cite{peirlinck23a}.
Out of all possible discovery algorithms, 
this is the most honest, unbiased, and transparent approach.\\[6.pt]
Taken together, our study suggests that $L_p$ regularized constitutive neural networks are a powerful technology for automated model discovery that allows us to identify interpretable constitutive models from data. 
We anticipate that our results generalize to $L_p$ regularization for model discovery with other techniques such as symbolic regression or system identification, and, more broadly, to model discovery in other fields such as biology, chemistry, or medicine.
The ability to discover new knowledge from data could have tremendous applications in generative material design where it could shape the path to manipulate matter, alter properties of existing materials, and discover new materials with targeted properties.
\subsection*{Author contributions}
JAMC and SRSP 
method development, 
simulation,
data analysis, 
result interpretation, 
manuscript writing;
KL and EK 
study design,
method development,
simulation, 
data analysis,
result interpretation,
manuscript writing.
\subsection*{Acknowledgments}
This work was supported 
by the National Science Foundation Graduate Research Fellowship 
to Jeremy McCulloch and Skyler St. Pierre, 
by the DAAD Fellowship to Kevin Linka, and 
by the NSF CMMI Award 2320933 {\it{Automated Model Discovery for Soft Matter}} 
to Ellen Kuhl.
\subsection*{Conflict of interest}
The authors declare no potential conflict of interests.

\end{document}